\documentclass[12pt]{article}
\def\ParSkip{}

\usepackage{url}
\usepackage[normalem]{ulem}
\usepackage{amsfonts}
\usepackage{amsmath,amssymb,color,bbm}
\usepackage{mathrsfs}
\usepackage{graphicx}
\usepackage{enumitem}
\usepackage{multirow}
\usepackage{bm}
\usepackage{natbib}
\usepackage{verbatim}
\usepackage{float}
\usepackage[toc,page]{appendix}
\usepackage{setspace,fullpage}
\usepackage{setspace}
\usepackage{adjustbox}
\usepackage{rotating}
\usepackage{subcaption}
\usepackage{amsthm}
\usepackage{booktabs,dcolumn,caption}
\usepackage{xltabular}
\usepackage{booktabs}
\usepackage{longtable}
\usepackage{tabularx}
\usepackage{threeparttablex} 
\usepackage{accents}
\usepackage{algpseudocode}
\usepackage{algorithm}

\usepackage{xr}

\usepackage{tikz}
\usetikzlibrary{shapes.geometric, arrows, positioning}

\tikzstyle{startstop} = [rectangle, rounded corners, 
minimum width=1.5cm, 
minimum height=1cm,
text centered, 
draw=black]

\tikzstyle{io} = [trapezium, 
trapezium stretches=true, 
trapezium left angle=70, 
trapezium right angle=110, 
minimum width=3cm, 
minimum height=1cm, text centered, 
draw=black, fill=blue!30]

\tikzstyle{process} = [rectangle, 
minimum width=3cm, 
minimum height=1cm, 
text centered, 
text width=3cm, 
draw=black, 
fill=orange!30]

\tikzstyle{decision} = [diamond, 
minimum width=2cm, 
minimum height=1cm, 
text centered, 
draw=black]
\tikzstyle{arrow} = [thick,->,>=stealth]

\usepackage{array}
\newcolumntype{L}[1]{>{\raggedright\let\newline\\\arraybackslash\hspace{0pt}}p{#1}}
\newcolumntype{C}[1]{>{\centering\let\newline\\\arraybackslash\hspace{0pt}}p{#1}}
\newcolumntype{R}[1]{>{\raggedleft\let\newline\\\arraybackslash\hspace{0pt}}p{#1}}

\ifdefined\TimesFont 
\usepackage{times} 
\fi

\ifdefined\ParSkip 
\usepackage{parskip} 
\fi

\newtheorem{theorem}{Theorem}

\newtheorem{corollary}{Corollary}

\newtheorem{lemma}{Lemma}

\newtheorem{remark}{Remark}

\newtheorem{assumption}{Assumption}
\newcounter{assumptionitem}[assumption]  
\renewcommand{\theassumptionitem}{\theassumption.\arabic{assumptionitem}}  
\newcommand{\assumptionitem}{%
  \refstepcounter{assumptionitem}
  \makebox[2em][l]{\theassumptionitem.}
}

\makeatletter
\newcommand*\rel@kern[1]{\kern#1\dimexpr\macc@kerna}
\newcommand*\widebar[1]{%
  \begingroup
  \def\mathaccent##1##2{%
    \rel@kern{0.8}%
    \overline{\rel@kern{-0.8}\macc@nucleus\rel@kern{0.2}}%
    \rel@kern{-0.2}%
  }%
  \macc@depth\@ne
  \let\math@bgroup\@empty \let\math@egroup\macc@set@skewchar
  \mathsurround\z@ \frozen@everymath{\mathgroup\macc@group\relax}%
  \macc@set@skewchar\relax
  \let\mathaccentV\macc@nested@a
  \macc@nested@a\relax111{#1}%
  \endgroup
}
\makeatother

\DeclareMathOperator*{\argmin}{argmin}

\DeclareMathOperator{\rank}{rank}

\DeclareMathOperator{\diag}{diag}

\DeclareMathOperator{\tr}{tr}

\newcommand{\aaa}{{\boldsymbol \alpha}}
\newcommand{\bbb}{{\boldsymbol \beta}}

\newcommand{\rrr}{\boldsymbol \gamma}
\newcommand{\GGamma}{\boldsymbol \Gamma}

\newcommand{\ttt}{\boldsymbol \theta}
\newcommand{\TTT}{\boldsymbol \Theta} 
\newcommand{\SG}{\boldsymbol \Sigma}

\newcommand{\aaaa}{\mathbf a}
\newcommand{\A}{\mathbf A}
\newcommand{\AAA}{\mathcal A}

\newcommand{\BB}{\mathbf B}
\newcommand{\BBB}{\mathcal B}
\newcommand{\bb}{\mathbf b}

\newcommand{\DD}{\mathbf D}

\newcommand{\DDDD}{\mathcal D}
\newcommand{\EE}{\mathbf 1}

\newcommand{\ee}{\mbox{$\mathbf e$}}
\newcommand{\EEE}{\mathbb E}
\newcommand{\EEEE}{\mathcal E}

\newcommand{\FF}{\mathbf F}

\newcommand{\GG}{\mathbf G}

\newcommand{\JJ}{\mathbf J}

\newcommand{\MM}{\mathbf M}

\newcommand{\XX}{\mathbf X}
\newcommand{\YY}{\mathbf Y}

\newcommand{\HHH}{\mathcal H}
\newcommand{\hh}{\mathbf h}
\newcommand{\II}{\mathbf I}
\newcommand{\III}{\mathcal I}

\newcommand{\OOO}{\mathbb O}

\newcommand{\PP}{\mathbf P}
\newcommand{\PPP}{\mathcal P}
\newcommand{\R}{\mathbb R}
\newcommand{\RR}{\mathbf R}

\newcommand{\SSS}{\mathbf S}

\newcommand{\uu}{\mathbf u}
\newcommand{\UU}{\mathbf U}
\newcommand{\UUU}{\mathcal U}
\newcommand{\VV}{\mathbf V}
\newcommand{\vv}{\mathbf v}
\newcommand{\VVV}{\mathcal V}
\newcommand{\WW}{\mathbf W}

\newcommand{\xx}{\mathbf x}

\newcommand{\Y}{\mathbf Y}

\newcommand{\Z}{\mathbf Z}

\newcommand{\QQ}{\mathbf Q}

\newcommand{\1}{\uppercase\expandafter{\romannumeral1}}
\newcommand{\2}{\uppercase\expandafter{\romannumeral2}}

\newcommand{\0}{\textbf{0}}

\newcommand{\ve}{\text{vec}}

\newcommand{\SVD}{\text{SVD}}
\newcommand{\Eigen}{\text{Eigen}}

\renewcommand{\arraystretch}{1.2}

\newcommand{\keywords}[1]{\textbf{Keywords:} #1}

\usepackage[hidelinks]{hyperref}

\definecolor{codexgreen}{RGB}{0,128,96}

\begin{document}

\title{\vspace{-2em}Model-Based Learning of Latent Euclidean Embedding from Relational Data}
\author{Yinghang Chen$^{1}$\thanks{Email: \texttt{yh.chen018@gmail.com}.},
Yunxiao Chen$^{2}$\thanks{Email: \texttt{y.chen186@lse.ac.uk}.},
and Haoran Zhang$^{1}$\thanks{Corresponding author. Email: \texttt{zhanghr@sustech.edu.cn}.}\\[6pt]
\small $^{1}$Department of Statistics and Data Science\\[-3pt]
\small Southern University of Science and Technology\\[-3pt]
\small 1088 Xueyuan Avenue, Nanshan District, Shenzhen, Guangdong 518055, P.R. China\\[2pt]
\small $^{2}$Department of Statistics, London School of Economics and Political Science\\[-3pt]
\small Houghton Street, London WC2A 2AE, United Kingdom
}
\date{}
\maketitle
\vspace{-1.5em}

\onehalfspacing
\begin{abstract}%
Visualizing and interpreting the dependence structure in relational data often benefits from geometric representations via low-dimensional Euclidean embeddings. However, existing approaches are typically restricted to specific response types, rely on burdensome high-dimensional nonconvex optimization, or lack uniform theoretical guarantees for simultaneous coordinate recovery. To address these limitations, we introduce a unified, distance-based latent space framework for both unipartite and bipartite relational data---accommodating continuous, binary, ordinal, and count-valued observations---via a modular two-stage procedure that decouples signal estimation from geometric recovery. 
First, we estimate the low-rank signal matrix using spectral denoising.
For models with nonlinear link functions, we further apply one-step row- and column-wise refinement to achieve uniform entrywise accuracy. Second, we recover the latent configurations through direct spectral analysis for unipartite data and a low-dimensional alignment optimization for bipartite data. We establish uniform recovery guarantees that characterize the propagation of error from signal estimation to the embedding space. To illustrate the scalability and interpretability of our approach, we apply the proposed method to two real-world datasets: a statistician citation network and European Parliament voting records.
\end{abstract}

\noindent\keywords{relational data, Euclidean embedding, low-rank model, network data, latent space model}\par

\doublespacing

\section{Introduction}\label{sec:intro}
Relational data arise in a wide range of scientific and machine-learning applications, including social and information networks, recommender systems, voting records, citation patterns, and user--item interactions. Such data record associations either among objects from a single set (unipartite relational data, such as networks) or between objects from two distinct sets (bipartite relational data, such as user--item interactions). These two settings are geometrically connected. After pooling the two sets, a bipartite relation can be represented as a symmetric relation matrix in which the cross-set blocks are observed, whereas the two within-set blocks are structurally missing. 

A central task is to uncover the relational structure among these objects for prediction, scientific interpretation, and exploratory visualization. 
A broad range of low-dimensional representation methods has been developed for this purpose. Multidimensional scaling (MDS) represents one set of objects from pairwise dissimilarities \citep{torgerson1952multidimensional, borg2005modern}, whereas multidimensional unfolding (MDU) jointly represents two sets of objects from a rectangular matrix of cross-set relations or preferences \citep{schonemann1970metric, borg2005modern}. Latent space models assign embedding vectors to network nodes and relate the observed network to these embedding vectors \citep{hoff2002latent, handcock2007model}. Correspondence analysis (CA) and biplots provide joint low-dimensional displays of the rows and columns of a data table \citep{gabriel1971biplot, gabriel2002goodness}, while factor models represent two sets of objects through latent factor scores and loading vectors \citep{bai2003inferential, skrondal2004generalized, bartholomew2008analysis}. More broadly, machine learning has developed a diverse family of low-dimensional embedding methods for visualization and representation learning, including manifold and spectral methods, neighbor embeddings, and neural parametric approaches \citep{tenenbaum2000global, belkin2003laplacian, hinton2006reducing, vandermaaten2008visualizing, mcinnes2018umap, debodt2025low}. Although all these methods produce low-dimensional coordinates, proximity in
the resulting representation does not have a universal interpretation. Depending on the method, the coordinates are designed to represent pairwise distances, inner-product interactions, standardized departures from independence, or local neighborhood relations.

\begin{figure}[!htb]
    \centering

    \begin{subfigure}[t]{0.48\textwidth}
        \centering
        \includegraphics[width=\linewidth]
        {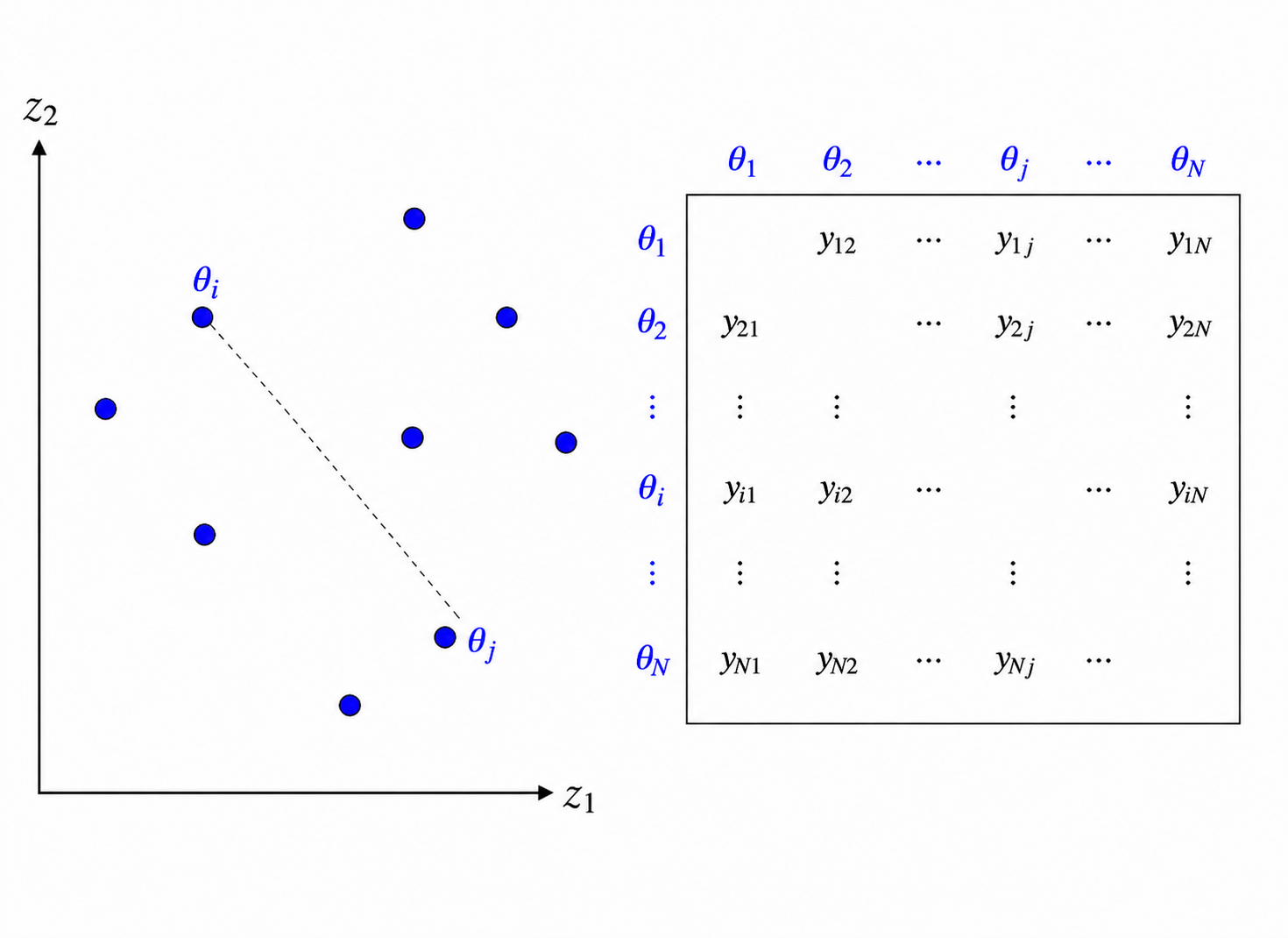}
    \end{subfigure}
    \hfill
    \begin{subfigure}[t]{0.48\textwidth}
        \centering
        \includegraphics[width=\linewidth]
        {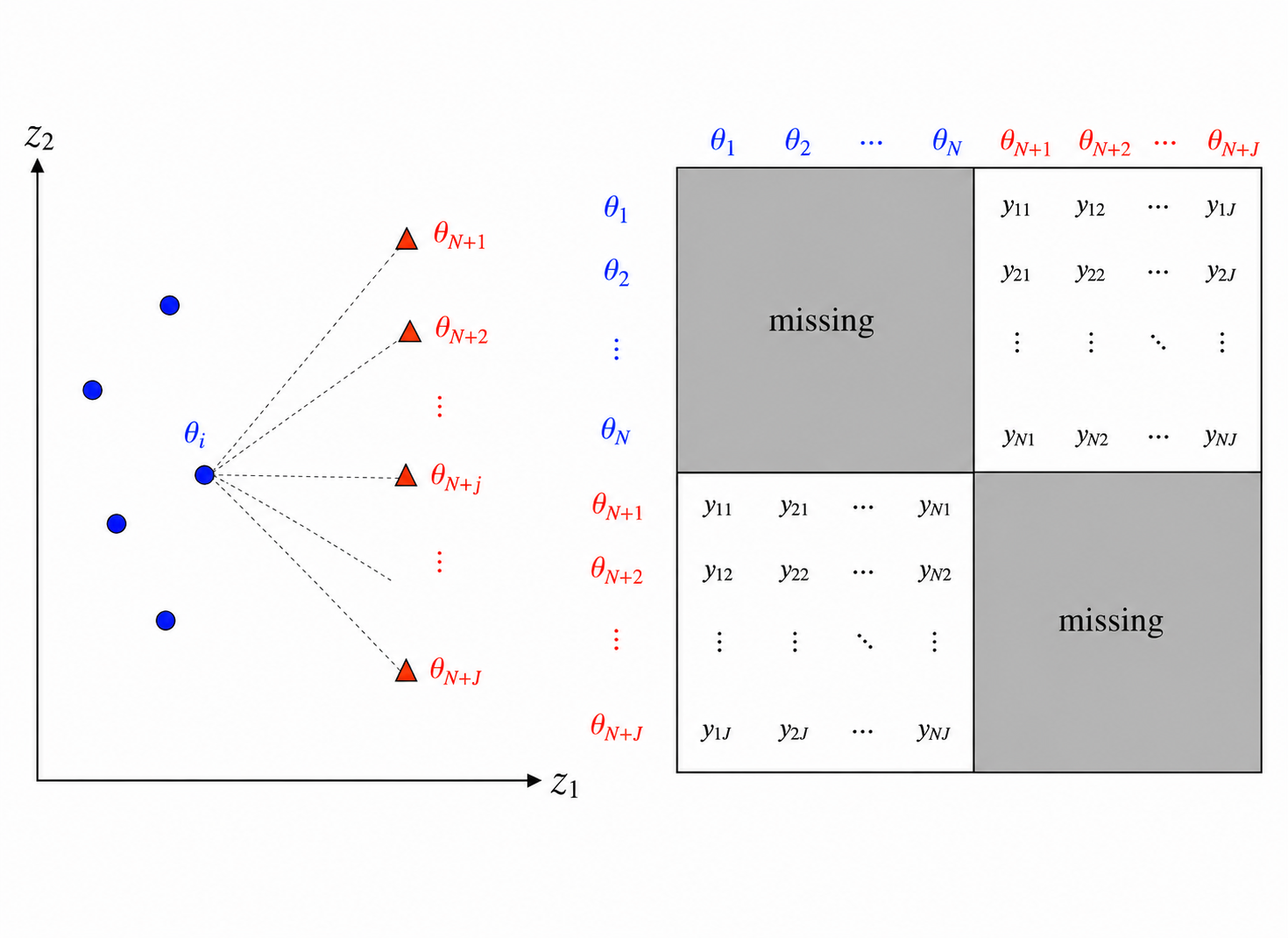}
    \end{subfigure}

    \caption{
Schematic illustrations of latent Euclidean configurations for
unipartite (left) and bipartite (right) relational data.
In the unipartite setting, relations are observed among objects from
a single set; in the bipartite setting, objects from two distinct sets
are jointly represented in a common Euclidean space while only
cross-set relations are observed.
    }
    \label{fig:illustration}
\end{figure}

Among these representations, a common low-dimensional Euclidean configuration is particularly attractive because it places all objects in a shared metric space, so that Euclidean distance has a direct relational interpretation \citep{schonemann1970metric,borg2005modern}.
Depending on the application, proximity may encode similarity,
preference, approval, affinity, or interaction intensity. The resulting geometry
can reveal clusters, latent continua, polarization, and objects occupying
intermediate positions. 
Figure~\ref{fig:illustration} provides a schematic illustration
of this common Euclidean view for unipartite and bipartite relational data, where the smaller the distance between objects in the embedding space, the stronger their similarity tends to be as reflected by the observed relational data. 

Despite the utility of low-dimensional representations, existing approaches have four major limitations. 
\begin{enumerate}

\item \textbf{Validity of joint geometric representation}. Many methods can provide a joint display of objects in a low-dimensional space, but this does not necessarily provide a valid joint metric geometry. In that case, the interpretation of low-dimensional coordinates depends on the structure that the method is designed to represent.
For example, factor models represent row objects by factor
scores and column objects by loadings, whose bilinear signal is
invariant under paired nonsingular transformations
\citep{skrondal2004generalized,bartholomew2008analysis}.
Biplots and correspondence analysis provide joint displays, but their cross-set interpretation is based on inner products, projections, or standardized associations rather than row--column Euclidean distances \citep{gabriel1971biplot,gabriel2002goodness}.
Neighborhood-embedding methods primarily preserve local proximity
rather than global Euclidean geometry \citep{debodt2025low,vandermaaten2008visualizing,
mcinnes2018umap}.
    \item \textbf{Modeling scope.} 
While model-based approaches can yield valid joint geometric interpretations, current methods remain limited to specific observation structures and response types. For instance, multidimensional scaling (MDS), multidimensional unfolding (MDU), and distance-based latent space models are typically tailored to narrow settings \citep{borg2005modern, chen2021unfolding, hoff2002latent, handcock2007model}, whereas more flexible generalized factor models do not induce a common Euclidean metric geometry \citep{skrondal2004generalized, bartholomew2008analysis}. Consequently, a unified distance-based framework that handles both unipartite and bipartite relational data across continuous, binary, ordinal, and count-valued responses is currently lacking.

    \item \textbf{Computational scalability.}
    Classical MDS admits a direct eigendecomposition-based solution
    \citep{torgerson1952multidimensional,borg2005modern}. In contrast, metric and nonmetric MDS, classical and probabilistic MDU, and likelihood-based latent space models commonly estimate one embedding vector per object through a high-dimensional nonconvex optimization \citep{kruskal1964multidimensional,busing2005avoiding,
    hoff2002latent,chen2021unfolding}.
    The dimension of such optimization problems grows with the number of objects, making these methods unsuitable for large data sets.
    
    \item \textbf{Uniform configuration recovery theory.} 
    By uniform configuration recovery, we mean simultaneous control of the estimation errors of all individual embedding vectors, rather than only an average error over the entire configuration.
    Such guarantees have been studied in selected unipartite settings \citep{zhang2016distance, li2020central, little2023analysis}, whereas model-based theory for bipartite Euclidean unfolding---where only cross-set relations are observed---has primarily concerned average configuration recovery under more specific response and identifiability assumptions \citep{schonemann1970metric, chen2021unfolding}. A unified uniform-recovery theory covering both observation structures is lacking.
    
\end{enumerate}

To address these limitations, we develop a unified distance-based formulation for learning latent Euclidean configurations from both unipartite and bipartite relational data. We assume that the mean of each observation is linked, through a prespecified strictly increasing link function, to a drift parameter minus the squared Euclidean distance between the corresponding embedding vectors.
This formulation accommodates continuous, binary, ordinal, and count-valued
relations while retaining a common proximity-based interpretation. 

We propose a novel estimation procedure that uses a common signal-denoising stage under both relational settings, while adapting the geometric recovery step to their different observation structures. 
By separating signal estimation from geometric recovery, the procedure avoids joint optimization over all latent positions, and the recovery stage can be combined with any sufficiently accurate signal estimator.
We first estimate the signal matrix using a global spectral procedure; for models with nonlinear link functions, we further apply one-step row- and column-wise estimating-equation refinement to obtain uniform
entrywise accuracy.
Given the denoised signal matrix, the unipartite configuration is recovered by the classical MDS construction based on double centering and eigendecomposition \citep{torgerson1952multidimensional,gower1966some}.
For bipartite data, the structurally missing within-set blocks prevent this construction from recovering the joint configuration: double centering identifies only the cross-product of the two centered sets of embedding vectors.
To tackle this issue, we profile out the high-dimensional singular spaces and formulate this remaining ambiguity as an alignment problem, which is solved by an optimization whose dimension depends only on the latent dimension, rather than on the numbers of row and column objects.

Theoretically, we provide convergence guarantees linking the estimated signal matrix to the recovered Euclidean configuration. 
For bipartite data, we establish general identifiability conditions combining a quantitative measure of local
identification strength with global separation. The resulting error bounds explicitly adapt to the local identification strength and are more general than the anchor-point conditions imposed in previous work \citep{chen2021unfolding}.
Based on these identifiability conditions, our results further quantify how uniform accuracy of the estimated signal matrix propagates to simultaneous recovery of all embedding vectors. 
Taken together, these results establish a unified signal-to-configuration recovery theory for both unipartite and bipartite relational data.

\subsection{Related Literature}

Multidimensional scaling (MDS) and multidimensional unfolding (MDU) are most closely related to our work in terms of the target parameter, as both aim to recover a low-dimensional Euclidean configuration from relational or proximity data
\citep{torgerson1952multidimensional,schonemann1970metric,borg2005modern}.
Classical and modern MDS/MDU methods recover such configurations through spectral or optimization-based procedures
\citep{borg2005modern,deleeuw2009smacof}.
Statistical formulations of unfolding further treat the latent coordinates as model parameters and establish identifiability and configuration recovery guarantees under probabilistic response models
\citep{chen2021unfolding}.
Our work shares the goal of recovering an interpretable Euclidean geometry, but considers a common mean formulation for both unipartite and bipartite data, separates signal estimation from geometric recovery, and develops uniform configuration recovery guarantees.

Distance-based latent space models provide another closely related framework for model-based geometric representation of network data, in which each node is assigned a latent position and edge probabilities depend on distances between latent positions \citep{hoff2002latent,handcock2007model}. Estimation has traditionally relied on likelihood or Bayesian procedures, while more scalable methods based on nuclear-norm regularization and projected gradient algorithms have been developed for broad classes of latent space models \citep{ma2020universal}. Recent work has also established uniform consistency and asymptotic distribution theory for latent-position estimators under a range of edge types and link functions \citep{li2025statistical}. These developments primarily concern unipartite networks, whereas our framework treats unipartite and bipartite observation structures jointly and, in the latter case, addresses configuration recovery when only cross-set relations are observed.

Our signal-estimation problem is also connected to the literature on high-dimensional factor and generalized latent factor models, which represents a large data matrix through low-dimensional factors and loadings and targets the resulting bilinear low-rank structure
\citep{bai2003inferential,fan2016projected,chen2020structured}.
Principal-component, projected-PCA, maximum-likelihood, and related scalable methods have been developed for estimating such latent structures in high dimensions
\citep{fan2016projected,wang2022maximum,kidzinski2022generalized}.
This literature provides consistency, convergence rates, inferential theory, and results on factor identifiability under various high-dimensional regimes.
In our setting, however, the low-rank factorization is primarily an intermediate device for estimating the signal matrix: the final target is a common Euclidean configuration whose row and column coordinates have the same metric interpretation, requiring an additional geometric recovery step beyond factor estimation.

Finally, our signal-estimation procedure is related to a growing
literature on spectral estimation and refinement for high-dimensional
latent-variable models. 
SVD-based methods exploit low-rank structure in the population signal to construct computationally efficient estimators or initializers, for example in high-dimensional item factor analysis, mixed-membership models, and deep latent-variable models
\citep{zhang2020note,chen2024spectral,lee2026deep}.
Related spectral-initialization ideas have also been developed for low-rank
matrix and tensor estimation, with statistical and computational guarantees
\citep{xia2021tensor}.
More recent work combines spectral initialization with model-specific one-step refinement to improve the statistical accuracy of the initial spectral estimator in high-dimensional latent-variable models \citep{chen2024entrywise,lyu2026spectral}.
Our procedure follows this broad spectral-plus-refinement principle for
models with nonlinear link functions, while its output is a uniformly
accurate estimate of a distance-based signal matrix that is subsequently
passed to a separate configuration recovery procedure; in the bipartite
setting, this additionally requires resolving the non-orthogonal alignment
ambiguity left by the cross-product structure.

\subsection{Notation}
We introduce some notation for the rest of the paper.
For a positive integer $N$, let $[N]:=\{1,\ldots,N\}$.
For a matrix $\MM$, let $\ve(\MM)$ denote its column-wise vectorization. Let
$\mathcal S_K:=\{\diag(q_1,\ldots,q_K):q_k\in\{-1,1\}\}$ denote the set of $K\times K$ diagonal sign matrices. For two integers $a,b$, let $a \vee b = \max\{a, b\}$ and $a \wedge b = \min \{a, b\}$. 
We let $\Vert \xx \Vert$ denote the standard Euclidean norm of a vector $\xx=(x_1,\dots,x_n)^\top$ and let $\Vert \xx \Vert_\infty=\max_i\vert x_i \vert$ denote the infinity norm of $\xx$. For a matrix $\mathbf{X}=(x_{ij})_{m \times n}$ and a function $f: \R \to \R$, let $f(\XX) :=(f(x_{ij}))_{m \times n} $. The matrix max-norm is denoted by $\Vert\XX\Vert_{\max} = \max_{i,j} |x_{ij}|$. In general, we define the operator $p$-norm of a matrix by $\Vert\XX\Vert_p = \sup_{\Vert\uu\Vert_p=1} \Vert\XX \uu\Vert_p$. In particular, $\Vert\XX\Vert_1 = \max_j \sum_{i=1}^m |x_{ij}|$; $\Vert\XX\Vert_\infty = \max_i \sum_{j=1}^n |x_{ij}|$; and $\Vert\cdot\Vert$ denotes the spectral norm, or equivalently the matrix $2$-norm $\Vert\cdot\Vert_2$, when there is no ambiguity. Let $\Vert\XX\Vert_F$ and $\Vert\XX\Vert_{2\to\infty}$ denote its Frobenius and two-to-infinity norms, respectively, which are defined as $\Vert\XX\Vert_F=\left(\sum_{i=1}^m\sum_{j=1}^n|x_{ij}|^2\right)^{\frac{1}{2}}$ and $\Vert\XX\Vert_{2\to\infty}=\max_{i}\left(\sum_{j=1}^n x_{ij}^2\right)^{\frac{1}{2}}$. $\|\XX\|_*$ denotes the nuclear norm of $\XX$, which is the sum of all singular values. For any matrix $\A\in\R^{N\times J}$ with singular value decomposition $\A=\sum_{i=1}^{N\wedge J}\sigma_i\uu_i \vv_i^\top$ and $\sigma_1\geq \cdots \geq \sigma_{N\wedge J} \geq 0$, we define $\SVD_K(\A):=\sum_{i=1}^{K}\sigma_i\uu_i\vv_i^\top$. For any symmetric matrix $\MM\in\R^{N\times N}$ with eigendecomposition $\MM = \sum_{i=1}^N \lambda_i\uu_i\uu_i^\top$ and $\lambda_1\geq \cdots\geq \lambda_N$, we define $\Eigen_K(\MM):=\sum_{i=1}^{K}\lambda_i\uu_i\uu_i^\top$. For $a<b$, let $\PPP_{[a,b]}(x)=a\,1_{\{x\le a\}}+x\,1_{\{a<x<b\}}+b\,1_{\{x\ge b\}}$. For a matrix $\MM=(m_{ij})$, let $\PPP_{[a,b]}(\MM)$ denote the matrix whose $(i,j)$-th entry is $\PPP_{[a,b]}(m_{ij})$. For a constant $C>0$, let the projection operator $\PPP_{C,2\to\infty}$ map an $n \times p$ matrix onto the constraint set ${\{\A \in \R^{n \times p}:\|\A\|_{2 \to \infty} \leq C\}}$. For a matrix $\VV = (\vv_1,\cdots,\vv_n)^\top \in \R^{n \times p}$, $\PPP_{C,2\to\infty}(\VV) = (\tilde \vv_1,\cdots,\tilde \vv_n)^\top$, where $\tilde \vv_i = \vv_i$ if $\|\vv_i\| \leq C$ and $\tilde \vv_i = (C/\|\vv_i\|)\vv_i$ otherwise. Additionally, let  $\JJ_n = \II_n - \EE_n\EE_n^\top/n$ denote the centering matrix and we sometimes omit the subscript when the dimension is clear from matrix multiplication. Let $\OOO_K$ denote the set of all $K$-by-$K$ orthogonal matrices. For a square matrix $\MM$, let $\Delta(\MM)$ denote the matrix whose off-diagonal entries are the same as those of $\MM$ while the diagonal entries are set to $0$. Let $B_\aaaa^{(K)}(C)$ denote the ball in $\R^K$ centered at $\aaaa \in \R^K$ with radius $C$. The $\psi_1$-Orlicz norm of a random variable $X$, where $\psi_1(x)=\exp(x)-1$, is defined as $\|X\|_{\psi_1}:=\inf\{t>0:\EEE[\exp(|X|/t)]\le 2\}$. Note that $X$ is sub-exponential if $\|X\|_{\psi_1}<\infty$. Throughout the paper, $c>0$ denotes a generic constant independent of $(N,J)$, whose value may change from line to line. 

The remainder of the paper is organized as follows. Section~\ref{sec:models} introduces
the generalized distance-based latent space models and their connections to existing low-dimensional representation methods. Section~\ref{sec:methods} presents the proposed estimation procedure, including refined spectral denoising and configuration
recovery. Section~\ref{sec:theory} establishes the corresponding theoretical guarantees for distance-signal estimation and configuration recovery. Section~\ref{sec:simulation} presents simulation studies, and Section~\ref{sec:realdata} applies the proposed method to a citation network of statisticians and European Parliament roll-call data.
Section~\ref{sec:discuss} concludes with a discussion, and the appendices contain proofs and additional numerical results.

\section{Generalized Distance-Based Latent Space Models}\label{sec:models} 

To describe the unipartite and bipartite
settings within a common notation, let $\mathcal P_1$ and $\mathcal P_2$
denote the object sets associated with the row and column indices of the
observed relational matrix, respectively, and let
$\mathcal P=\mathcal P_1\cup\mathcal P_2$ denote the global object set. Define $N = |\PPP_1|$ and $J = |\PPP_2|$.
In the unipartite setting, $\mathcal P_1=\mathcal P_2 = [N]$, whereas in the bipartite setting, $\mathcal P_1 = [N]$ and $\mathcal P_2 = N + [J] = \{N+1,\ldots,N+J\}$. 

Each object $u\in\mathcal P$ is associated with an embedding vector
$\ttt_u\in\R^K$. We collect all embedding vectors into a global
configuration $\TTT=(\ttt_u^\top)_{u\in\mathcal P}^\top$. For $g\in\{1,2\}$, let $ \TTT_{\mathcal P_g} = (\ttt_u^\top)_{u\in\mathcal P_g}^\top$ denote the configuration consisting of the objects in
$\mathcal P_g$. 
We use $u$ and $v$ to denote global object indices in $\mathcal P$, whereas $i$ and $j$ denote the local row and column indices of the observed relational matrix $\Y=(y_{ij})$. The correspondence between the local indices $(i,j)$ and the associated global objects depends on
the observation structure and will be specified separately for the unipartite and bipartite settings.

For each observed entry $y_{ij}$, let $d_{ij}$ denote the squared Euclidean distance between the two corresponding global objects, and define $m_{ij}$ as
$$
m_{ij}=\tau-d_{ij}\quad\text{with}\quad
d_{ij} = 
\left\{
\begin{aligned}
&\|\ttt_i - \ttt_j\|^2,\qquad  &&\text{unipartite}, \\
&\|\ttt_i - \ttt_{N+j}\|^2,\qquad &&\text{bipartite},
\end{aligned}
\right.
$$
where $\tau\in\R$ is an unknown drift parameter. We treat the embedding vectors as fixed unknown parameters and assume that the observations 
satisfy an independence structure appropriate to the underlying
relational setting, as specified in
Sections~\ref{subsec:uni model} and~\ref{subsec:bi model}. We assume
\begin{equation}\label{eq:model}
    \EEE(y_{ij})
    =
    f(m_{ij})
    =
    f(\tau-d_{ij}),
\end{equation}
where $f:\R\to\R$ is a prespecified strictly increasing link function. We
refer to $\MM=(m_{ij})$ as the signal matrix.
Under \eqref{eq:model}, a smaller latent distance corresponds to a larger
expected relation. The parameter $\tau$ controls the overall level of
the observed responses. If the observed entries represent
dissimilarities rather than similarities, one may instead model $\EEE(y_{ij})=f(-m_{ij})$. For simplicity, we focus on the formulation in \eqref{eq:model} throughout the paper. 

The mean formulation in \eqref{eq:model} accommodates different response
types through the choice of $f$. For continuous responses, one may take
$f$ to be the identity function and regard $y_{ij}$ as a noisy
observation of $m_{ij}$. For binary responses, one may take $f$ to be a logistic or probit function, so that the probability of observing a positive relation
decreases with the latent distance. Ordinal responses can be handled
analogously through an ordinal response model. For count data, one may
take $f(x)=\exp(x)$, leading to a Poisson mean model with intensity
$\exp(m_{ij})$.

The object of interest is the latent Euclidean configuration $\TTT$, rather than only the signal matrix or fitted mean relations.
Because Euclidean distances are invariant under common shifts and orthogonal transformations, configuration recovery is understood modulo these intrinsic geometric ambiguities.
In our theoretical and simulation analyses, recovery accuracy is measured by the largest row-wise Euclidean discrepancy between the estimated and true configurations after appropriate alignment, corresponding to a two-to-infinity norm loss. Specifically, let $\TTT^*$ be the true latent configuration and $\widehat\TTT$ be any estimator. 
The recovery accuracy is measured by 
$$
    d_c(\widehat\TTT,\TTT^*) = \min_{\QQ\in\OOO_K,\bb\in\R^K}\|\widehat\TTT - \EE_{|\PPP|}\bb^\top - \TTT^*\QQ \|_{2\to\infty}.
$$

For unipartite data, the centered configuration is given by its Gram matrix $\JJ_N\TTT(\JJ_N\TTT)^{\top}$ and can therefore be recovered spectrally once the signal matrix is known. For bipartite data, however, only cross-set relations are observed, so the double-centered signal identifies only a cross-product of the two centered configurations as $\JJ_N\TTT_{\PPP_1}(\JJ_J\TTT_{\PPP_2})^\top$; 
recovering the joint Euclidean configuration therefore requires additional identifiability conditions and an alignment step.
We give a more detailed discussion in Section~\ref{subsec:method_conf_recover}.

\subsection{Unipartite Model}\label{subsec:uni model}

We first consider unipartite relational data among a single set of $N$ objects. Under the notation introduced above, we have $
\mathcal P_1=\mathcal P_2=\mathcal P=[N]$,
so that
$\TTT_{\mathcal P_1}=\TTT_{\mathcal P_2}=\TTT
=(\ttt_1,\ldots,\ttt_N)^\top$.
The relational data matrix is symmetric, denoted by
$\Y=(y_{ij})_{N\times N}$, where $y_{ij}=y_{ji}$ records the observed
relation between objects $i$ and $j$. The diagonal entries correspond
to self-relations and typically do not carry meaningful relational
information. We set $y_{ii}=0$ for notational convenience. 

We assume that $\{y_{ij}\}_{i<j\in[N]}$ are mutually independent. Let $\DD = (d_{ij})_{N\times N}$ with $d_{ij} = \|\ttt_i - \ttt_j\|^2$. Thus, \eqref{eq:model} gives 
\begin{equation}\label{eq:model_uni}
    \EEE(y_{ij}) = f(m_{ij}) = f(\tau - \|\ttt_i-\ttt_j\|^2),\qquad 1\leq i<j\leq N.     
\end{equation}
Therefore, objects that are closer in the latent space tend to have a larger expected relation.

\begin{remark}[Connection to latent space models]
The unipartite model \eqref{eq:model} is closely related to distance-based latent space models for network data \citep{hoff2002latent,handcock2007model}. In the classical binary setting, each node is assigned an embedding vector and the edge probability is modeled as a decreasing function of the latent distance, often through a logistic form such as 
$    
\operatorname{logit}\Pr(y_{ij}=1\mid \ttt_i,\ttt_j)
=
\alpha + x_{ij}^{\top}\beta - \|\ttt_i-\ttt_j\|.
$
The intercept $\alpha$ controls the baseline density of the network, while the distance term captures homophily and transitivity. The model may also include effect of additional covariates.
Our formulation keeps the same distance-based geometric mechanism but relaxes the parametric assumption and only assumes the expectation is linked to the signal through a general $f$ as in \eqref{eq:model}. 
This extension allows continuous, binary, ordinal, and count-valued relations to be handled within the same distance-based framework. 
\end{remark}

\begin{remark}[Connection to classical MDS]
The unipartite model reduces to classical multidimensional scaling once the signal matrix $\mathbf M$ is known \citep{borg2005modern}.   Since $\MM=\tau\EE_N\EE_N^\top-\DD$, double centering removes the drift term and converts the negative squared distance matrix into the centered Gram matrix:
$$
    \frac{1}{2}\JJ_N\MM\JJ_N
    =
    -\frac{1}{2}\JJ_N\DD\JJ_N = (\JJ_N\TTT)(\JJ_N\TTT)^\top\qquad\text{with}~~ \TTT=(\ttt_1,\ldots,\ttt_N)^\top.
$$
Thus, in the generalized and noisy setting, unipartite configuration recovery consists of first denoising $\MM$ and then applying the standard double-centering and eigendecomposition step.
We refer the reader to Section~\ref{subsec:method_conf_recover} for further details.
\end{remark}

\subsection{Bipartite Model}\label{subsec:bi model}
We next consider bipartite relational data consisting of interactions
between $N$ row objects and $J$ column objects. 
In this case, $\mathcal P_1=[N], \mathcal P_2=N+[J]$ and $\mathcal P=[N+J]$.
Accordingly, $
\TTT
=
(\ttt_1,\ldots,\ttt_{N+J})^\top$ with
$\TTT_{\mathcal P_1}
=
(\ttt_1,\ldots,\ttt_N)^\top$ and $\TTT_{\mathcal P_2}
=
(\ttt_{N+1},\ldots,\ttt_{N+J})^\top$.
The observations are represented by an $N\times J$ matrix
$\Y=(y_{ij})_{N\times J}$, where the local entry $y_{ij}$ corresponds to the relation between global objects $i$ and $N+j$. We assume that
$\{y_{ij}\}_{i\in[N],\,j\in[J]}$ are mutually independent. Let
$\DD=(d_{ij})_{N\times J}$, where $d_{ij}=\|\ttt_i-\ttt_{N+j}\|^2$.
Then \eqref{eq:model} becomes
\begin{equation}\label{eq:model_bi}
    \EEE(y_{ij})
    =
    f(m_{ij})
    =
    f\bigl(\tau-\|\ttt_i-\ttt_{N+j}\|^2\bigr),
    \qquad i\in[N],~j\in[J].
\end{equation}





\begin{remark}[Connection to multidimensional unfolding]
The bipartite model is rooted in multidimensional unfolding (MDU), 
where objects from the two index blocks are embedded in a common Euclidean space and the observed responses are represented through their mutual distances \citep{schonemann1970metric,borg2005modern}. Classical MDU methods typically recover the joint configuration by directly optimizing a high-dimensional stress function over all embedding vectors. Such optimization can be computationally
demanding and may suffer from degeneracy or non-identifiability \citep{busing2005avoiding}.
The probabilistic unfolding model of \citet{chen2021unfolding} provides a
statistical formulation for binary response data by linking response probabilities
to latent distances, and establishes identifiability conditions and convergence
rates for the recovered configuration. However, its estimation still relies on a
high-dimensional optimization over the embedding vectors. Here we keep the same
unfolding geometry, but work with the model \eqref{eq:model_bi}
with an additional unknown drift parameter $\tau$. This formulation accommodates different response types through the link function $f$ and separates the overall response level from the relative geometry of the embedding vectors.
\end{remark}

\begin{remark}
The bipartite model is related to generalized linear factor models (GLFM)
\citep{skrondal2004generalized, bartholomew2008analysis}. A typical GLFM specifies
$$
    \EEE(y_{ij}\mid \ttt_i,\aaaa_j)
    =
    g(\mu_j+\ttt_i^\top\aaaa_j),
$$
where the item-specific intercept $\mu_j$ is a main effect that controls the overall
level of item $j$, while the inner product $\ttt_i^\top\aaaa_j$ captures the interaction
between row and column latent factors. Because the main effects and the interaction
terms are freely parameterized, the latent factors $\ttt_i$ and loadings $\aaaa_j$ are generally identifiable only up to nonsingular linear
transformations. This weakens their direct geometric interpretation.
In contrast, our bipartite signal can be rewritten as
$$
m_{ij}
=
\tau-\|\ttt_i-\aaaa_j\|^2
=
\tau-\|\ttt_i\|^2-\|\aaaa_j\|^2+2\ttt_i^\top\aaaa_j,
$$
which also contains a bilinear interaction term. The important difference is that
the main-effect-like terms $-\|\ttt_i\|^2$ and $-\|\aaaa_j\|^2$ are not free parameters:
they are tied to the same embedding vectors that generate the interaction term
$2\ttt_i^\top\aaaa_j$. This coupling turns the low-rank factor structure into a squared-distance structure, giving the embedding vectors a direct proximity-based interpretation. The price is that estimation is more involved than ordinary GLFM
estimation: after recovering the signal matrix, one must further recover a
Euclidean configuration satisfying the squared-distance representation. This motivates
the denoising and configuration recovery procedures developed in the following sections.
\end{remark}

\section{Proposed Methods}\label{sec:methods}

We first introduce the refined spectral denoising procedure for signal matrix estimation and then present the configuration recovery procedures for unipartite and bipartite relational data.

\subsection{Refined Spectral Denoising}\label{sec:denoise}

Note that the same rank bound applies to both data types. In the bipartite case, recall that
$\TTT_{\mathcal P_1} = (\ttt_1,\ldots,\ttt_N)^\top$ and  $\TTT_{\mathcal P_2} = (\ttt_{N+1},\ldots,\ttt_{N+J})^\top,
$
and define
$ \bm d_{\mathcal P_1} = (\|\ttt_1\|^2,\ldots,\|\ttt_N\|^2)^\top$ and $\bm d_{\mathcal P_2}
=
(\|\ttt_{N+1}\|^2,\ldots,\|\ttt_{N+J}\|^2)^\top.
$
We have
\begin{equation}\label{eq:unified_bi_low_rank}
    \MM
    =
    2\TTT_{\mathcal P_1}\TTT_{\mathcal P_2}^\top
    +\tau\EE_N\EE_J^\top
    -\bm d_{\mathcal P_1}\EE_J^\top
    -\EE_N\bm d_{\mathcal P_2}^\top,
    \qquad
    \rank(\MM)\le K+2.
\end{equation}

At the level of the low-rank signal representation, the unipartite formula
is obtained by taking $\mathcal P_1=\mathcal P_2=[N]$, so that $\TTT_{\mathcal P_1}=\TTT_{\mathcal P_2}=\TTT$. Hence, letting
$ \bm d = (\|\ttt_1\|^2,\ldots,\|\ttt_N\|^2)^\top$,
we have
\begin{equation}\label{eq:unified_uni_low_rank}
    \MM
    =
    2\TTT\TTT^\top
    +\tau\EE_N\EE_N^\top
    -\bm d\EE_N^\top
    -\EE_N\bm d^\top,
    \qquad
    \rank(\MM)\le K+2.
\end{equation}

We propose Algorithm~\ref{alg:rsd}, which gives an estimator for the signal matrix.
This algorithm consists of two stages: the spectral denoising stage gives a fast global denoiser by exploiting the rank-$(K+2)$ structure in \eqref{eq:unified_bi_low_rank} and \eqref{eq:unified_uni_low_rank}; 
for nonlinear links, the spectral denoising stage is followed by a
one-step refinement, which uses the rescaled right singular vectors as anchors, updates the row space by solving estimating equations for each row $i$, and then updates the column space.
For unipartite data, the final symmetrization and rank-$(K+2)$ truncation retain the leading signal components while enforcing the one-mode structure.
For the bipartite case, we assume without loss of generality that $N \ge J$.

\begin{algorithm}[!htb]
    \caption{Refined Spectral Denoising}
    \label{alg:rsd}
    \begin{enumerate}
        \item \textbf{Input:} the observation matrix $\YY$, latent dimension $K$, link function $f(\cdot)$, truncation parameters $a_{NJ}$ and $b_{NJ}$, and projection constant $C$. For unipartite data, set $J = N$.

        \item \textbf{Spectral denoising.} Define $\widehat\XX=(\widehat x_{ij})$ by
        $$
            \widehat\XX=
            \begin{cases}
            \SVD_{K+2}(\YY), & \text{if } f(x)=x,\\
            (b_{NJ}-a_{NJ})\SVD_{\widetilde r}(\widetilde\YY)+a_{NJ}\EE_N\EE_J^\top, & \text{otherwise},
            \end{cases}
        $$
        where
        $$
        \begin{aligned}
            &\widetilde \YY = (\widetilde y_{ij}) \text{ with } \widetilde y_{ij}
            =
            \{\PPP_{[a_{NJ},b_{NJ}]}(y_{ij})-a_{NJ}\}/(b_{NJ}-a_{NJ}), \\
            &\widetilde r = \max\{K+2, \max\{k: \sigma_k(\widetilde \YY) \geq 1.01\sqrt{N}\}\}.
        \end{aligned}
        $$
        Let
        $$
        \widehat\MM^{\SVD}
        =
        \begin{cases}
        \PPP_{[-C,C]}(\widehat\XX), & f(x)=x,\\
        f^{-1}\!\left\{\PPP_{[f(-C),f(C)]}(\widehat\XX)\right\},
        & \text{otherwise}.
        \end{cases}
        $$
        If $f(x) = x$, then return $\widehat\MM^{\mathrm{SVD}}$.
Otherwise, perform the one-step refinement below.

        \item \textbf{One-step refinement for nonlinear link functions.}        
        Compute the rank-$(K+2)$ SVD of $\widehat\MM^{\SVD}$ and let $\widehat\VVV\in\R^{J\times (K+2)}$ be its right singular matrix. Set
        $$
            \widehat\AAA=(\widehat\aaa_1,\ldots,\widehat\aaa_J)^\top
            =
            \PPP_{C,2\to\infty}(\sqrt{J}\widehat\VVV).
        $$
        For each row $i$, obtain $\widetilde\bbb_i$ by solving
            \begin{equation}\label{eq:solve_row}
                \sum_{j\in\mathcal J_i}
                \{y_{ij}-f(\widehat\aaa_j^\top\bbb_i)\}\widehat\aaa_j
                =
                \0_{K+2},
            \end{equation}
        where $\mathcal J_i=[J]$ for bipartite data and $\mathcal J_i=[N]\setminus\{i\}$ for unipartite data.
        Then, for each column $j$, obtain $\widetilde\aaa_j$ by solving
            \begin{equation}\label{eq:solve_col}
                \sum_{i\in\mathcal I_j}
                \{y_{ij}-f(\aaa_j^\top\widetilde\bbb_i)\}\widetilde\bbb_i
                =
                \0_{K+2},
            \end{equation}
         where $\mathcal I_j=[N]$ for bipartite data and $\mathcal I_j=[N]\setminus\{j\}$ for unipartite data.

        \item \textbf{Output:} For bipartite data, return $\widehat\MM^{\text{RSD}}=\widetilde\BBB\widetilde\AAA^\top$, where $\widetilde\BBB=(\widetilde\bbb_1,\ldots,\widetilde\bbb_N)^\top$ and $\widetilde\AAA=(\widetilde\aaa_1,\ldots,\widetilde\aaa_J)^\top$. For unipartite data, return
        $$
            \widehat\MM^{\text{RSD}}
            =
            \SVD_{K+2}\left\{
            \frac{\widetilde\BBB\widetilde\AAA^\top+\widetilde\AAA\widetilde\BBB^\top}{2}
            \right\}.
        $$
    \end{enumerate}
\end{algorithm}

\begin{remark}
We discuss the spectral denoising stage.
If $f(\cdot)$ is the identity function, then we directly apply singular value decomposition to $\YY$ and select the first $K+2$ signals to obtain $\widehat\XX$, which is formally defined in Algorithm~\ref{alg:rsd}.
Note that for a general linear function $f(x)=kx+b$ with $k>0$, for any observed pair corresponding to global objects $u$ and $v$,
$$
f\bigl(\tau-\|\ttt_u-\ttt_v\|^2\bigr)
=
(k\tau+b)-\|\sqrt{k}\ttt_u-\sqrt{k}\ttt_v\|^2,
$$
which reduces to the case of the identity 
link function where the drift becomes $k\tau+b$ and the embedding vectors are rescaled by $\sqrt{k}$.
For nonlinear $f(\cdot)$, instead of directly performing singular value decomposition to $\YY$, we first truncate all $y_{ij}$ to $[a_{NJ},b_{NJ}]$ and rescale them to $[0,1]$, after which we apply singular value thresholding \citep{chatterjee2015matrix} to the transformed matrix with the threshold $1.01\sqrt{N}$. The truncation levels $a_{NJ}$ and $b_{NJ}$ may be chosen according to the response space and tail behavior of $y_{ij}$. One such choice, together with the resulting convergence rates, is given in Corollary~\ref{cor:support_adaptive_truncation}. The spectral estimate is then transformed back to the original response scale to obtain $\widehat\XX$. Finally, for a nonlinear link, the estimate is first projected onto $f([-C,C])$ and then transformed by $f^{-1}$.
\end{remark}

\begin{remark}
We discuss the refinement stage, which is required if $f$ is a nonlinear function.
Let $\MM^*$ denote the true signal matrix, and write its singular value decomposition as
$\MM^*=\UUU^*\DDDD^*\VVV^{*\top}$. Further let $\AAA^* = \sqrt{J}\,\VVV^*\widehat \PP$ and $\BBB^* = J^{-1/2}\,\UUU^*\DDDD^*\widehat \PP$, where $\widehat \PP \in \argmin_{\PP\in\OOO_{K+2}} \|\widehat{\VVV}-\VVV^*\PP\|_F$.
If we replace $y_{ij}$ by its expectation $f(m_{ij}^*)$ in \eqref{eq:solve_row}, then it becomes
$$
\sum_{j\in\mathcal J_i}\left\{f(m_{ij}^*)-f\left(\widehat{\aaa}_j^\top\widetilde\bbb_i\right)\right\}\widehat{\aaa}_j
=
\0_{K+2}.
$$
Therefore, $\widetilde\bbb_i\approx \bbb_i^*$ as long as $\widehat\AAA \approx \AAA^*$, which is guaranteed by the spectral denoising stage. Similar analysis applies to \eqref{eq:solve_col}.
Both \eqref{eq:solve_row} and \eqref{eq:solve_col} involve only $K+2$ variables which could be solved efficiently, and there exists a unique solution with probability approaching 1, as will be shown in Section~\ref{subsec:theory1}.
Moreover, for each $i\in[N]$, \eqref{eq:solve_row} does not depend on other $\bbb_j$ for $j\neq i$. Therefore, the procedure could be accelerated by parallel computing, and the same applies to equation \eqref{eq:solve_col}.
\end{remark}

\subsection{Configuration Recovery}\label{subsec:method_conf_recover}

In this section, we turn to configuration recovery based on an estimate of the signal matrix $\widehat\MM$.
Although $\widehat\MM$ can be taken as the signal estimator returned by Algorithm~\ref{alg:rsd}, the recovery procedures apply more generally to any sufficiently accurate estimator of $\MM^*$. Under the regularity and identifiability conditions specified in Section~\ref{subsec:theory_conf_recover}, uniform consistency for signal estimation translates into uniform consistency for configuration recovery.

\subsubsection{Unipartite Configuration Recovery}
We first consider configuration recovery for the unipartite case, where the obtained $\widehat\MM$ is supposed to be symmetric.
Let $\TTT^*=(\ttt_1^*,\ldots,\ttt_N^*)^\top$ denote the true latent configuration. Note that $ \frac12\JJ_N\MM^*\JJ_N = (\JJ_N\TTT^*)(\JJ_N\TTT^*)^\top$. To estimate $\TTT^*$, let
$$
\Eigen_K(\JJ_N\widehat\MM\JJ_N/2)
=
\sum_{k=1}^K\widehat\lambda_k\widehat\uu_k\widehat\uu_k^\top.
$$
Then,
$$
\widehat \TTT
=
((\widehat\lambda_1)_+^{1/2}\widehat\uu_1,\ldots,
(\widehat\lambda_K)_+^{1/2}\widehat\uu_K),
$$
where $(\widehat\lambda_k)_+ = \max\{0,\widehat\lambda_k\}$ for $k\in[K]$.

\subsubsection{Bipartite Configuration Recovery}
To present the proposed method for bipartite configuration recovery, we first introduce some generic notations.
Let $\Z = \bigl(\sqrt N\,\UU^\top,\sqrt J\,\WW^\top\bigr)^\top \in\R^{(N+J)\times K}$,
where $\UU\in\R^{N\times K}$ and $\WW\in\R^{J\times K}$.
Given an invertible matrix $\FF\in\R^{K\times K}$ and a vector $\bb\in\R^K$, define
$\rrr = (\tau, \ve(\FF)^\top, \bb^\top)^\top \in \R^{K(K+1)+1}$, where $\tau$ is the intercept parameter in \eqref{eq:model}.
Given $\Z$ and $\rrr$, further define the induced configurations as
\begin{equation}\label{eq:unified_TA}
\TTT_{\mathcal P_1}(\rrr;\Z)
=
\sqrt{N}\UU\FF,
\qquad
\TTT_{\mathcal P_2}(\rrr;\Z)
=
\sqrt{J}\WW(\FF^{-1})^\top
+
\EE_J\bb^\top.
\end{equation}
The corresponding global configuration is
$$
\TTT(\rrr;\Z)
=
\begin{pmatrix}
\TTT_{\mathcal P_1}(\rrr;\Z)\\
\TTT_{\mathcal P_2}(\rrr;\Z)
\end{pmatrix}.
$$
We also define the induced signal matrix
$\MM(\rrr;\Z)=(m_{ij}(\rrr;\Z))_{N\times J}$ with
\begin{equation}\label{eq:unified_M}
m_{ij}(\rrr;\Z)
=
\tau-\|\ttt_i(\rrr;\Z)-\ttt_{N+j}(\rrr;\Z)\|^2,
\end{equation}
where $\ttt_i(\rrr;\Z)$ and $\ttt_{N+j}(\rrr;\Z)$ are the embedding vectors of global objects $i$ and $N+j$ in $\TTT(\rrr;\Z)$, respectively.

The key idea is to profile out the high-dimensional singular spaces, which
are estimated directly from the double-centered signal matrix. Conditional on
these spaces, summarized by $\Z$, configuration recovery only requires fixing a
low-dimensional transformation: an invertible matrix $\FF$ and a shift vector
$\bb$, together with the drift parameter $\tau$. The following flow chart
summarizes how these remaining parameters induce a configuration and a signal matrix, which is then aligned with $\widehat\MM$.







\begin{center}
\begin{adjustbox}{max width=\linewidth}
\begin{tikzpicture}[
    >=stealth,
    stage/.style={
        draw=black!70,
        fill=black!2,
        rounded corners=2pt,
        align=center,
        text width=0.22\linewidth,
        inner xsep=5pt,
        inner ysep=5pt,
        font=\footnotesize
    },
    flow/.style={
        ->,
        thick,
        draw=black!70
    }
]

\node[stage] (parameter) {
    \parbox[c][1.55cm][c]{\linewidth}{
        \centering
        \textbf{Parameters}\\[4pt]
        $\rrr=(\tau,\ve(\FF)^\top,\bb^\top)^\top$
    }
};

\node[
    stage,
    right=2mm of parameter
] (configuration) {
    \parbox[c][1.55cm][c]{\linewidth}{
        \centering
        {\scriptsize\bfseries\mbox{Induced configuration}}\\[7pt]
        {\scriptsize
        $\TTT_{\mathcal P_1}(\rrr;\Z),
        ~~
        \TTT_{\mathcal P_2}(\rrr;\Z)$
        }
    }
};

\node[
    stage,
    right=2mm of configuration
] (signal) {
    \parbox[c][1.55cm][c]{\linewidth}{
        \centering
        {\scriptsize\bfseries\mbox{Induced signal}}\\[4pt]
        $\MM(\rrr;\Z)$
    }
};

\node[
    stage,
    right=2mm of signal
] (alignment) {
    \parbox[c][1.55cm][c]{\linewidth}{
        \centering
        \textbf{Align with $\widehat\MM$}\\[4pt]
        {\scriptsize
        $\displaystyle
        \min_{\rrr}
        \|\MM(\rrr;\Z)-\widehat\MM\|_F^2$
        }
    }
};

\draw[flow] (parameter) -- (configuration);
\draw[flow] (configuration) -- (signal);
\draw[flow] (signal) -- (alignment);

\end{tikzpicture}
\end{adjustbox}
\end{center}

We first illustrate the idea for the case where $\MM^*$ is known. Since the distances between embedding vectors are invariant under shifting, we assume
$\TTT_{\mathcal P_1}^{*\top}\EE_N = {\bf0}$
without loss of generality. By double-centering $\MM^*$, we obtain
$
\frac{1}{2}\JJ_N\MM^*\JJ_J
=
\TTT_{\mathcal P_1}^*
\bigl(\JJ_J\TTT_{\mathcal P_2}^*\bigr)^\top.
$
Let
$
\frac{1}{2}\JJ_N\MM^*\JJ_J
=
\UU^*\SG^*\VV^{*\top}
$
be the singular value decomposition.
To properly scale the singular vectors, we define
$\WW^*=\frac{1}{\sqrt{NJ}}\VV^*\SG^*$ and construct
$
\Z^*
=
(\sqrt{N}\UU^{*\top},\,\sqrt{J}\WW^{*\top})^\top.
$
Notice that
$$
\TTT_{\mathcal P_1}^*
\bigl(\JJ_J\TTT_{\mathcal P_2}^*\bigr)^\top
=
\frac{1}{2}\JJ_N\MM^*\JJ_J
=
(\sqrt{N}\UU^*)(\sqrt{J}\WW^*)^\top,
$$
which guarantees the existence of a vector
$\rrr_0=(\tau_0,\ve(\FF_0)^\top,\bb_0^\top)^\top$
such that
$$
\TTT_{\mathcal P_1}^*
=
\TTT_{\mathcal P_1}(\rrr_0;\Z^*),
\qquad
\TTT_{\mathcal P_2}^*
=
\TTT_{\mathcal P_2}(\rrr_0;\Z^*).
$$
Equivalently,
$
\TTT^*=\TTT(\rrr_0;\Z^*).
$
Consequently, we have
$\MM^*=\MM(\rrr_0;\Z^*)$.
If $\MM^*$ is known, the true parameter $\rrr_0$ could be recovered by minimizing the Frobenius norm $\|\MM(\rrr;\Z^*)-\MM^*\|_F^2$.

For parameter estimation, the true $\MM^*$ is unknown, and we rely on the perturbed signal matrix $\widehat\MM$.
For any $\rrr,\Z$ and $\MM$, we measure the reconstruction error based on $\rrr,\Z$ from $\MM$ by the following objective function
$$
l(\rrr;\Z,\MM)
=
\|\MM(\rrr;\Z)-\MM\|_F^2,
$$
where $\MM(\rrr;\Z)$ is the signal matrix defined in \eqref{eq:unified_M}.
The pair $(\rrr_0,\Z^*)$ exactly reproduces the true drifted distance matrix $\MM^*$, yielding $l(\rrr_0;\Z^*,\MM^*)=0$.
We proceed by computing the rank-$K$ singular value decomposition as $\SVD_K\left(\JJ_N\widehat\MM\JJ_J/2\right) = \widehat\UU\widehat\SG\widehat\VV^\top$,
where $\widehat\UU\in\R^{N\times K}$, $\widehat\VV\in\R^{J\times K}$, and
$\widehat\SG=\diag(\widehat\sigma_1,\ldots,\widehat\sigma_K)$.
Similar to the case of $\MM^*$, we define
$\widehat\WW=\frac{1}{\sqrt{NJ}}\widehat\VV\widehat\SG$ and construct the matrix $\widehat\Z = (\sqrt{N}\widehat\UU^\top,\,\sqrt{J}\widehat\WW^\top)^\top$.
To estimate $\rrr_0$, we solve the following constrained optimization problem:
\begin{equation}\label{eq:opti}
\begin{aligned}
&\widehat\rrr \in 
\argmin_{\rrr\in\GGamma}
l(\rrr;\widehat\Z,\widehat\MM)
\quad\text{with}\quad
l(\rrr;\widehat\Z,\widehat\MM)
=
\|\MM(\rrr;\widehat\Z)-\widehat\MM\|_F^2
\quad\text{and}
\\
&
\GGamma
=
\Bigl\{
\rrr
=
(\tau,\ve(\FF)^\top,\bb^\top)^\top
\in\R^{K(K+1)+1}:
\|\rrr\|\le C,\ 
\det(\FF)\neq0,
\\
&
\phantom{\GGamma=\Bigl\{}
\text{and}\
\FF^\top\FF
\text{ is diagonal with non-increasing entries}
\Bigr\}.
\end{aligned}
\end{equation}
for a sufficiently large fixed constant $C>0$. 

\begin{remark}
The objective function $l(\rrr;\widehat\Z,\widehat\MM)$ is a plug-in version of the oracle objective $l(\rrr;\Z^*,\MM^*)$, obtained by replacing the unknown quantities $\Z^*$ and $\MM^*$ with their
estimators $\widehat\Z$ and $\widehat\MM$, respectively.
This suggests that $\widehat\rrr$ should be close to $\rrr_0$ if $\widehat\MM\approx\MM^*$ and $\widehat\Z\approx\Z^*$. The parameter space $\GGamma$ is bounded by the constraint $\|\rrr\|\le C$, while the condition $\det(\FF)\neq0$ excludes singular configurations. Although $\GGamma$ is not compact due to the constraint $\det(\FF)\neq0$, the minimum of $l(\rrr;\widehat\Z,\widehat\MM)$ over $\GGamma$ is attained when $\|\widehat\MM-\MM^*\|_{\max}$ is sufficiently small. A formal statement is provided in Theorem~\ref{thm:bi_conf} below.
The diagonal constraint on $\FF^\top\FF$ removes the rotational ambiguity of the configuration and ensures identifiability up to column-wise sign flipping. Indeed, it follows from \eqref{eq:unified_TA} that constraining $\FF$ fixes the orientation of $\TTT_{\mathcal P_1}(\rrr;\Z)$ and hence that of the global configuration $\TTT(\rrr;\Z)$.
\end{remark}

It is worth mentioning that the optimization problem \eqref{eq:opti} involves only $K(K+1)+1$ parameters, so it is computationally fast. This is in contrast to the SMACOF method \citep{deleeuw2009smacof} and likelihood-based optimization \citep{chen2021unfolding}, which involve at least $(N+J)K$ parameters. We summarize the complete estimation procedure in Algorithm~\ref{alg:recover}.

\begin{algorithm}[!htb]
\caption{Bipartite configuration recovery}
\label{alg:recover}
\begin{enumerate}
\item \textbf{Input:} Perturbed signal matrix $\widehat \MM$.

\item Compute
$\SVD_K(\JJ_N\widehat\MM\JJ_J/2)
=
\widehat\UU\widehat\SG\widehat\VV^\top$,
where the signs of each singular-vector pair are chosen so that the first nonzero entry of each column of $\widehat\UU$ is positive. Define
$\widehat\WW
=
\widehat\VV\widehat\SG/\sqrt{NJ}$
and
$\widehat\Z
=
(\sqrt{N}\widehat\UU^\top,\,
\sqrt{J}\widehat\WW^\top)^\top$.

\item Solve
$\widehat\rrr
\in
\argmin_{\rrr\in\GGamma}
l(\rrr;\widehat\Z,\widehat\MM)$,
where $l(\rrr;\widehat\Z,\widehat\MM)$ and $\GGamma$ are given in
\eqref{eq:opti}.

\item \textbf{Output:}
$$
\widehat\TTT_{\mathcal P_1}
=
\TTT_{\mathcal P_1}(\widehat\rrr;\widehat\Z)
=
\sqrt{N}\widehat\UU\widehat\FF,
$$
$$
\widehat\TTT_{\mathcal P_2}
=
\TTT_{\mathcal P_2}(\widehat\rrr;\widehat\Z)
=
\sqrt{J}\widehat\WW(\widehat\FF^{-1})^\top
+
\EE_J\widehat\bb^\top,
$$
and
$$
\widehat\TTT
=
\begin{pmatrix}
\widehat\TTT_{\mathcal P_1}\\
\widehat\TTT_{\mathcal P_2}
\end{pmatrix}.
$$
\end{enumerate}
\end{algorithm}

\section{Theoretical Guarantees}\label{sec:theory}
Throughout this section, we take the latent dimension $K$ to be fixed, and suppose $J\leq N = J^{O(1)}$. 
For unipartite data, we set $J=N$.

\subsection{Signal Matrix Estimation}\label{subsec:theory1}

We study the asymptotic properties of the signal estimators produced by Algorithm~\ref{alg:rsd}. The purpose of this section is to show that Algorithm~\ref{alg:rsd} produces an entrywise accurate estimate of the signal matrix $\MM^*$. This entrywise control will serve as the input condition for the configuration recovery analysis in Section~\ref{subsec:theory_conf_recover}.

Let $\TTT^*=(\ttt_u^*)_{u\in\mathcal P}^\top$ be the true global latent configuration, $\MM^*$ be the true signal matrix, and $\tau_0$ be the true value of the drift parameter $\tau$. For unipartite data, $\mathcal P=[N]$ and $\TTT^*=(\ttt_1^*,\ldots,\ttt_N^*)^\top$. For bipartite data, $\mathcal P=[N+J]$, with configurations
$$
\TTT_{\mathcal P_1}^*
=
(\ttt_1^*,\ldots,\ttt_N^*)^\top,
\qquad
\TTT_{\mathcal P_2}^*
=
(\ttt_{N+1}^*,\ldots,\ttt_{N+J}^*)^\top.
$$
Let $\MM^*=\UUU^*\DDDD^*\VVV^{*\top}$ be a rank-$(K+2)$ singular value decomposition of $\MM^*$. For the spectral estimator $\widehat\MM^{\mathrm{SVD}}$, let
$\widehat{\UUU}\widehat\DDDD\widehat{\VVV}^{\top}:=\SVD_{K+2}(\widehat\MM^{\mathrm{SVD}})$
denote its rank-$(K+2)$ truncated singular value decomposition.

To derive the theoretical results, we impose several regularity conditions on the boundedness of the configuration and signal matrix, the tail behavior of the random noise, the smoothness and monotonicity of the link function, and the spectral structure of the signal matrix.

\begin{assumption}\label{assumption:algo1}
Suppose the following conditions hold.

\assumptionitem \label{assumption:algo1_bounded}
$\max_{u\in\mathcal P}\|\ttt_u^*\|\le C$ and
$\|\MM^*\|_{\max}\le C$ for some constant $C>0$.

\assumptionitem \label{assumption:algo1_subexponential}
$\max_{i\in[N],\,j\in[J]}
\|y_{ij}-f(m_{ij}^*)\|_{\psi_1}
=
O(1)$ as $N,J\to\infty$.

\assumptionitem \label{assumption:algo2_link}
The link function $f$ is twice continuously differentiable and $f'(x)>0$ for $x\in\R$.

\assumptionitem \label{assumption:algo2_incoherence}
The singular vectors of $\MM^*$ satisfy
$\|\UUU^*\|_{2\to\infty}\lesssim N^{-1/2}$ and
$\|\VVV^*\|_{2\to\infty}\lesssim J^{-1/2}$.

\assumptionitem \label{assumption:algo2_spikes}
The singular values of $\MM^*$ satisfy
$\sigma_{K+2}(\MM^*)\gtrsim\sqrt{NJ}$.
\end{assumption}

\begin{remark}
Assumption~\ref{assumption:algo1_bounded} requires that the global latent configuration lies in a compact subset of $\R^K$, which is a standard condition in the literature of low-rank matrix estimation and multidimensional unfolding \citep{chen2021unfolding, zhang2020note}.
Assumption~\ref{assumption:algo1_subexponential} controls the tail behavior of the random noises, which holds for many model types in \eqref{eq:model}, including linear, logistic, Probit, and Poisson models.
Assumption~\ref{assumption:algo2_link} imposes smoothness and monotonicity conditions on the link function. Since $f'$ is continuous and strictly positive, it is bounded away from zero on any fixed compact interval. In particular, under Assumption~\ref{assumption:algo1_bounded},
$
\sup_{y\in[f(-C),f(C)]}|(f^{-1})'(y)|
=
\sup_{|m|\le C}\frac{1}{f'(m)}
<\infty.
$
Moreover, the continuity of $f'$ and $f''$ provides the derivative bounds
required in the Taylor expansions of the refinement analysis.
Assumptions~\ref{assumption:algo2_incoherence} and~\ref{assumption:algo2_spikes} are standard incoherence and signal strength conditions that prevent overly concentrated singular vectors and degenerate singular values.
\end{remark}

\subsubsection{Error Bounds for Spectral Denoising}

We next establish error bounds for the signal matrix estimator $\widehat\MM^{\mathrm{SVD}}$ obtained from the spectral denoising stage of Algorithm~\ref{alg:rsd}.

\begin{theorem}\label{thm:converge in prob}
For $\widehat\MM^{\text{SVD}}$, the following error bounds hold.
\begin{enumerate}
\item \textbf{(Identity link)}
Suppose Assumption~\ref{assumption:algo1_bounded} and \ref{assumption:algo1_subexponential} hold. If $f(x)=x$, then
$$
\frac{1}{NJ}
\|\widehat\MM^{\mathrm{SVD}}-\MM^*\|_F^2
=
O_p(J^{-1}).
$$
If, in addition, Assumptions~\ref{assumption:algo2_incoherence}
and~\ref{assumption:algo2_spikes} hold, then
\[
\|\widehat\MM^{\mathrm{SVD}}-\MM^*\|_{\max}
=
O_p\left(
J^{-\frac{1}{2}}\sqrt{\log(NJ)}\right).
\]
\item \textbf{(Nonlinear link)}
Suppose Assumption~\ref{assumption:algo1_bounded} and \ref{assumption:algo2_link} hold. 
Let $a_{NJ}<b_{NJ}$ be deterministic truncation levels and define $R_{NJ}=b_{NJ}-a_{NJ}$. 
Further suppose
$R_{NJ}+r_{NJ}=o(\sqrt J)$, where
$$
r_{NJ} =
\begin{cases}
\displaystyle
\max_{i\in[N],\,j\in[J]}
\left|
\EEE\left\{
\PPP_{[a_{NJ},b_{NJ}]}(y_{ij})
-
f(m_{ij}^*)
\right\}
\right|,
& \text{bipartite},
\\[3mm]
\displaystyle
\max_{i\neq j}
\left|
\EEE\left\{
\PPP_{[a_{NJ},b_{NJ}]}(y_{ij})
-
f(m_{ij}^*)
\right\}
\right|,
& \text{unipartite}.
\end{cases}
$$
For bipartite data, the normalized Frobenius error satisfies
$$
\frac{1}{NJ}
\|\widehat\MM^{\mathrm{SVD}}-\MM^*\|_F^2
=
O_p\left(
R_{NJ}
\left\{
\frac{(R_{NJ}+r_{NJ})^2}{J}
\right\}^{\frac{1}{K+2}}
+
r_{NJ}(R_{NJ}+r_{NJ})
\right).
$$
The same bound holds for unipartite data, provided that
$f(\tau_0)\in[a_{NN},b_{NN}]$.
\end{enumerate}
\end{theorem}

\begin{remark}\label{rem:bound}
For the identity link, the rate $O_p(J^{-1})$ agrees with the minimax order for fixed-rank matrix estimation under additive noise and full observation \citep{agarwal2012noisy}.
Under the additional incoherence and signal strength conditions in
Assumption~\ref{assumption:algo1}, the same spectral estimator is also consistent in max-norm.
The logarithmic factor reflects the cost of controlling all entries simultaneously, relative to the $J^{-1/2}$ scale, 
which is consistent with existing literature on entrywise perturbation for low-rank spectral estimators \citep{abbe2020entrywise}.
Consequently, for the identity link, the spectral estimator itself already provides the uniform signal accuracy required for subsequent configuration recovery, and no refinement step is needed.
For nonlinear links, Theorem~\ref{thm:converge in prob} provides a
rate for the proposed spectral estimator under a mean
formulation without requiring a fully specified likelihood.
Under stronger distributional assumptions, results for related
generalized latent factor models suggest the parametric benchmark
$O_p(J^{-1})$ under the normalized Frobenius loss for fixed $K$ and
$N\ge J$ \citep{chen2020structured}.
When a full likelihood is available, the first-stage signal estimator may instead be constructed using, for example, a low-rank constrained maximum likelihood estimator or a nuclear-norm penalized likelihood estimator. The subsequent refinement, when applicable, and configuration recovery procedures are modular and can be combined with such alternative signal estimators provided that the required initial estimation accuracy is satisfied.
\end{remark}

\begin{corollary}
\label{cor:support_adaptive_truncation}
Suppose Assumptions~\ref{assumption:algo1_bounded},
\ref{assumption:algo1_subexponential}, and
\ref{assumption:algo2_link} hold, and let $f$ be nonlinear.
Let $\mathcal Y_{NJ}$ denote the closure of the union of the supports
of all response variables $y_{ij}$ under consideration, and suppose that $\inf\mathcal Y_{NJ}<\sup\mathcal Y_{NJ}$ for all sufficiently large $N$ and $J$.
Then there exists a fixed constant $\kappa>0$ such that, taking
$a_{NJ}=\max\{\inf\mathcal Y_{NJ},-\kappa\log J\}$ and $b_{NJ}=
\min\{\sup\mathcal Y_{NJ},\kappa\log J\}$, we have
\[
\frac{1}{NJ}
\left\|
\widehat\MM^{\mathrm{SVD}}-\MM^*
\right\|_F^2
=
O_p\left(
J^{-\frac{1}{K+2}}
(\log J)^{\frac{K+4}{K+2}}
\right).
\]
If, in addition, the response supports are uniformly bounded, that is, $\sup_{N,J}\sup_{y\in\mathcal Y_{NJ}}|y|<\infty$, then, for all sufficiently large $J$,
$a_{NJ}=\inf\mathcal Y_{NJ}$ and
$b_{NJ}=\sup\mathcal Y_{NJ}$.
Consequently,
\[
\frac{1}{NJ}
\left\|
\widehat\MM^{\mathrm{SVD}}-\MM^*
\right\|_F^2
=
O_p\left(
J^{-\frac{1}{K+2}}
\right).
\]
The same conclusions hold for unipartite data with $J=N$, provided that
$f(\tau_0)\in[a_{NN},b_{NN}]$ for all sufficiently large $N$.
\end{corollary}

\begin{remark}\label{rem:distribution_rates} 
We illustrate the preceding results with several common response distributions. For Gaussian responses with the identity link, Theorem~\ref{thm:converge in prob} gives \[ \frac{1}{NJ} \|\widehat\MM^{\mathrm{SVD}}-\MM^*\|_F^2 = O_p(J^{-1}). \] For binomial responses $y_{ij}\sim\mathrm{Binomial}(L,\sigma(m_{ij}^*))$ with fixed $L$ and $\sigma(m)=(1+e^{-m})^{-1}$, the link function is $f(m)=L\sigma(m)$ and the response space is $\mathcal Y=\{0,\ldots,L\}$. Taking $a_{NJ}=0$ and $b_{NJ}=L$ gives $R_{NJ}=L$ and $r_{NJ}=0$, and hence \[ \frac{1}{NJ} \|\widehat\MM^{\mathrm{SVD}}-\MM^*\|_F^2 = O_p\left( J^{-\frac{1}{K+2}} \right). \] The Bernoulli model corresponds to $L=1$. For Poisson responses $y_{ij}\sim\mathrm{Poisson}(\exp(m_{ij}^*))$, the response space is $\mathcal Y=\{0,1,\ldots\}$. Taking $a_{NJ}=0$ and $b_{NJ}=\kappa\log J$ for any fixed $\kappa>0$ gives $R_{NJ}\asymp\log J$, while $r_{NJ}$ vanishes faster than any polynomial in $J$. Therefore, \[ \frac{1}{NJ} \|\widehat\MM^{\mathrm{SVD}}-\MM^*\|_F^2 = O_p\left( J^{-\frac{1}{K+2}} (\log J)^{\frac{K+4}{K+2}} \right). \] 
\end{remark}

\subsubsection{Entrywise Consistency for One-Step Refinement}

For the identity link, Theorem~\ref{thm:converge in prob} shows that
the spectral estimator $\widehat\MM^{\mathrm{SVD}}$ is already uniformly
consistent under the max norm, so no refinement is required.
For nonlinear links, the spectral denoising results above provide
normalized Frobenius consistency but do not directly yield entrywise
control. We therefore apply the one-step refinement to upgrade a
Frobenius-norm accurate initializer to a max-norm accurate estimator.
The following generic refinement result applies to both observation
structures.
Throughout the refinement analysis, the projection radius $C$ is chosen as a sufficiently large fixed constant such that
$\|\AAA^*\|_{2\to\infty}\le C$, with $J=N$ for unipartite data.

\begin{theorem}\label{thm:refinement}
Suppose that $f$ is nonlinear and Assumption~\ref{assumption:algo1} holds. Let $\MM^{\mathrm{init}}$ be an initial estimator of $\MM^*$ satisfying
$\|\MM^{\mathrm{init}}-\MM^*\|_F=O_p(e_{\MM,F})$ for some deterministic
rate sequence $e_{\MM,F}$ such that
$$
\sqrt N
\lesssim
e_{\MM,F}
\ll
\sqrt{NJ}(\log(NJ))^{-2}.
$$
Apply the one-step refinement stage of Algorithm~\ref{alg:rsd} with
initialization $\MM^{\mathrm{init}}$. Then, with probability tending to one,
all row- and column-wise estimating equations admit unique solutions.
Let $\widehat\MM^{\mathrm{ref}}$ denote the resulting refined estimator. Then
$$
\|\widehat\MM^{\mathrm{ref}}-\MM^*\|_{\max}
=
O_p\left(
(NJ)^{-1/2}(\log(NJ))^2e_{\MM,F}
\right).
$$
\end{theorem}

\begin{remark}
Theorem~\ref{thm:refinement} shows that the one-step refinement stage upgrades a Frobenius-norm accurate initializer into an entrywise accurate estimator while preserving its normalized Frobenius convergence rate up to a squared logarithmic factor. The intuition can be seen under the factor representation $\MM^*=\BBB^*\AAA^{*\top}$. The initial estimator gives an accurate estimate of the leading right singular subspace and hence provides an anchor matrix $\widehat\AAA$ close to $\AAA^*$. Given the anchor, the estimation errors for $\widetilde\BBB$ from the row-wise estimating equations come from both the random noise and the plug-in error of $\widehat\AAA$. The column-wise estimating equations then use $\widetilde\BBB$ as the anchor matrix and similarly refine $\widehat\AAA$. These two factor-level refinements yield two-to-infinity norm error controls for both $\widetilde\BBB$ and $\widetilde\AAA$, and hence yield an entrywise error bound for $\widehat\MM^{\mathrm{ref}}$.
\end{remark}

\begin{corollary}\label{cor:refinement_algo1}
Suppose that $f$ is nonlinear and Assumption~\ref{assumption:algo1} holds, with $J=N$ for
unipartite data. Let $a_{NJ}$ and $b_{NJ}$ be chosen as in Corollary~\ref{cor:support_adaptive_truncation}. For unipartite data, additionally suppose that $f(\tau_0)\in[a_{NN},b_{NN}]$ for all sufficiently large $N$.
Then
\[
\|\widehat{\MM}^{\mathrm{RSD}}-\MM^*\|_{\max}
=
O_p\left(
(\log(NJ))^2
J^{-\frac{1}{2(K+2)}}
(\log J)^{\frac{K+4}{2(K+2)}}
\right).
\]
If the response supports are uniformly bounded, the factor
$(\log J)^{\frac{K+4}{2(K+2)}}$ can be omitted.
\end{corollary}

Theorem~\ref{thm:converge in prob} and Corollary~\ref{cor:refinement_algo1} together show that Algorithm~\ref{alg:rsd} yields uniform entrywise consistency under both link regimes. For the identity link, this property is already achieved by the spectral estimator and no refinement is needed; for nonlinear links, the one-step refinement upgrades a Frobenius-consistent spectral initializer under the stated truncation choices to uniform entrywise consistency. The resulting max-norm bounds provide the uniform signal estimation accuracy required for the subsequent configuration recovery in both the unipartite and bipartite settings.

\subsection{Configuration Recovery}\label{subsec:theory_conf_recover} %

In this section, let $\widehat\MM$ be a generic estimator of $\MM^*$ such that $\xi=\|\widehat\MM-\MM^*\|_{\max}=o(1)$.

\subsubsection{Unipartite Configuration Recovery}

We first establish the consistency of the unipartite configuration recovery procedure introduced in Section~\ref{subsec:method_conf_recover}. The next theorem gives the configuration recovery error based on a generic uniformly consistent estimator $\widehat\MM$.

\begin{theorem}\label{thm:uni_conf}
Suppose Assumption~\ref{assumption:algo1_bounded} holds,
$\EE_N^\top\TTT^*=\mathbf 0$, and
\[
\liminf_{N\to\infty}
\lambda_{\min}\left(N^{-1}\TTT^{*\top}\TTT^*\right)>0.
\]
Let $\widehat\MM$ be symmetric and let $\widehat\TTT$ be the estimator obtained from the unipartite configuration recovery procedure in Section~\ref{subsec:method_conf_recover}. Then $d_c(\widehat\TTT,\TTT^*) \lesssim \xi$.
\end{theorem}

According to Theorem~\ref{thm:uni_conf}, for unipartite data, a uniformly accurate estimator of the signal matrix translates directly into uniform recovery of the latent configuration, without further deterioration in the convergence rate.

\subsubsection{Bipartite Configuration Recovery} 
For $\rrr =(\tau,\ve(\FF)^\top,\bb^\top)^\top$, define its sign-equivalence class by
$$
[\rrr]
:=
\left\{
\left(
\tau,\,
\ve(\FF\QQ)^\top,\,
(\QQ^\top\bb)^\top
\right)^\top
:
\QQ\in\mathcal S_K
\right\}.
$$
Thus, two parameter vectors are sign-equivalent if they differ only by column-wise sign changes of $\FF$ and the corresponding sign changes of $\bb$. To ensure the statistical consistency of the recovered latent configurations, we impose the following regularity conditions on the true parameters.  

\begin{assumption}\label{assumption:algo3}
Suppose $\rrr_0\in \GGamma$ and the following conditions hold.

    \assumptionitem \label{assumption:algo3_diag}
    $\EE_N^\top\TTT_{\mathcal P_1}^* = {\bf 0}$ and
    $N^{-1}\TTT_{\mathcal P_1}^{*\top}\TTT_{\mathcal P_1}^*$ is diagonal and converges to a matrix with strictly decreasing positive diagonal entries.
    Further,
    $
    \min\limits_{1\le k\le K}
    \left\{
    \sigma_k\left(\JJ_N\MM^*\JJ_J\right)
    -
    \sigma_{k+1}\left(\JJ_N\MM^*\JJ_J\right)
    \right\}
    =
    \Omega(\sqrt{NJ}),
    $ and the first nonzero entry of each left singular vector
    $\uu_k^*$ of $\JJ_N\MM^*\JJ_J$ is positive.

    \assumptionitem \label{assumption:algo3_identifiability}
    There exists a diverging sequence $q_{NJ}$ such that
    $$
    \liminf\limits_{(N,J)\to\infty}
    q_{NJ}^{-1}
    \sigma_{\frac{(K+1)(K+2)}{2}}
    \bigl(\partial \MM(\rrr_0;\Z^*)\bigr)>0,
    $$ 
    where $\partial\MM(\rrr;\Z)\in\R^{NJ\times(K(K+1)+1)}$ is the Jacobian matrix of the mapping $\rrr\mapsto\ve(\MM(\rrr;\Z))$.

    
    \assumptionitem \label{assumption:algo3_global} 
    There exist positive deterministic sequences $\rho_{NJ}$ and $c_{NJ}$ such that 
    $$ 
    \liminf\limits_{(N,J)\to\infty} c_{NJ}^{-1} \inf\limits_{ \rrr\in\GGamma:\, \inf_{\widetilde\rrr\in[\rrr_0]} \|\rrr-\widetilde\rrr\| \ge \rho_{NJ} } (NJ)^{-1}l(\rrr;\Z^*,\MM^*)>0. 
    $$
\end{assumption}

\begin{remark}
As discussed in Section~\ref{subsec:method_conf_recover}, Assumption~\ref{assumption:algo3_diag} fixes the shift and rotational ambiguities of the configuration up to column-wise sign changes. In particular, $\FF_0^\top\FF_0 = \TTT_{\mathcal P_1}^{*\top}\TTT_{\mathcal P_1}^*/N$ is diagonal with asymptotically distinct positive diagonal entries, which motivates the diagonal constraint on $\FF^\top\FF$ in $\GGamma$ in \eqref{eq:opti}. The singular-value gap condition ensures the stability of the SVD-based recovery, while the sign normalization on $\UU^*$ fixes the representation of the population singular vectors. 
The quantity $q_{NJ}$ measures the local identification strength of the configuration. As shown in the Theorem~\ref{thm:bi_conf} below, $\sqrt{NJ}/q_{NJ}$ acts as an amplification
factor that determines how estimation error in the signal matrix propagates to the recovered configuration. 
Assumption~\ref{assumption:algo3_identifiability} quantifies local identification through the Jacobian $\partial\MM(\rrr_0;\Z^*)$: its first $(K+1)(K+2)/2$ singular values are bounded below at scale $q_{NJ}$, while the remaining $K(K-1)/2$ directions correspond to rotational invariance of the unconstrained parameterization. Assumption~\ref{assumption:algo3_global} complements this local condition by requiring separation, at scale $c_{NJ}$, outside a $\rho_{NJ}$-neighborhood of the sign-equivalence class $[\rrr_0]$. Together, these conditions characterize local identification and global separation of the true parameter modulo column-wise sign changes. Unlike the anchor-point conditions used in \citet{chen2021unfolding}, they formulate configuration identifiability directly through the local Jacobian and the population loss, without requiring particular anchor
points.
\end{remark}

\begin{theorem}\label{thm:bi_conf} 
Suppose Assumptions~\ref{assumption:algo1_bounded} and~\ref{assumption:algo3} hold, and suppose further that 
$\xi^2=o(c_{NJ})$ and $\rho_{NJ} = o\left(q_{NJ}/\sqrt{NJ}\right)$. Then, for all sufficiently large $N$ and $J$, the minimum in \eqref{eq:opti} is attained. Let $\widehat\rrr$ be any global minimizer of \eqref{eq:opti}. Then
$$ 
d_c(\TTT(\widehat\rrr;\widehat\Z), \TTT^*)
\lesssim
\frac{\sqrt{NJ}}{q_{NJ}}\xi.
$$ 
\end{theorem}

\begin{remark}
The additional rate conditions in Theorem~\ref{thm:bi_conf} have distinct roles in the perturbation analysis. The condition $\xi^2=o(c_{NJ})$ ensures that the squared perturbation of the signal matrix is asymptotically smaller than the global separation in Assumption~\ref{assumption:algo3_global}, so that any global minimizer is localized within a $\rho_{NJ}$-neighborhood of $[\rrr_0]$. 
The condition $\rho_{NJ}=o(q_{NJ}/\sqrt{NJ})$ ensures that, within this neighborhood, the linear term of the mapping
$\rrr\mapsto\ve\{\MM(\rrr;\Z^*)\}$ governed by the Jacobian
$\partial\MM(\rrr_0;\Z^*)$ dominates the higher-order remainder. Given the conditions, the uniform error rate for configuration recovery is $\sqrt{NJ}\xi/q_{NJ}$, which is determined by both the estimation error $\xi$ for the signal matrix and the local identification strength $q_{NJ}$ of the true configuration. 
In the standard well-conditioned regime $q_{NJ}\asymp\sqrt{NJ}$, the configuration recovery error has the same order as the signal estimation error.
\end{remark}

\begin{corollary} \label{cor:config_rsd} 
Suppose Assumption~\ref{assumption:algo1} holds, together with the
additional conditions of Theorem~\ref{thm:uni_conf} for unipartite data or Assumption~\ref{assumption:algo3} for bipartite data. Let $\widehat\MM$ be the output of Algorithm~\ref{alg:rsd}, that is,
\[
\widehat\MM
=
\begin{cases}
\widehat\MM^{\mathrm{SVD}}, & f(x)=x,\\
\widehat\MM^{\mathrm{RSD}}, & f \text{ is nonlinear}.
\end{cases}
\]
where, in the nonlinear case, $a_{NJ}$ and $b_{NJ}$ are chosen as in Corollary~\ref{cor:support_adaptive_truncation}. For nonlinear unipartite data, additionally suppose that $f(\tau_0)\in[a_{NN},b_{NN}]$ for all sufficiently large $N$.
Define 
$$ 
\delta_{NJ} = 
\begin{cases} 
\sqrt{\log(NJ)}J^{-1/2}, & f(x)=x,\\
(\log(NJ))^2 J^{-\frac{1}{2(K+2)}} (\log J)^{\frac{K+4}{2(K+2)}}, & f \text{ is nonlinear}.
\end{cases} 
$$ 
For bipartite data, suppose further that $\delta_{NJ}^2=o(c_{NJ})$ and $\rho_{NJ} = o\left(q_{NJ}/\sqrt{NJ}\right)$. Let $\widehat\TTT$ denote the configuration estimator obtained by applying the corresponding configuration recovery procedure to $\widehat\MM$.
$$ 
d_c(\widehat\TTT, \TTT^*)
= \begin{cases} O_p(\delta_{NJ}), & \text{for unipartite data},\\[2mm] O_p\!\left( \dfrac{\sqrt{NJ}}{q_{NJ}}\delta_{NJ} \right), & \text{for bipartite data}. \end{cases} 
$$ 
If the response supports are uniformly bounded, the factor $(\log J)^{\frac{K+4}{2(K+2)}}$ in $\delta_{NJ}$ can be omitted.
\end{corollary}

\begin{remark}
Corollary~\ref{cor:config_rsd} follows by combining the identity-link max-norm bound in Theorem~\ref{thm:converge in prob} and the nonlinear-link max-norm bound in Corollary~\ref{cor:refinement_algo1} with the configuration recovery bounds in Theorems~\ref{thm:uni_conf} and~\ref{thm:bi_conf}. For unipartite data, the configuration recovery step preserves the rate of the signal estimator returned by Algorithm~\ref{alg:rsd}. For bipartite data, the recovery rate additionally depends on the local identification strength through the factor $\sqrt{NJ}/q_{NJ}$; when $q_{NJ}\asymp\sqrt{NJ}$, it has the same order as the corresponding signal estimator. For nonlinear responses with uniformly bounded response supports, the sharper rate follows by omitting the additional logarithmic factor in Corollary~\ref{cor:refinement_algo1}.
\end{remark}

\section{Simulation Study}\label{sec:simulation} %

In this section, we conduct simulation studies to examine the finite-sample performance of the proposed method. 

\subsection{Simulation Setup}\label{subsec:setup}
We consider latent dimensions $K\in\{3,5\}$. For the bipartite setting, we set $N=2J$ with $J\in\{300,500,1000,2000\}$. For the unipartite setting, we set $N=J$ and consider symmetric $N\times N$ matrices with $N\in\{300,500,1000,2000\}$. Each setting is replicated 50 times.

For each case, $\TTT_{\mathcal P_1}^*$ is constructed from an $N\times K$ matrix with i.i.d.\ $\mathrm{Uniform}(-1,1)$ entries. We first center each column and orthogonalize the centered columns using a QR decomposition, and then rescale each column. Specifically, we rescale the columns so that
$$
\frac{1}{N}
(\TTT_{\mathcal P_1}^*)^\top\TTT_{\mathcal P_1}^*
=
\operatorname{diag}\bigl(\boldsymbol{\lambda}^{(K)}\bigr),
$$
where $\boldsymbol{\lambda}^{(3)} = (0.20,0.18,0.16)^\top$ and $\boldsymbol{\lambda}^{(5)} =(0.20,0.18,0.16,0.14,0.12)^\top$. The decreasing entries of $\boldsymbol{\lambda}^{(K)}$ give distinct, nondegenerate latent directions, which is consistent with the normalization and nondegeneracy conditions in Assumption~\ref{assumption:algo3_diag}. For the bipartite setting, $\TTT_{\mathcal P_2}^*$ is independently generated from a $J\times K$ matrix with i.i.d. $\mathrm{Uniform}(-1,1)$ entries and rescaled column-wise so that the mean squared entry of each column matches that of the corresponding column of $\TTT_{\mathcal P_1}^*$. 
For the unipartite setting, $\mathcal P_1=\mathcal P_2=[N]$.

Recall that
$$
m_{ij}^*
=
\begin{cases}
\tau_0-\|\ttt_i^*-\ttt_{N+j}^*\|^2,
& \text{for the bipartite setting},\\
\tau_0-\|\ttt_i^*-\ttt_j^*\|^2,
& \text{for the unipartite setting}.
\end{cases}
$$
We use the same data-generating process for the latent configurations across the Gaussian, Binary, and Poisson observation models. We choose $\tau_0$ so that the average of the entries of $\MM^*$ equals 0 for the Gaussian and Binary models. For the Poisson model, we instead center the entries of $\MM^*$ to $0.5$, yielding larger expected counts and thereby reducing the prevalence of zero observations. Thus, all three observation models share the same latent geometry, while the Poisson natural-parameter matrix includes an additional intercept shift of $0.5$. We then generate observations according to the corresponding observation model. For unipartite data, observations are generated for $i<j$, and we set $y_{ji}=y_{ij}$ and $y_{ii}=0$.

\begin{itemize}
    \item \textbf{Homoskedastic Gaussian model.}
    Let $y_{ij}\sim N(m_{ij}^*,1)$.

    \item \textbf{Heteroskedastic Gaussian model.}
    Let $y_{ij}\sim N(m_{ij}^*,\sigma_{ij}^2)$, where
    $\sigma_{ij}^2\overset{\mathrm{i.i.d.}}{\sim}
    \mathrm{Uniform}(0.4,1.6)$.

    \item \textbf{Binary model.}
    Let $y_{ij}\sim\mathrm{Bernoulli}(p_{ij})$, where
    $p_{ij}=1/\{1+\exp(-m_{ij}^*)\}$.

    \item \textbf{Poisson model.}
    Let $y_{ij}\sim\mathrm{Poisson}(\lambda_{ij})$, where
    $\lambda_{ij}=\exp(m_{ij}^*)$.
\end{itemize}

For the nonlinear-link models, we set $(a_{NJ},b_{NJ})=(0,1)$ for the
binary model and $(a_{NJ},b_{NJ})=(0,3\log J)$ for the Poisson model. We use the projection constant $C=7$ in the spectral-denoising stage for all observation models and also in the one-step-refinement stage for the binary and Poisson models.

\subsection{Results}
\label{subsec:simulation-results}

To assess the effect of refinement, we compare
$\widehat\MM^{\mathrm{SVD}}$ and $\widehat\MM^{\mathrm{RSD}}$
under the max-norm loss. For the Gaussian models,
$\widehat\MM^{\mathrm{RSD}}$ is computed only as a diagnostic comparison,
since Algorithm~\ref{alg:rsd} returns
$\widehat\MM^{\mathrm{SVD}}$ under the identity link.
Configuration recovery is always based on the final output of
Algorithm~\ref{alg:rsd}, namely $\widehat\MM^{\mathrm{SVD}}$ for the
Gaussian models and $\widehat\MM^{\mathrm{RSD}}$ for the binary and
Poisson models.
For $r\in\{\mathrm{SVD},\mathrm{RSD}\}$, we denote
$\mathcal{E}_{\max}^{(r)}
=
\|\widehat{\MM}^{r}-\MM^{*}\|_{\max}$.
For configuration recovery, we report $d_c(\widehat\TTT,\TTT^*)$.
We also report
$|\widehat\tau-\tau_0|$ under bipartite settings.
Results for $K=3$ are presented below, while those for $K=5$ are provided in Appendix~\ref{app:additional-results}.

\begin{figure}[!htb]
    \centering

    \includegraphics[
        width=0.94\textwidth,
        height=0.25\textheight,
        keepaspectratio
    ]
    {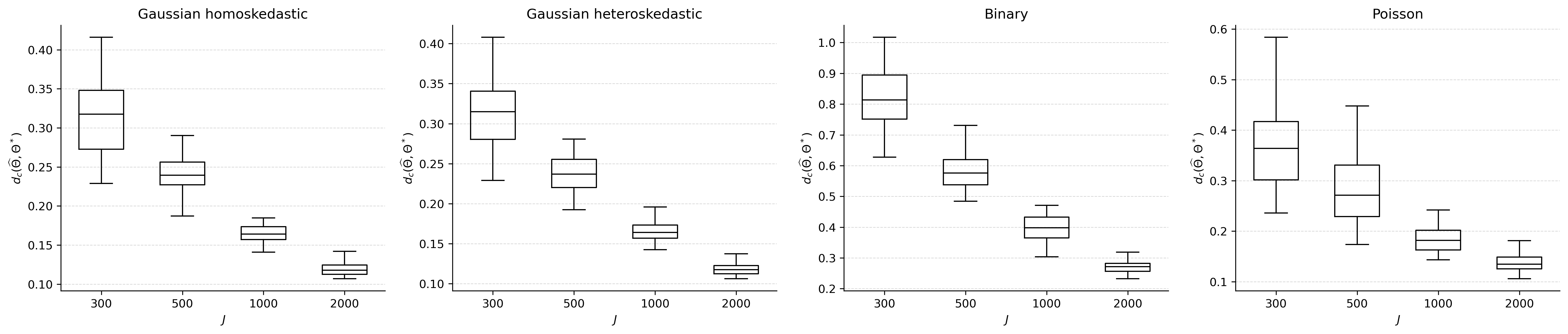}

    \caption{
        Configuration recovery error $d_c(\widehat\TTT,\TTT^*)$
        for the unipartite model with $K=3$.
    }
    \label{fig:uni_k3-recovery}
\end{figure}

\begin{figure}[!htb]
    \centering

    \includegraphics[width=0.96\textwidth]
    {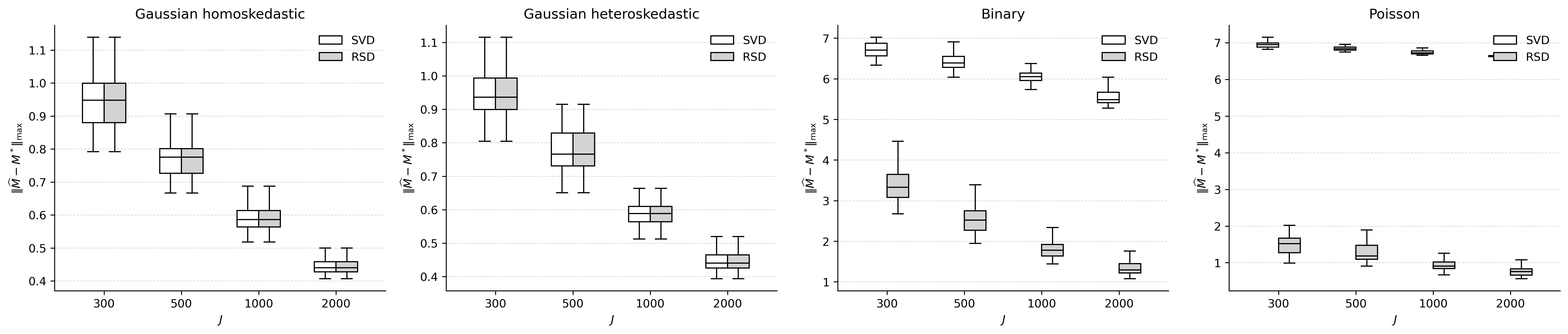}

    \caption{
        Max-norm errors of $\widehat\MM^{\mathrm{SVD}}$ and
$\widehat\MM^{\mathrm{RSD}}$ for the bipartite model with $K=3$.
    }
    \label{fig:bi_k3-entrywise}
\end{figure}

\begin{figure}[!htb]
    \centering

    \begin{subfigure}[t]{0.94\textwidth}
        \centering
        \includegraphics[
            width=\linewidth,
            height=0.21\textheight,
            keepaspectratio
        ]
        {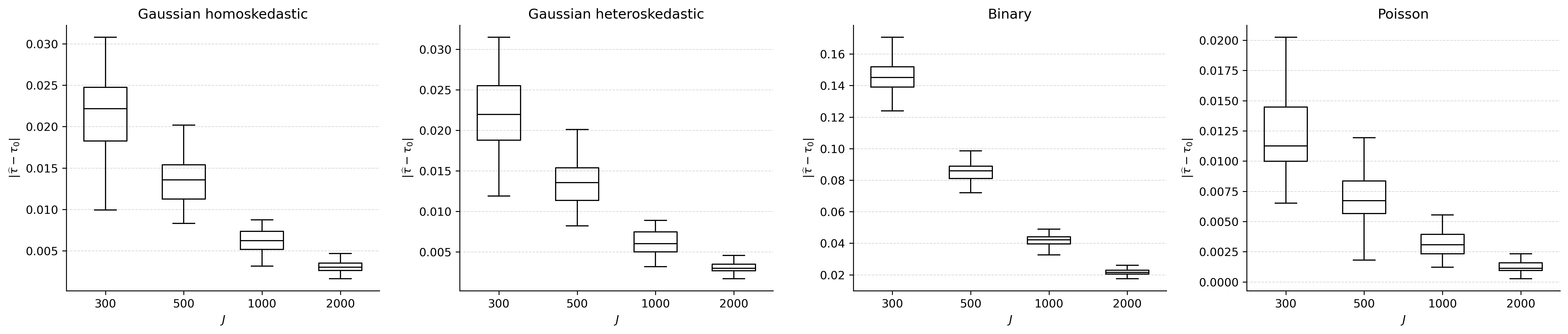}
        \caption{
            $|\widehat{\tau}-\tau_0|$.
        }
        \label{fig:bi_k3-tau}
    \end{subfigure}

    \vspace{0.4em}

    \begin{subfigure}[t]{0.94\textwidth}
        \centering
        \includegraphics[
            width=\linewidth,
            height=0.21\textheight,
            keepaspectratio
        ]
        {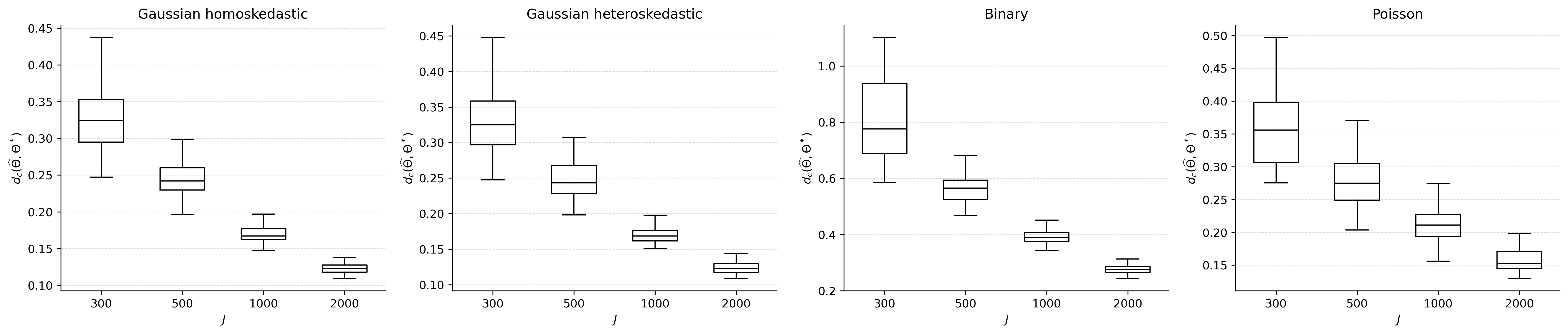}
        \caption{
            $d_c(\widehat\TTT,\TTT^*)$.
        }
        \label{fig:bi_k3-p1-configuration}
    \end{subfigure}

    \caption{
        $|\widehat{\tau}-\tau_0|$ and $d_c(\widehat\TTT,\TTT^*)$ for the
        bipartite model with $K=3$. 
    }
    \label{fig:bi_k3-recovery}
\end{figure}

Figures~\ref{fig:uni_k3-entrywise}--\ref{fig:uni_k3-recovery} show the
results for the unipartite setting, while
Figures~\ref{fig:bi_k3-entrywise}--\ref{fig:bi_k3-recovery} show the
corresponding results for the bipartite setting, together with the drift parameter error.
Across both observation structures, the one-step refinement
substantially reduces the entrywise maximum error for the binary and
Poisson models, and the refined max-norm error decreases steadily as
the matrix size increases. In contrast, the max-norm error of the
spectral estimator remains comparatively large for these nonlinear models. This pattern is consistent with the role of refinement in our
theory: the spectral stage provides a Frobenius-consistent initializer,
while the refinement yields uniform entrywise control.
For the Gaussian models, Algorithm~\ref{alg:rsd} returns
$\widehat\MM^{\mathrm{SVD}}$, which is already max-norm consistent by
Theorem~\ref{thm:converge in prob}. In the bipartite setting, the
additional refinement has little empirical effect. In the unipartite
setting, a small finite-sample deterioration is observed after
refinement, possibly because the spectral estimator directly exploits
the symmetric one-mode structure, whereas refinement temporarily
estimates the row and column factors separately before reimposing
symmetry at the final step.
Across all four models, the configuration recovery error
decreases with the matrix size; the bipartite drift estimation error shows the same overall trend.

\begin{figure}[!htb]
    \centering

    \includegraphics[width=0.96\textwidth]
    {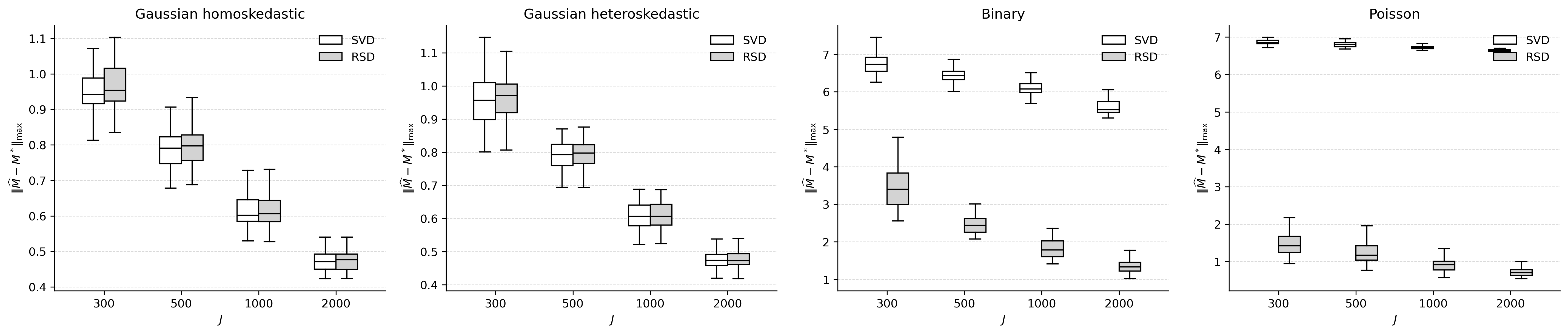}

    \caption{Max-norm errors of $\widehat\MM^{\mathrm{SVD}}$ and
$\widehat\MM^{\mathrm{RSD}}$ for the unipartite model with $K=3$.
    }
    \label{fig:uni_k3-entrywise}
\end{figure}

Table~\ref{tab:runtime} reports the computational time of the proposed estimation procedure.\footnote{All runtime experiments were conducted on a server equipped with two AMD EPYC 9654 96-Core processors and approximately 1.1 TiB of RAM, using Python 3.11.14.} For each setting, we record the wall-clock time of the complete estimation pipeline, including spectral denoising and configuration recovery for the Gaussian models, and spectral denoising, one-step refinement, and configuration recovery for the binary and Poisson models.
Each experiment is repeated 50 times, and the mean runtime is reported. As shown in Table~\ref{tab:runtime}, the runtime increases with the matrix size, as expected, while the difference between $K=3$ and $K=5$ is generally moderate. The Gaussian models are computationally less demanding than the binary and Poisson models, for which the refinement step accounts for most of the computation. Even for the largest setting considered, with $J=2000$, the mean runtime remains below $20$ seconds across all cases. These results demonstrate that the proposed procedure scales efficiently to large relational matrices, completing the full estimation pipeline within seconds even at the largest simulated sizes.

\begin{table}[!htb]
\centering
\scriptsize
\renewcommand{\arraystretch}{1.08}
\setlength{\tabcolsep}{3.5pt}

\begin{tabularx}{\textwidth}{
    >{\centering\arraybackslash}p{0.85cm}
    >{\centering\arraybackslash}X
    >{\centering\arraybackslash}p{1.75cm}
    >{\centering\arraybackslash}p{1.75cm}
    >{\centering\arraybackslash}p{1.75cm}
    >{\centering\arraybackslash}p{2.00cm}
}
\toprule
$K$ & Observation model & $J=300$ & $J=500$ & $J=1000$ & $J=2000$ \\
\midrule
\multicolumn{6}{c}{\textit{Unipartite}} \\
\midrule
\multirow[c]{4}{*}{$K=3$} & Homoskedastic Gaussian & 0.007 (0.002) & 0.018 (0.015) & 0.118 (0.082) & 1.102 (0.800) \\
 & Heteroskedastic Gaussian & 0.007 (0.005) & 0.014 (0.014) & 0.065 (0.048) & 0.715 (0.638) \\
 & Binary & 0.186 (0.230) & 0.291 (0.069) & 0.947 (0.253) & 7.016 (3.703) \\
 & Poisson & 0.219 (0.186) & 0.457 (0.350) & 1.100 (0.272) & 6.318 (3.337) \\[2pt]
\multirow[c]{4}{*}{$K=5$} & Homoskedastic Gaussian & 0.008 (0.006) & 0.013 (0.004) & 0.090 (0.052) & 0.803 (0.627) \\
 & Heteroskedastic Gaussian & 0.006 (0.002) & 0.012 (0.006) & 0.065 (0.037) & 0.779 (0.652) \\
 & Binary & 0.232 (0.263) & 0.338 (0.059) & 1.169 (0.507) & 7.166 (2.187) \\
 & Poisson & 0.204 (0.040) & 0.402 (0.046) & 1.185 (0.295) & 4.775 (1.356) \\
\midrule
\multicolumn{6}{c}{\textit{Bipartite}} \\
\midrule
\multirow[c]{4}{*}{$K=3$} & Homoskedastic Gaussian & 0.037 (0.015) & 0.151 (0.118) & 0.862 (0.373) & 3.029 (2.069) \\
 & Heteroskedastic Gaussian & 0.047 (0.035) & 0.062 (0.070) & 0.613 (0.416) & 3.282 (2.183) \\
 & Binary & 0.277 (0.247) & 0.620 (0.649) & 3.120 (2.468) & 19.618 (11.382) \\
 & Poisson & 0.388 (0.657) & 0.560 (0.610) & 3.416 (2.639) & 10.991 (6.360) \\[2pt]
\multirow[c]{4}{*}{$K=5$} & Homoskedastic Gaussian & 0.050 (0.010) & 0.090 (0.072) & 0.695 (0.331) & 2.610 (1.815) \\
 & Heteroskedastic Gaussian & 0.066 (0.095) & 0.079 (0.062) & 0.587 (0.471) & 1.527 (1.205) \\
 & Binary & 0.246 (0.037) & 0.586 (0.560) & 2.049 (1.204) & 17.583 (10.147) \\
 & Poisson & 0.338 (0.207) & 0.660 (0.351) & 2.641 (1.176) & 11.719 (4.618) \\
\bottomrule
\end{tabularx}

\caption{
Mean runtime (seconds), with standard errors in parentheses,
over 50 replications.
}
\label{tab:runtime}
\end{table}

\subsection{Comparison with Likelihood-Based Estimation}

We further compare the proposed procedure with likelihood-based alternatives in the bipartite binary setting, as considered in Section~\ref{subsec:setup}. 
The bipartite binary setting is most directly connected to existing likelihood-based probabilistic unfolding methods, including the approach of \cite{chen2021unfolding}, and therefore provides a natural benchmark for our procedure. 
We compare RSD $+$ CR with two likelihood-based alternatives designed to separate the effects of likelihood-based signal estimation and direct enforcement of the Euclidean distance geometry.
In particular, we compare three methods:
(i) the proposed RSD $+$ CR procedure, consisting of spectral denoising,
one-step refinement, and configuration recovery;
(ii) a two-stage likelihood-based procedure (low-rank MLE $+$ CR), which first estimates the low-rank signal matrix by maximum likelihood and then applies the same configuration recovery procedure as RSD $+$ CR; 
(iii) the joint maximum likelihood estimator (joint MLE), which directly estimates the latent configurations and the intercept by maximizing the Bernoulli likelihood using alternating optimization.
The joint maximum likelihood estimator is similar to the estimator in \cite{chen2021unfolding}, except that we have an additional drift parameter $\tau$ here.

Table~\ref{tab:binary_method_comparison} shows a clear
statistical--computational tradeoff among the three procedures.
For configuration recovery, joint MLE achieves the smallest errors
throughout, while low-rank MLE $+$ CR generally improves modestly over RSD $+$ CR and lies between the other two methods. The advantage of joint MLE is expected under the correctly specified
model, since it imposes the $K$-dimensional Euclidean distance structure throughout likelihood optimization, whereas low-rank MLE $+$ CR first
estimates a more general rank-$(K+2)$ signal matrix and only enforces the distance geometry in the subsequent recovery step; the two additional signal dimensions introduce extra nuisance flexibility and
can lead to some loss of statistical efficiency.
For the drift parameter, joint MLE is also consistently the most accurate, while low-rank MLE $+$ CR provides only a modest improvement over RSD $+$ CR.

In contrast, RSD $+$ CR is substantially faster than both likelihood-based alternatives.  Across the four sample sizes considered, it is approximately $11$--$28$ times faster than low-rank MLE $+$ CR and $13$--$34$ times faster than joint MLE. This substantial computational advantage arises because RSD $+$ CR relies on spectral denoising, one-step low-dimensional row- and column-wise refinements, and a final configuration-recovery optimization whose dimension depends only on $K$, whereas both likelihood-based alternatives require repeated iterative optimization.

\begin{table}[!htb]
\centering
\scriptsize
\renewcommand{\arraystretch}{1.08}

\begin{tabularx}{\textwidth}{
    >{\centering\arraybackslash}p{0.9cm}
    >{\centering\arraybackslash}X
    >{\centering\arraybackslash}p{2.4cm}
    >{\centering\arraybackslash}p{2.4cm}
    >{\centering\arraybackslash}p{2.8cm}
}
\toprule
$J$ & Method & $d_c(\widehat{\TTT},\TTT^*)$ & $|\widehat{\tau}-\tau_0|$ & Runtime (seconds) \\
\midrule
\multirow[c]{3}{*}{300} & RSD $+$ CR & 0.668 (0.064) & 0.145 (0.011) & 0.277 (0.247) \\
 & Low-rank MLE $+$ CR & 0.643 (0.053) & 0.143 (0.011) & 7.884 (5.356) \\
 & Joint MLE & 0.609 (0.061) & 0.088 (0.010) & 8.037 (1.304) \\
\midrule
\multirow[c]{3}{*}{500} & RSD $+$ CR & 0.505 (0.037) & 0.085 (0.007) & 0.620 (0.649) \\
 & Low-rank MLE $+$ CR & 0.492 (0.036) & 0.084 (0.007) & 14.353 (3.104) \\
 & Joint MLE & 0.465 (0.037) & 0.051 (0.006) & 21.046 (5.960) \\
\midrule
\multirow[c]{3}{*}{1000} & RSD $+$ CR & 0.374 (0.039) & 0.042 (0.004) & 3.120 (2.468) \\
 & Low-rank MLE $+$ CR & 0.362 (0.024) & 0.040 (0.003) & 57.062 (12.360) \\
 & Joint MLE & 0.332 (0.027) & 0.024 (0.003) & 106.488 (28.128) \\
\midrule
\multirow[c]{3}{*}{2000} & RSD $+$ CR & 0.272 (0.023) & 0.022 (0.002) & 19.618 (11.382) \\
 & Low-rank MLE $+$ CR & 0.266 (0.019) & 0.020 (0.002) & 210.383 (35.019) \\
 & Joint MLE & 0.242 (0.021) & 0.012 (0.002) & 258.160 (53.052) \\
\bottomrule
\end{tabularx}

\caption{
Comparison of RSD $+$ CR, low-rank MLE $+$ CR, and joint MLE
in the nonlinear bipartite binary setting with $K=3$ and $N=2J$. Each entry reports the mean, with standard errors in parentheses, over 50 replications.
}
\label{tab:binary_method_comparison}
\end{table}

\section{Real Data Analysis}\label{sec:realdata} %
For the real-data analyses, we select the latent dimension by inspecting the leading spectral values of the double-centered estimated signal matrix and choose $K$ at the dominant spectral elbow. In both applications, the leading spectral values indicate that a two-dimensional representation captures the main geometric structure, and we therefore set $K=2$ for
visualization.

\subsection{Analysis of a Statistician Citation Network}
We apply the proposed method to a real-world citation network of statisticians. The data set, collected by Ji et al. (2021), contains publication records from 36 statistical journals between 1975 and 2015.
To obtain a clear and interpretable visualization, we restrict the analysis to the top $0.3\%$ most active authors, resulting in a set of 142 statisticians. Based on their citation relationships, we construct an undirected intellectual network with adjacency matrix $\YY=(y_{ij})_{142\times142}$.
Specifically, the edge weight between statisticians $i$ and $j$ is defined as the symmetrized citation frequency after a log transformation to reduce the influence of large citation counts, $y_{ij}=\log(1+(C_{ij}+C_{ji})/{2})$, where $C_{ij}$ denotes the number of citations from statistician $i$ to statistician $j$. This symmetrization procedure converts the directed citation relationship into an undirected measure of intellectual connection strength.
We apply the proposed unipartite method to $\YY$ under a Gaussian observation model with the identity link.

Figure~\ref{fig:stat_network} presents the estimated two-dimensional latent configuration of the 142 statisticians. By applying K-means clustering with five clusters to these estimated embedding vectors, we partition the network into five distinct research communities. In the visualization, we manually select representative statisticians from the five K-means clusters and highlight them in dark red.
The blue dashed contours, generated based on the K-means partition, delineate the spatial boundaries of these communities. A legend listing the five research groups is provided in the top-right corner of the plot. The resulting visualization reveals clear separation among research
communities in the statistician citation network.

To interpret the five K-means clusters, we examine the main research areas of the statisticians in each cluster. The clusters can be broadly characterized as follows:

\begin{figure}[!htb]
    \centering
    \includegraphics[width=0.7\linewidth]{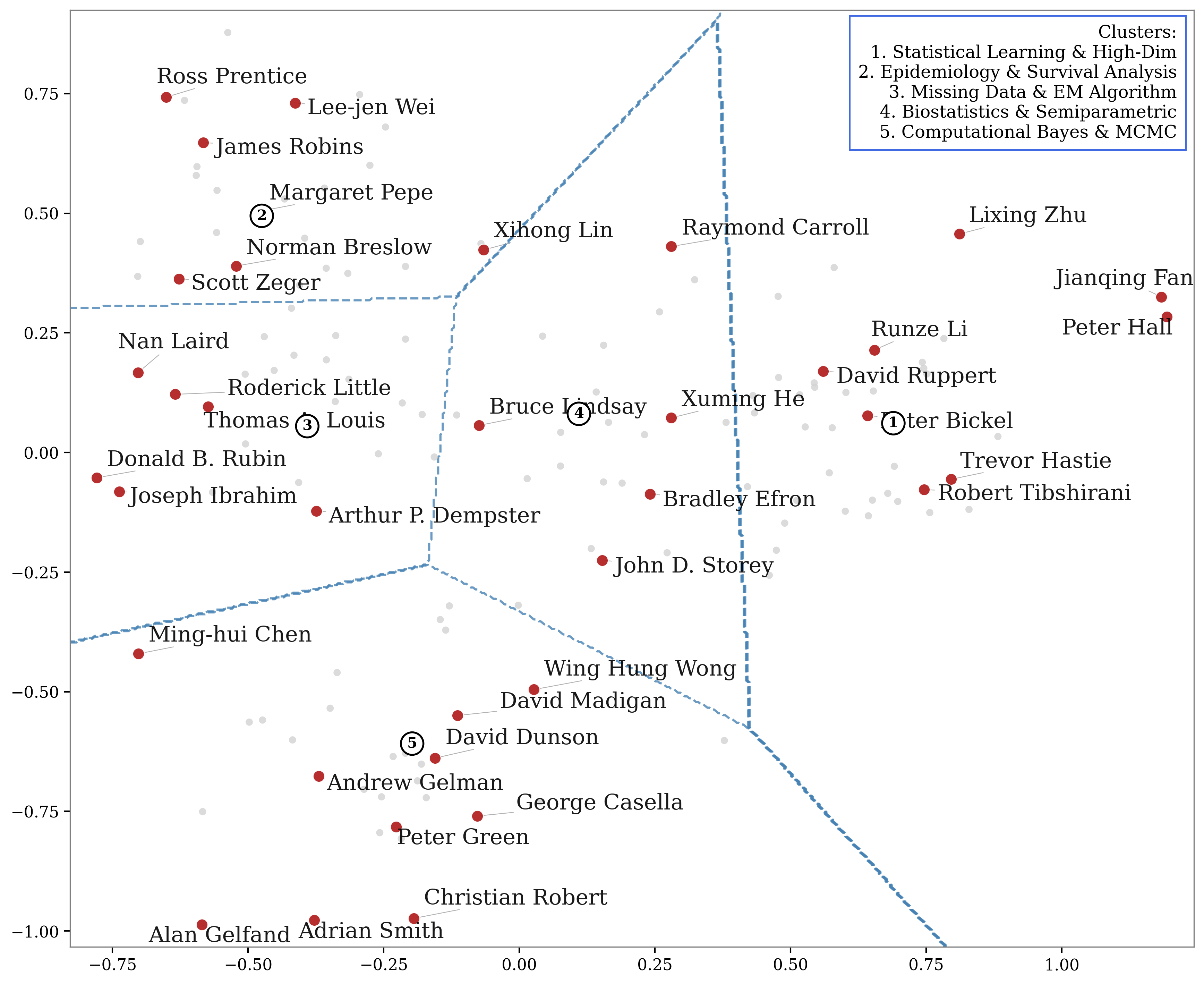}
    \caption{Two-dimensional visualization of the estimated embedding vectors for $142$ statisticians.}
    \label{fig:stat_network}
\end{figure}

\begin{itemize}
    \item \textbf{Statistical Learning and High-Dimensional Inference}: this cluster includes researchers working on nonparametric methods, regularization, resampling, and high-dimensional statistical learning. Representative figures include Jianqing Fan, Peter Hall, Robert Tibshirani, Trevor Hastie, Bradley Efron, and Peter Bickel.
    
    \item \textbf{Epidemiology, Survival Analysis, and Causal Inference}: this cluster centers on statisticians known for their foundational contributions to epidemiological studies and survival analysis. Notable statisticians in this group include James Robins, Ross Prentice, Lee-jen Wei, Norman Breslow, and Scott Zeger.
    
    \item \textbf{Missing Data, EM Algorithm, and Potential Outcomes}: this cluster comprises researchers renowned for their work on causal inference under the potential outcomes framework and missing-data methodologies. Key members include Donald B. Rubin, Nan Laird, Roderick Little, Thomas A. Louis, Joseph Ibrahim, and Arthur P. Dempster.
    
    \item \textbf{Biostatistics and Semiparametric/Nonparametric Regression}: this cluster brings together statisticians who have made significant advances in biostatistics, measurement error models, and semiparametric regression. Representative authors include Xihong Lin, Raymond Carroll, Xuming He, Bruce Lindsay, and David Ruppert.
    
    \item \textbf{Computational Bayes, MCMC, and Graphical Models}: this cluster consists of leading contributors to Markov chain Monte Carlo methods, computational Bayesian inference, and graphical models. Prominent names include Alan Gelfand, Adrian Smith, Christian Robert, David Dunson, Andrew Gelman, George Casella, and Peter Green.
\end{itemize}

The K-means clustering identifies several interpretable research communities without utilizing prior research-field labels in either model estimation or clustering. Moreover, the estimated embedding vectors provide geometric insights beyond discrete groupings. Statisticians whose work spans several research areas tend to appear near the cluster boundaries. For instance, Bradley Efron is positioned near the intersection of the statistical learning and biostatistics groups, which reflects his broad expertise spanning bootstrap resampling, empirical Bayes methods, and classical statistical inference. Thus, the recovered latent space effectively captures both the cohesive community structure of the citation network and the gradual, continuous relationships bridging different subfields of statistics.

\subsection{Analysis of the European Parliament Voting Data}

We further evaluate the proposed method using roll-call voting data from the European Parliament. The data set is obtained from the \textit{HowTheyVote.eu} repository,\footnote{The data is available at
\url{https://github.com/HowTheyVote/data/}.} which provides voting records for Members of the European Parliament (MEPs). Our analysis focuses on the 9th parliamentary term, covering the period from July 2, 2019, to July 15, 2024.

To formalize the analysis, we construct an observation matrix in which each row corresponds to a specific MEP and each column corresponds to a legislative procedure. The observed voting outcomes fall into three categories: votes in favor, votes against, and abstentions.
We code votes in favor, votes against, and abstentions as $1$, $-1$, and $0$, respectively. Our analysis includes all $867$ MEPs recorded in the data set who cast at least one vote during this term, and retains $1650$ distinct legislative procedures that contain at least one valid vote. Missing entries, including non-participation records, are imputed with $0$, treating them as a neutral midpoint on the coded voting scale and yielding a complete data matrix of size $867 \times 1650$. Let $x_{ij}$ denote the voting outcome for the $i$-th MEP on the $j$-th legislative procedure. Let $y_{ij} = x_{ij} + 1\in\{0,1,2\}$, and one possible specification is that
$
y_{ij}\sim\mathrm{Binomial}(2,\sigma(\tau-\|\ttt_i-\aaaa_j\|^2))
$
with $\sigma(\cdot)$ being the sigmoid function.
The induced link is $f(m)=2/(1+\exp(-m)).
$

\begin{figure}[!htb]
    \centering
    \begin{subfigure}[t]{0.5\textwidth}
        \centering
        \includegraphics[width=\linewidth]
        {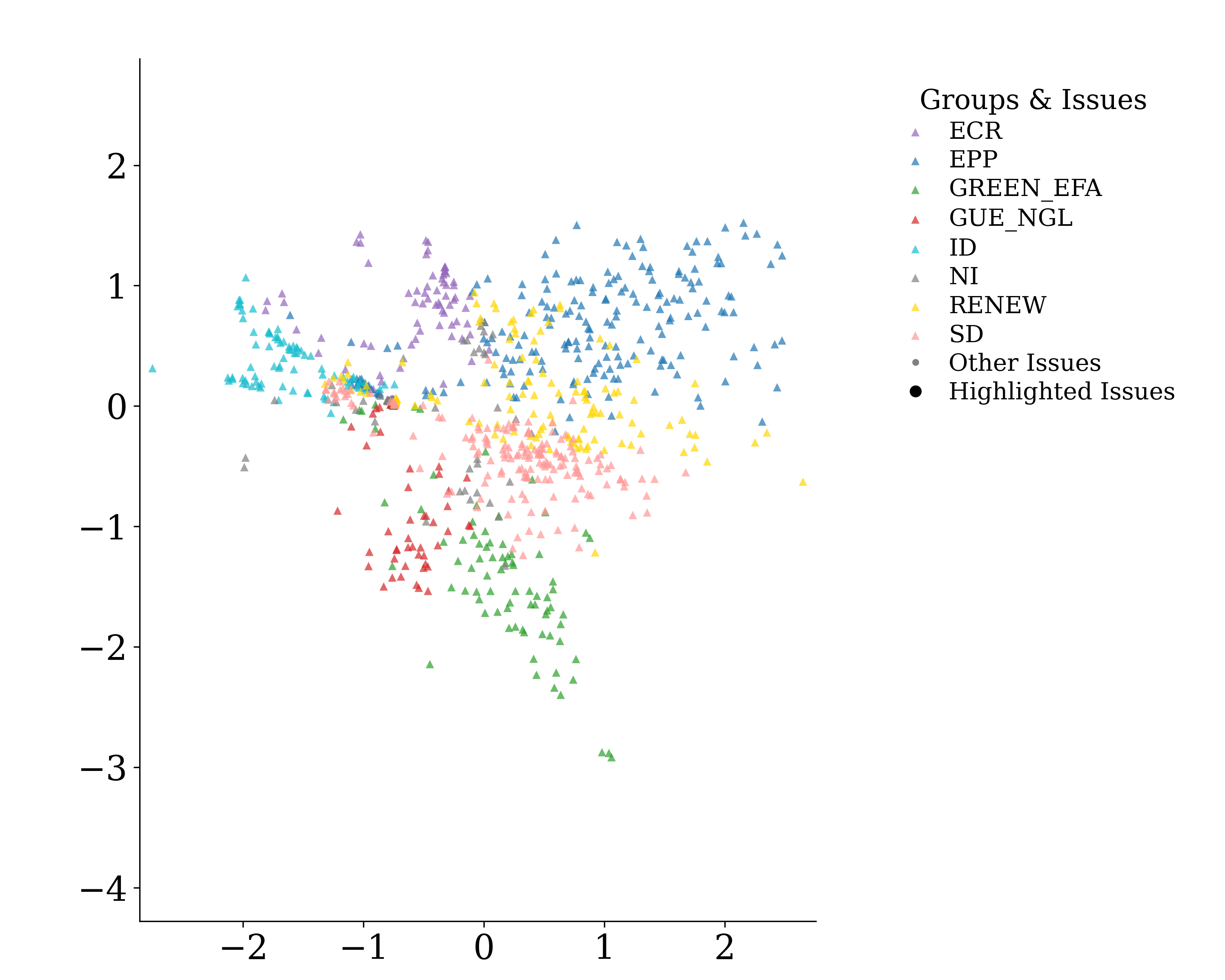}
    \end{subfigure}
    \hspace{-0.015\textwidth}
    \begin{subfigure}[t]{0.5\textwidth}
        \centering
        \includegraphics[width=\linewidth]
        {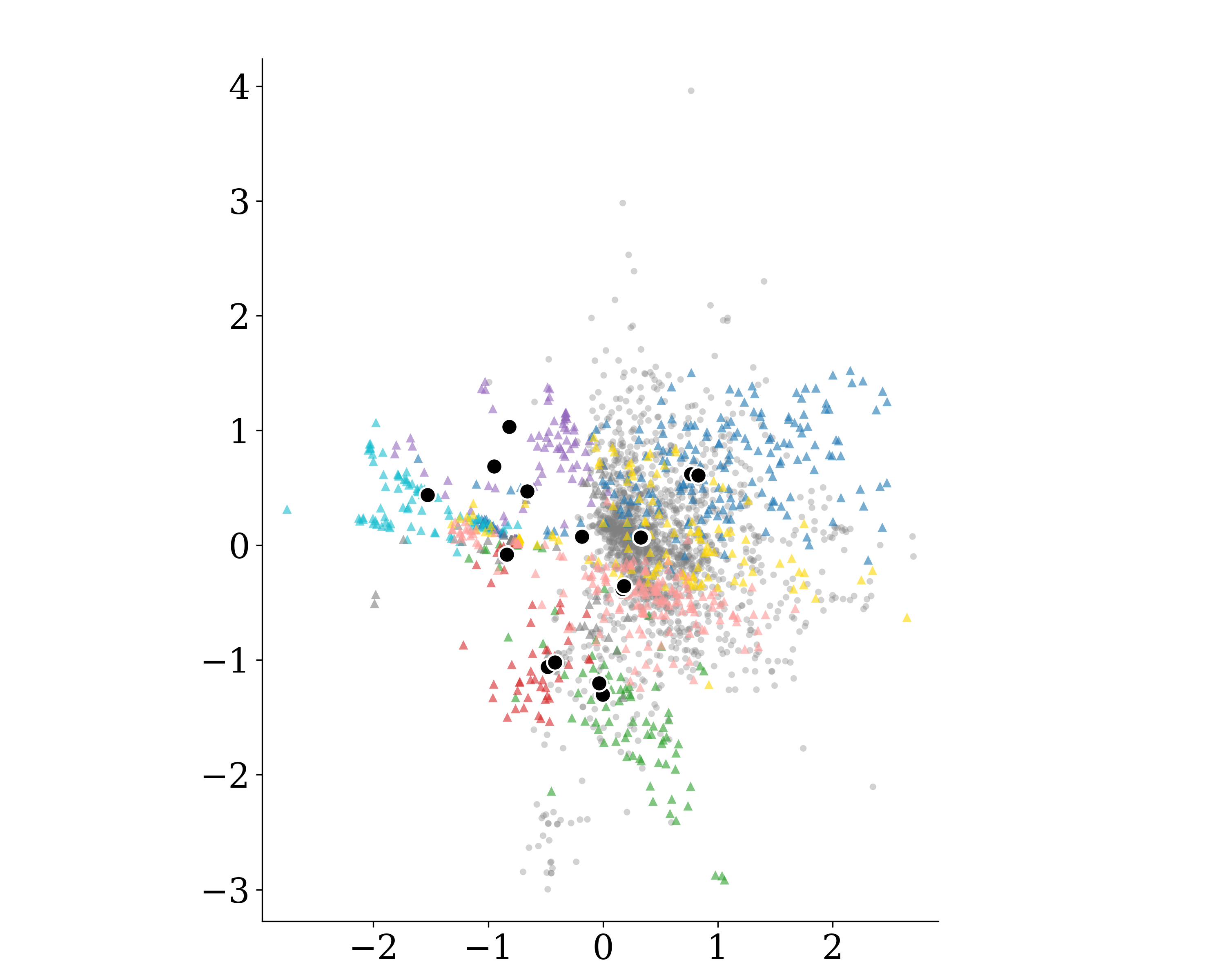}
    \end{subfigure}

    \caption{
    Two views of the recovered European Parliament latent configuration.
    Left: estimated embedding vectors of $867$ MEPs, colored by official
    political group affiliation. Right: joint representation of MEPs and
    legislative procedures in the recovered bipartite configuration.
    }
    \label{fig:euro_joint}
\end{figure}

\begin{table}[!htb]
\centering
\small
\renewcommand{\arraystretch}{1.08}

\begin{tabularx}{\textwidth}{
    >{\raggedright\arraybackslash}X
    >{\centering\arraybackslash}X
    >{\centering\arraybackslash}X
    >{\centering\arraybackslash}X
    >{\centering\arraybackslash}X
}
\toprule
Group
& Dim 1 Mean
& Dim 1 Median
& Dim 2 Mean
& Dim 2 Median \\
\midrule

ID 
& -1.626 
& -1.690 
& 0.365 
& 0.306 \\

GUE\_NGL 
& -0.616 
& -0.614 
& -0.921 
& -1.038 \\

ECR 
& -0.599 
& -0.466 
& 0.776 
& 0.848 \\

NI 
& -0.556 
& -0.788 
& -0.030 
& 0.047 \\

GREEN\_EFA 
& 0.023 
& 0.118 
& -1.210 
& -1.300 \\

SD 
& 0.181 
& 0.336 
& -0.378 
& -0.396 \\

RENEW 
& 0.333 
& 0.416 
& 0.075 
& 0.048 \\

EPP 
& 0.794 
& 0.879 
& 0.624 
& 0.565 \\

\bottomrule
\end{tabularx}

\caption{Coordinate statistics of political groups on the first and second latent dimensions.}
\label{tab:party_coords}

\end{table}

As shown in Figure~\ref{fig:euro_joint} (left), the recovered latent configurations exhibit a strong alignment with the official political group affiliations of the MEPs, which is highly consistent with the well-known two-dimensional model of the European Parliament proposed in political science literature \citep{hix2006dimensions, greco2025twodimensional}. Moreover, the two resolved dimensions can be broadly interpreted in terms of attitudes toward European integration and traditional socio-economic ideologies. The first latent dimension appears to capture a cleavage related to attitudes toward European integration. The signs of the displayed axes are chosen for interpretability and have no effect on the underlying Euclidean distances. As shown in Figure~\ref{fig:euro_joint} (left) and summarized in Table~\ref{tab:party_coords}, EPP, RENEW, and SD are located on the positive side of this dimension, whereas ID, ECR, and GUE\_NGL are located on the negative side. This ordering is broadly consistent with the distinction between groups that are generally more supportive of European integration and groups that exhibit more critical attitudes toward supranational integration. The second latent dimension appears to capture the conventional left--right ideological cleavage in the European Parliament. 
As shown in Figure~\ref{fig:euro_joint} (left) and summarized in Table~\ref{tab:party_coords}, GREEN\_EFA and GUE\_NGL are located on the negative side of this dimension, followed by SD and RENEW near the center, whereas EPP, ECR, and ID occupy more positive positions. Importantly, this dimension is distinct from the European integration dimension: for example, ECR and GUE\_NGL are located at similar positions along the first dimension but differ substantially along the second dimension.

A key advantage of the bipartite framework is that MEPs and legislative procedures are represented jointly in the same Euclidean space. Figure~\ref{fig:euro_joint} (right) illustrates this joint representation, allowing legislative procedures to be interpreted relative to the locations of political groups in the recovered configuration. For illustration, for each political group we identify the two distinct roll-call issues closest to the centroid of its MEP embedding vectors. For example, the procedures nearest to the SD center concern humanitarian assistance in Gaza and the ILO Violence and Harassment Convention, while those near the GREEN\_EFA center involve the role of tax policy during crises and the establishment of the EU ethics body. The ECR center lies close to procedures concerning migration-related flexibility instruments and additional technical criteria for climate-related economic activities. The complete set of group--procedure associations is reported in Table~\ref{tab:supp_party_issues} in Appendix~\ref{app:additional-results}.

\section{Discussion}\label{sec:discuss}

This paper develops a computationally efficient framework for recovering Euclidean configurations from unipartite and bipartite relational
data under a general mean formulation. The proposed approach
separates low-rank signal estimation from geometric recovery, and in the
bipartite setting reduces the remaining ambiguity to a low-dimensional
alignment problem. The recovery step is modular and can be combined with
alternative sufficiently accurate signal estimators. The theory establishes
uniform configuration recovery for both settings and formulates bipartite
identifiability through local identification strength and global separation,
without requiring explicit anchor-point conditions.

Several directions merit further study. A natural extension is to
multipartite relational data with structured blockwise missingness, where
global identifiability may depend jointly on the observed block pattern
and the latent geometry. Such extensions would broaden the applicability
of the framework to complex relational systems involving multiple types of
entities and interactions.
Another direction is to develop statistically efficient first-stage
estimators for nonlinear-link models. Replacing the spectral estimator
with likelihood-based or other efficient estimators may improve the signal
estimation accuracy while retaining the modular configuration recovery
procedure developed here. This combination could provide a foundation for
establishing asymptotic distributions and uncertainty quantification for
recovered embeddings, pairwise distances, and other geometrically
interpretable quantities.
A further direction is to extend the framework to dynamic relational
systems, where latent configurations evolve over time. Incorporating
temporal dependence and developing scalable procedures for tracking
changes in latent geometry would enable applications to longitudinal
network and interaction data.

\bigskip
\appendix


\section{Additional Simulation and Real-Data Results}
\label{app:additional-results}

This appendix reports additional simulation and real-data results that
complement the numerical analyses in the main text.

\subsection{Additional Simulation Results for \texorpdfstring{$K=5$}{K=5}}

This section reports the simulation results for latent dimension $K=5$. The simulation settings and evaluation criteria are the same as those in Section~\ref{subsec:simulation-results} of the main text. Overall, the results are consistent with those for $K=3$ and exhibit similar patterns as the matrix size increases.

\begin{figure}[!htb]
    \centering

    \includegraphics[width=0.96\textwidth]
    {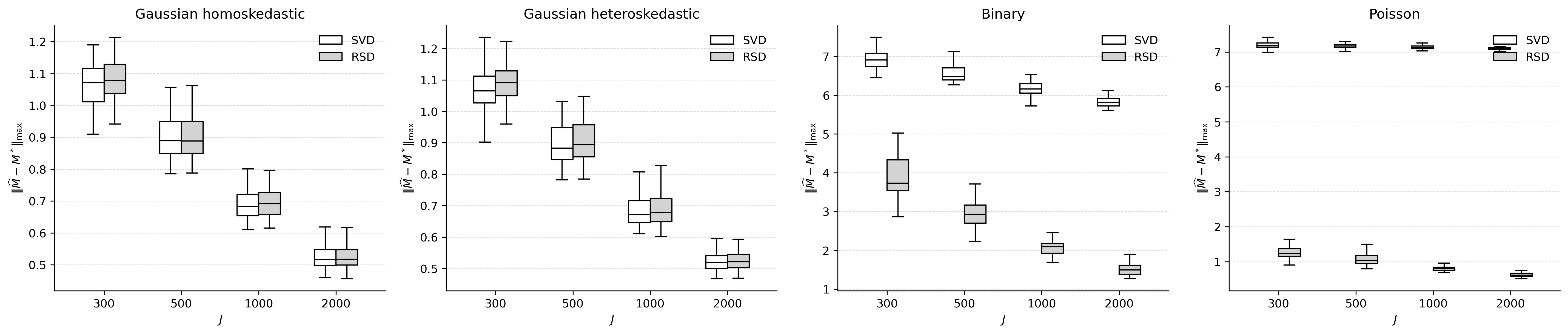}

    \caption{Max-norm errors of $\widehat\MM^{\mathrm{SVD}}$ and
$\widehat\MM^{\mathrm{RSD}}$ for the unipartite model with $K=5$.
    }
    \label{fig:uni_k5-entrywise}
\end{figure}

\begin{figure}[!htb]
    \centering

    \includegraphics[
        width=0.94\textwidth,
        height=0.25\textheight,
        keepaspectratio
    ]
    {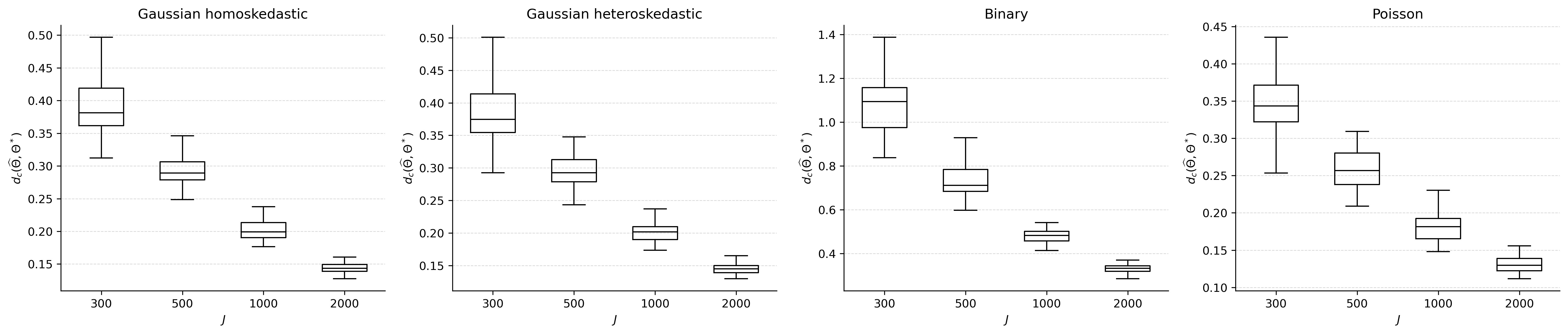}

    \caption{
        Configuration recovery error $d_c(\widehat\TTT,\TTT^*)$
        for the unipartite model with $K=5$.
    }
    \label{fig:uni_k5-recovery}
\end{figure}

\begin{figure}[!htb]
    \centering

    \includegraphics[width=0.96\textwidth]
    {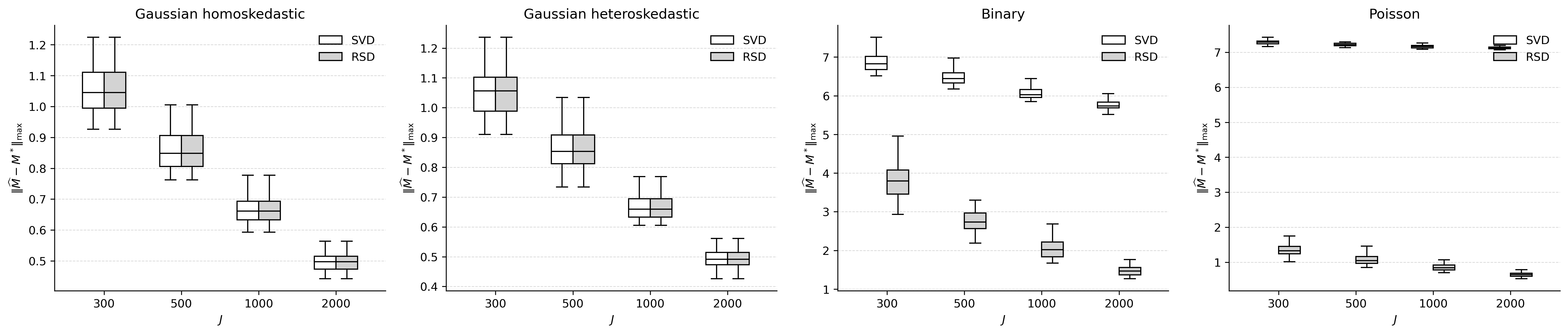}

    \caption{
        Max-norm errors of $\widehat\MM^{\mathrm{SVD}}$ and
$\widehat\MM^{\mathrm{RSD}}$ for the bipartite model with $K=5$.
    }
    \label{fig:bi_k5-entrywise}
\end{figure}

\begin{figure}[!htb]
    \centering

    \begin{subfigure}[t]{0.94\textwidth}
        \centering
        \includegraphics[
            width=\linewidth,
            height=0.21\textheight,
            keepaspectratio
        ]
        {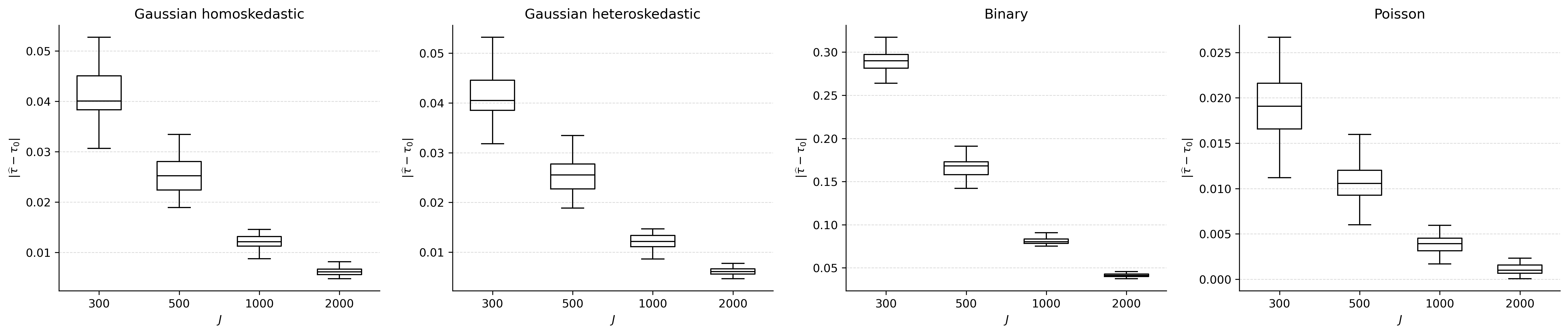}
        \caption{
            $|\widehat{\tau}-\tau_0|$.
        }
        \label{fig:bi_k5-tau}
    \end{subfigure}

    \vspace{0.4em}

    \begin{subfigure}[t]{0.94\textwidth}
        \centering
        \includegraphics[
            width=\linewidth,
            height=0.21\textheight,
            keepaspectratio
        ]
        {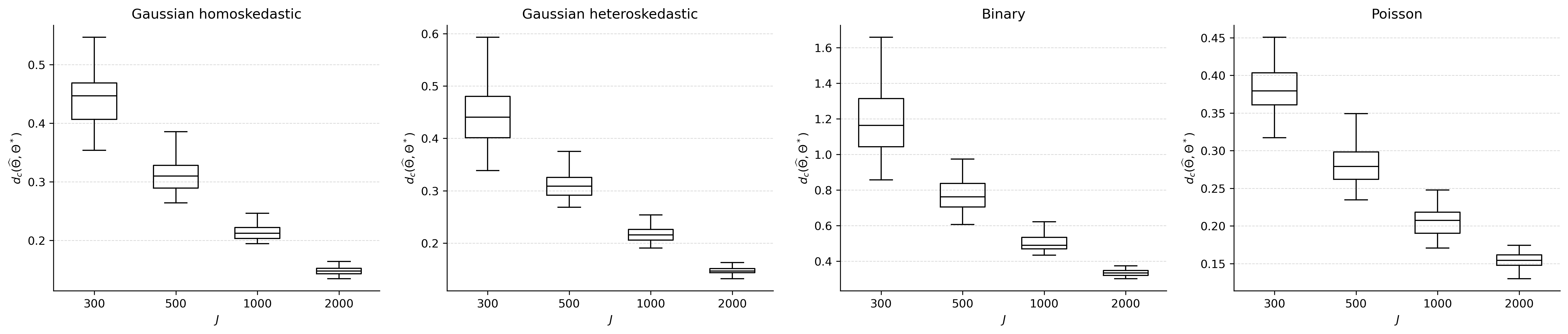}
        \caption{
            $d_c(\widehat\TTT,\TTT^*)$.
        }
        \label{fig:bi_k5-p1-configuration}
    \end{subfigure}

    \caption{
        $|\widehat{\tau}-\tau_0|$ and $d_c(\widehat\TTT,\TTT^*)$ for the
        bipartite model with $K=5$. 
    }
    \label{fig:bi_k5-recovery}
\end{figure}

Figures~\ref{fig:uni_k5-entrywise}--\ref{fig:uni_k5-recovery} show the results for the unipartite setting, while Figures~\ref{fig:bi_k5-entrywise}--\ref{fig:bi_k5-recovery} show the corresponding results for the bipartite setting, together with the drift parameter error.

The results for $K=5$ broadly agree with those for $K=3$ in the main text. Across the observation models, the signal-estimation and configuration recovery errors generally decrease as the matrix size increases.
The refinement is particularly effective in reducing the entrywise maximum error for the nonlinear observation models. For the bipartite setting, the drift parameter error also generally decreases with the matrix size.
These results exhibit the same overall finite-sample patterns as those reported in the main text.

\subsection{Additional Results for the European Parliament Voting Data}

Table~\ref{tab:supp_party_issues} reports the two legislative
procedures closest to the centroid of each political group in the estimated
joint configuration. These results complement
Figure~\ref{fig:euro_joint} (right) in the main text by providing the
complete list of group--procedure proximities used in the visualization.
As discussed in the main text, these associations are descriptive and
should be interpreted in terms of proximity in the recovered voting geometry
rather than as direct measures of issue importance or uniform political
support.

\begin{table}[!htb]
\centering
\small
\renewcommand{\arraystretch}{1.08}

\begin{tabularx}{\textwidth}{
>{\raggedright\arraybackslash}p{3.0cm}
>{\scriptsize\raggedright\arraybackslash}X
>{\scriptsize\raggedright\arraybackslash}X
}
\toprule
\textbf{Political Group}
& \textbf{Nearest Legislative Procedure}
& \textbf{Second-Nearest Legislative Procedure} \\
\midrule

ECR
& Flexibility Instrument: adjustments to the amounts mobilised for 2019 to be used for migration, refugee inflows and security threats
& Additional technical screening criteria for climate change mitigation and adaptation activities \\

EPP
& The EU Trade Policy Review
& Organic production: date of application and certain other dates \\

GREEN\_EFA
& Role of tax policy in times of crisis
& The establishment of the EU ethics body \\

GUE\_NGL
& Renewal of the authorisation for products containing genetically modified soybean A5547-127
& Renewal of the authorisation for products containing genetically modified cotton 281-24-236 $\times$ 3006-210-23 \\

ID
& Acceptance period of vaccination certificates issued in the EU Digital COVID Certificate format
& Children's rights on the occasion of the 30th anniversary of the UN Convention on the Rights of the Child \\

NI
& The deteriorating situation of human rights in Egypt, including the case of EIPR activists
& Objection concerning maximum residue levels for tricyclazole \\

RENEW
& Exceptional macro-financial assistance to Ukraine and reinforcement of the Common Provisioning Fund
& Future of fisheries in the Channel, North Sea, Irish Sea and Atlantic Ocean \\

SD
& The immediate risk of mass starvation in Gaza and attacks on humanitarian aid deliveries
& Council decision inviting Member States to ratify the Violence and Harassment Convention, 2019 (No 190) of the International Labour Organization \\

\bottomrule
\end{tabularx}

\caption{Top-2 closest legislative procedures to each political group center.}
\label{tab:supp_party_issues}
\end{table}

\section{Proofs for Spectral Denoising}
\label{app:spectral-denoising-proofs}

The proof of Theorem~\ref{thm:converge in prob} is organized by data structure and link function. We first consider bipartite data and then give the corresponding arguments for unipartite data, where symmetry and the unobserved diagonal require additional treatment. For the identity link, we establish both the normalized Frobenius and max-norm bounds, whereas for nonlinear links we establish the normalized Frobenius bound used in the subsequent refinement analysis.

Throughout Appendix~\ref{app:spectral-denoising-proofs}, let $C_0>0$ denote a fixed constant such that Assumption~\ref{assumption:algo1_bounded} holds, in particular, $\|\MM^*\|_{\max}\le C_0$. To distinguish this population bound from the projection constant used in Algorithm~\ref{alg:rsd}, we denote the latter by $C_{\mathrm{proj}}$ throughout the proof and take $C_{\mathrm{proj}}\ge C_0$.

\subsection{Bipartite Spectral Denoising}

We treat the identity and nonlinear link functions separately.

\subsubsection{Identity Link}

In the identity-link case, $f(x)=x$, hence
$\YY=\MM^*+\EEEE$, where
$\EEEE=(y_{ij}-m_{ij}^*)_{N\times J}$ is the noise matrix.
Algorithm~\ref{alg:rsd} computes
$\widehat\XX=\SVD_{K+2}(\YY)$ and outputs
$\widehat\MM^{\SVD}=(\widehat m_{ij}^{\SVD})_{N\times J}$ with
\[
\widehat m_{ij}^{\SVD}
=
\PPP_{[-C_{\mathrm{proj}},C_{\mathrm{proj}}]}
(\widehat x_{ij}).
\]

We first establish the normalized Frobenius bound. By the
Eckart--Young--Mirsky theorem,
$\widehat\XX=\argmin_{\rank(\XX)\le K+2}\|\YY-\XX\|_F^2$.
Since $\rank(\MM^*)\le K+2$,
$\|\YY-\widehat\XX\|_F^2\le\|\YY-\MM^*\|_F^2$.
Let $\Delta=\widehat\XX-\MM^*$. Since
$\YY-\widehat\XX=\EEEE-\Delta$,
\[
\|\Delta\|_F^2
\le
2\langle\EEEE,\Delta\rangle
\le
2\|\EEEE\|_2\|\Delta\|_*
\le
2\sqrt{2(K+2)}\|\EEEE\|_2\|\Delta\|_F,
\]
where
$\rank(\Delta)\le2(K+2)$.
Consequently,
\begin{equation}\label{eq:id_X_bound_revised}
\|\widehat\XX-\MM^*\|_F
\le
2\sqrt{2(K+2)}\,\|\EEEE\|_2.
\end{equation}

Since
$m_{ij}^*\in[-C_0,C_0]\subseteq
[-C_{\mathrm{proj}},C_{\mathrm{proj}}]$,
entrywise projection cannot increase the error relative to $\MM^*$.
Thus,
\[
\|\widehat\MM^{\SVD}-\MM^*\|_F
\le
\|\widehat\XX-\MM^*\|_F,
\]
and hence
\begin{equation}\label{eq:id_M_bound_revised}
\frac{1}{NJ}
\|\widehat\MM^{\SVD}-\MM^*\|_F^2
\le
\frac{8(K+2)}{NJ}\|\EEEE\|_2^2.
\end{equation}

Conditional on the true latent configuration, the entries of $\EEEE$
are independent, mean zero, and uniformly sub-exponential under
Assumption~\ref{assumption:algo1_subexponential}. Therefore, by the
rectangular analogue of Corollary~3.5 of
\citet{bandeira2016sharp}, obtained from the rectangular argument of
their Theorem~3.1,
\[
\EEE\|\EEEE\|_2
\lesssim
\sqrt N+\sqrt J+\log J
\lesssim
\sqrt N,
\]
where $J\le N$. Hence $\|\EEEE\|_2=O_p(\sqrt N)$, and
\[
\frac{1}{NJ}
\|\widehat\MM^{\SVD}-\MM^*\|_F^2
=
O_p(J^{-1}).
\]

We next establish the max-norm bound under
Assumptions~\ref{assumption:algo2_incoherence}
and~\ref{assumption:algo2_spikes}. Put $r=K+2$ and
$L=\log(NJ)$, and write
\[
\MM^*=\mathbf U\mathbf D\mathbf V^\top,
\qquad
\widehat\XX=\SVD_r(\YY),
\]
where $\mathbf U\in\R^{N\times r}$ and
$\mathbf V\in\R^{J\times r}$ have orthonormal columns.
The rank bound together with
Assumption~\ref{assumption:algo2_spikes} implies
$\rank(\MM^*)=r$, and
\begin{equation}\label{eq:idmax_population}
\begin{gathered}
\sigma_r(\MM^*)\gtrsim\sqrt{NJ},
\qquad
\|\MM^*\|\lesssim\sqrt{NJ},
\\
\|\mathbf U\|_{2\to\infty}\lesssim N^{-1/2},
\qquad
\|\mathbf V\|_{2\to\infty}\lesssim J^{-1/2},
\\
\|\MM^*\|_{2\to\infty}\lesssim\sqrt J,
\qquad
\|\MM^{*\top}\|_{2\to\infty}\lesssim\sqrt N.
\end{gathered}
\end{equation}
Again, entrywise projection cannot increase the error, so it suffices
to control $\|\widehat\XX-\MM^*\|_{\max}$.

Let $\mathbf E=\YY-\MM^*$, let $\mathbf V_Y$ contain the leading $r$
right singular vectors of $\YY$, and define
\[
\mathbf P=\mathbf V\mathbf V^\top,
\qquad
\widehat{\mathbf P}=\mathbf V_Y\mathbf V_Y^\top,
\qquad
\mathbf H=\YY^\top\YY-\MM^{*\top}\MM^*.
\]
Writing
$\YY^\top\YY\mathbf V_Y
=\mathbf V_Y\boldsymbol\Lambda_Y$
and
$\lambda_Y=\lambda_{\min}(\boldsymbol\Lambda_Y)=\sigma_r(\YY)^2$,
we have, whenever $\lambda_Y>0$,
\begin{equation}\label{eq:idmax_offspace}
(\mathbf I_J-\mathbf P)\mathbf V_Y
=
(\mathbf I_J-\mathbf P)
\mathbf H\mathbf V_Y
\boldsymbol\Lambda_Y^{-1}.
\end{equation}
Consequently,
\[
\|(\mathbf I_J-\mathbf P)\mathbf V_Y\|
\le
\frac{\|\mathbf H\|}{\lambda_Y}.
\]
Moreover, for each $j\in[J]$,
\[
\|\mathbf e_j^\top
(\mathbf I_J-\mathbf P)\mathbf V_Y\|
\le
\lambda_Y^{-1}
\left\{
\|\mathbf e_j^\top\mathbf H\|
+
\|\mathbf e_j^\top\mathbf V\|\,\|\mathbf H\|
\right\}.
\]
Since right multiplication by $\mathbf V_Y^\top$ preserves the
Euclidean norm of vectors in $\R^r$,
\begin{equation}\label{eq:idmax_offspace_rows}
\|(\mathbf I_J-\mathbf P)\widehat{\mathbf P}\|_{2\to\infty}
\le
\frac{
\|\mathbf H\|_{2\to\infty}
+
\|\mathbf V\|_{2\to\infty}\|\mathbf H\|
}{
\lambda_Y
}.
\end{equation}

Let $\mathbf C_Y=\mathbf V^\top\mathbf V_Y$. Orthogonality gives
\[
\|\mathbf V^\top
(\mathbf I_J-\widehat{\mathbf P})\|^2
=
1-\sigma_r^2(\mathbf C_Y)
=
\|(\mathbf I_J-\mathbf P)\mathbf V_Y\|^2,
\]
and hence
\[
\|\mathbf P(\mathbf I_J-\widehat{\mathbf P})\|_{2\to\infty}
\le
\|\mathbf V\|_{2\to\infty}
\frac{\|\mathbf H\|}{\lambda_Y}.
\]
Using
$\widehat{\mathbf P}-\mathbf P
=(\mathbf I_J-\mathbf P)\widehat{\mathbf P}
-\mathbf P(\mathbf I_J-\widehat{\mathbf P})$,
we obtain
\begin{equation}\label{eq:idmax_projector_deterministic}
\|\widehat{\mathbf P}-\mathbf P\|_{2\to\infty}
\le
\frac{
\|\mathbf H\|_{2\to\infty}
+
2\|\mathbf V\|_{2\to\infty}\|\mathbf H\|
}{
\lambda_Y
}.
\end{equation}

Since
$\widehat\XX=\YY\widehat{\mathbf P}$
and $\MM^*\mathbf P=\MM^*$,
\begin{equation}\label{eq:idmax_reconstruction}
\widehat\XX-\MM^*
=
\mathbf E\mathbf P
+
\YY(\widehat{\mathbf P}-\mathbf P).
\end{equation}
Therefore,
\begin{equation}\label{eq:idmax_reconstruction_bound}
\|\widehat\XX-\MM^*\|_{\max}
\le
\|\mathbf E\mathbf V\|_{2\to\infty}
\|\mathbf V\|_{2\to\infty}
+
\|\YY\|_{2\to\infty}
\|\widehat{\mathbf P}-\mathbf P\|_{2\to\infty}.
\end{equation}

The preceding sub-exponential assumption gives
\begin{equation}\label{eq:idmax_noise_operator}
\|\mathbf E\|=O_p(\sqrt N),
\end{equation}
and a truncation argument followed by Bernstein's inequality and a
union bound gives
\begin{equation}\label{eq:idmax_noise_rows}
\|\mathbf E\|_{2\to\infty}=O_p(\sqrt J),
\qquad
\|\YY\|_{2\to\infty}=O_p(\sqrt J).
\end{equation}
Using the incoherence of $\mathbf U$ and $\mathbf V$ together with the
weighted sub-exponential Bernstein inequality
\citep[Theorem~2.8.2]{vershynin2018high}, we further obtain
\begin{equation}\label{eq:idmax_projected_noise}
\|\mathbf E\mathbf V\|_{2\to\infty}
=
O_p(\sqrt L),
\qquad
\|\mathbf E^\top\mathbf U\|_{2\to\infty}
=
O_p(\sqrt L).
\end{equation}

By \eqref{eq:idmax_population} and
\eqref{eq:idmax_noise_operator},
$\|\mathbf E\|/\sigma_r(\MM^*)=O_p(J^{-1/2})=o_p(1)$.
Thus, with probability tending to one,
$\sigma_r(\YY)\gtrsim\sqrt{NJ}$ and
$\lambda_Y^{-1}\lesssim(NJ)^{-1}$.
Since
\[
\mathbf H
=
\MM^{*\top}\mathbf E
+
\mathbf E^\top\MM^*
+
\mathbf E^\top\mathbf E,
\]
we have
\begin{equation}\label{eq:idmax_gram_operator}
\|\mathbf H\|
=
O_p(N\sqrt J),
\end{equation}
and, using
$\MM^*=\mathbf U\mathbf D\mathbf V^\top$,
\begin{equation}\label{eq:idmax_gram_rows}
\|\mathbf H\|_{2\to\infty}
=
O_p\left(
N+\sqrt{NJL}
\right).
\end{equation}
Substituting these bounds into
\eqref{eq:idmax_projector_deterministic} yields
\begin{equation}\label{eq:idmax_projector_rate}
\|\widehat{\mathbf P}-\mathbf P\|_{2\to\infty}
=
O_p\left(
\frac1J+\sqrt{\frac{L}{NJ}}
\right).
\end{equation}
Combining
\eqref{eq:idmax_reconstruction_bound},
\eqref{eq:idmax_noise_rows},
\eqref{eq:idmax_projected_noise}, and
\eqref{eq:idmax_projector_rate}, we obtain
\[
\begin{aligned}
\|\widehat\XX-\MM^*\|_{\max}
&=
O_p\left\{
\sqrt{\frac LJ}
+
\sqrt J
\left(
\frac1J+\sqrt{\frac{L}{NJ}}
\right)
\right\}\\
&=
O_p\left(
\sqrt{\frac LJ}
\right),
\end{aligned}
\]
where the last equality uses $N\ge J$.
Finally, entrywise projection gives
\[
\|\widehat\MM^{\mathrm{SVD}}-\MM^*\|_{\max}
=
O_p\left(
\sqrt{\frac{\log(NJ)}{J}}
\right).
\]
This proves the identity-link bounds in the bipartite setting.

\subsubsection{Nonlinear Link}

Let
$$
\YY^*
=
f(\MM^*)
=
(y_{ij}^*)_{N\times J},
\qquad
y_{ij}^*
=
f(m_{ij}^*).
$$
Define the truncated observation matrix
$\widehat\YY=(\widehat y_{ij})_{N\times J}$ with
$\widehat y_{ij} = \PPP_{[a_{NJ},b_{NJ}]}(y_{ij}).$
We first decompose the estimation error as
\begin{equation}\label{eq:bias_variance_split}
\frac{1}{NJ}
\EEE\|\widehat\XX-\YY^*\|_F^2
\le
\frac{2}{NJ}
\EEE\|
\widehat\XX-\EEE(\widehat\YY)
\|_F^2
+
\frac{2}{NJ}
\|
\EEE(\widehat\YY)-\YY^*
\|_F^2.
\end{equation}
By the definition of $r_{NJ}$,
$| \EEE\widehat y_{ij} - y_{ij}^* | \le r_{NJ},$
and hence
\begin{equation}\label{eq:bias_pure}
\frac{2}{NJ}
\|
\EEE(\widehat\YY)-\YY^*
\|_F^2
\le
2r_{NJ}^2.
\end{equation}

It remains to bound the first term in
\eqref{eq:bias_variance_split}. Let
$$
\widetilde r_0
:=
\max
\left\{
k:
\sigma_k(\widetilde\YY)
\ge
1.01\sqrt N
\right\},
$$
with $\widetilde r_0=0$ if the above set is empty, and define the
pure-threshold estimator
$$
\widetilde\XX^{\mathrm{USVT}}
:=
\SVD_{\widetilde r_0}(\widetilde\YY).
$$
Since $\widetilde y_{ij}\in[0,1]$,
$\operatorname{Var}(\widetilde y_{ij})\le1/4$.
Conditional on the true latent configuration, the entries of
$\widetilde\YY$ are independent in the bipartite setting.
Since $N\ge J$, we apply the variance-adjusted rectangular USVT argument
to $\widetilde\YY^\top\in\R^{J\times N}$.
Taking the fixed slack parameter so that the threshold equals
$1.01\sqrt N$, the argument underlying Theorem~1.1 of
\citet{chatterjee2015matrix} yields
\begin{equation}\label{eq:USVT_bound}
\frac{1}{NJ}
\EEE
\left\|
\widetilde\XX^{\mathrm{USVT}}
-
\EEE(\widetilde\YY)
\right\|_F^2
\le
c
\min
\left\{
\frac{\|\EEE(\widetilde\YY)\|_*}{J\sqrt N},
\frac{\|\EEE(\widetilde\YY)\|_*^2}{NJ},
1
\right\}
+
ce^{-cN}.
\end{equation}

Our estimator instead uses
$\widetilde r=\max\{K+2,\widetilde r_0\}$ and
$\widetilde\XX=\SVD_{\widetilde r}(\widetilde\YY)$.
If $\widetilde r_0\ge K+2$, then
$\widetilde\XX=\widetilde\XX^{\mathrm{USVT}}$.
Otherwise, all additionally retained singular values are smaller than
$1.01\sqrt N$, and hence
$$
\left\|
\widetilde\XX-\widetilde\XX^{\mathrm{USVT}}
\right\|_F^2
\le
(K+2)(1.01)^2N
\le
cN.
$$
Using
$\|A+B\|_F^2\le2\|A\|_F^2+2\|B\|_F^2$
and retaining only the first term in the minimum in
\eqref{eq:USVT_bound}, we obtain
\begin{equation}\label{eq:modified_USVT_bound}
\begin{aligned}
\frac{1}{NJ}
\EEE
\left\|
\widetilde\XX-\EEE(\widetilde\YY)
\right\|_F^2
\le\;&
c
\frac{\|\EEE(\widetilde\YY)\|_*}{J\sqrt N}
+
ce^{-cN}
+
\frac{c}{J}.
\end{aligned}
\end{equation}

Given the rescaling relations
$$
\EEE(\widehat y_{ij})
=
(b_{NJ}-a_{NJ})
\EEE(\widetilde y_{ij})
+
a_{NJ},
\qquad
\widehat x_{ij}
=
(b_{NJ}-a_{NJ})
\widetilde x_{ij}
+
a_{NJ},
$$
we have
$$
\|
\widehat\XX-\EEE(\widehat\YY)
\|_F^2
=
(b_{NJ}-a_{NJ})^2
\|
\widetilde\XX-\EEE(\widetilde\YY)
\|_F^2.
$$
Moreover,
$$
\EEE\widetilde\YY
=
\frac{1}{b_{NJ}-a_{NJ}}
\left\{
\EEE\widehat\YY
-
a_{NJ}\EE_N\EE_J^\top
\right\}.
$$
Substituting these relations into
\eqref{eq:modified_USVT_bound} gives
\begin{equation}\label{eq:variance_bound_1}
\begin{aligned}
\frac{1}{NJ}
\EEE
\left(
\|
\widehat\XX-\EEE(\widehat\YY)
\|_F^2
\right)
\le\;&
c(b_{NJ}-a_{NJ})^2
\frac{
\|\EEE\widetilde\YY\|_*
}{
J\sqrt N
}
+
c(b_{NJ}-a_{NJ})^2e^{-cN}
+
c\frac{(b_{NJ}-a_{NJ})^2}{J}\\
=\;&
c
\frac{b_{NJ}-a_{NJ}}{J\sqrt N}
\|
\EEE\widehat\YY
-
a_{NJ}\EE_N\EE_J^\top
\|_*\\
&+
c(b_{NJ}-a_{NJ})^2e^{-cN}
+
c\frac{(b_{NJ}-a_{NJ})^2}{J}.
\end{aligned}
\end{equation}

To bound
$\|
\EEE\widehat\YY-a_{NJ}\EE_N\EE_J^\top
\|_*$,
we construct configuration-dependent $\delta$-nets by selecting centers
from the true embedding vectors. Specifically, let
$$
\mathcal G_1
\subset
\{\ttt_i^*:i\in[N]\},
\qquad
\mathcal G_2
\subset
\{\ttt_{N+j}^*:j\in[J]\}
$$
be maximal $\delta$-separated subsets. Then
$\mathcal G_1$ and $\mathcal G_2$ are $\delta$-nets of the two
configurations
$\TTT_{\mathcal P_1}^*$ and
$\TTT_{\mathcal P_2}^*$, respectively, and every net center coincides
with a true embedding vector.

For any $\ttt_i^*$ and $\ttt_{N+j}^*$, let
$p(\ttt_i^*)$ and $p(\ttt_{N+j}^*)$ be the corresponding nearest points
in $\mathcal G_1$ and $\mathcal G_2$. Under
Assumption~\ref{assumption:algo1_bounded}, the cardinalities satisfy
$|\mathcal G_1|, |\mathcal G_2| \le N_\delta := c(C_0/\delta)^K.$

We construct the approximation matrix
$\widehat\YY_\delta=(\widehat y_{ij}^\delta)_{N\times J}$ with entries
$$
\widehat y_{ij}^{\delta}
=
f
\left(
\tau_0
-
\|
p(\ttt_i^*)
-
p(\ttt_{N+j}^*)
\|^2
\right).
$$
Since $\widehat\YY_\delta$ is generated by at most $N_\delta$ distinct
row vectors, its rank is bounded by
$$
\rank(\widehat\YY_\delta)
\le
\min\{
|\mathcal G_1|,
|\mathcal G_2|
\}
\le
N_\delta.
$$

We decompose the target nuclear norm into two parts:
\begin{equation}\label{eq:decomp_nuclear}
\|
\EEE\widehat\YY
-
a_{NJ}\EE_N\EE_J^\top
\|_*
\le
\underbrace{
\|
\EEE\widehat\YY
-
\widehat\YY_\delta
\|_*
}_{\text{(I)}}
+
\underbrace{
\|
\widehat\YY_\delta
-
a_{NJ}\EE_N\EE_J^\top
\|_*
}_{\text{(II)}}.
\end{equation}

\textbf{Part (I):}

Define
$m_{ij}^{\delta} = \tau_0 - \| p(\ttt_i^*) - p(\ttt_{N+j}^*) \|^2.$
Since the centers of $\mathcal G_1$ and $\mathcal G_2$ are selected from
the true embedding vectors, for every $(i,j)$ there exist
$i'\in[N]$ and $j'\in[J]$ such that
$$
p(\ttt_i^*)
=
\ttt_{i'}^*,
\qquad
p(\ttt_{N+j}^*)
=
\ttt_{N+j'}^*,
$$
and hence
$$
m_{ij}^{\delta}
=
\tau_0
-
\|
\ttt_{i'}^*
-
\ttt_{N+j'}^*
\|^2
=
m_{i'j'}^*.
$$
Therefore,
$|m_{ij}^{\delta}|\le C_0$.
Hence, both $m_{ij}^*$ and $m_{ij}^{\delta}$ lie in $[-C_0,C_0]$.
By Assumption~\ref{assumption:algo2_link}, let
$L_f := \sup_{|m|\le C_0}|f'(m)| < \infty.$
Thus, $f$ is $L_f$-Lipschitz on $[-C_0,C_0]$. Then
\begin{equation}
\begin{aligned}
|
\EEE\widehat y_{ij}
-
\widehat y_{ij}^{\delta}
|
&\le
|
\EEE\widehat y_{ij}
-
f(m_{ij}^*)
|
+
|
f(m_{ij}^*)
-
\widehat y_{ij}^{\delta}
|\\
&\le
r_{NJ}
+
L_f
\left|
\|
\ttt_i^*
-
\ttt_{N+j}^*
\|^2
-
\|
p(\ttt_i^*)
-
p(\ttt_{N+j}^*)
\|^2
\right|\\
&\le
r_{NJ}
+
8C_0L_f\delta.
\end{aligned}
\end{equation}
Therefore,
\begin{equation}\label{ineq:I}
\begin{aligned}
\text{(I)}
&\le
\sqrt J
\|
\EEE\widehat\YY-\widehat\YY_\delta
\|_F\\
&\le
\sqrt J\sqrt{NJ}
(r_{NJ}+8C_0L_f\delta)\\
&\le
(r_{NJ}+c\delta)J\sqrt N.
\end{aligned}
\end{equation}

\textbf{Part (II):}

For fixed $i$, the $i$th row of $\widehat\YY_\delta$ depends on $i$ only
through $p(\ttt_i^*)$. Therefore,
\begin{equation}\label{eq:two_routes_rank_f}
\rank
\left(
\widehat\YY_\delta
-
a_{NJ}\EE_N\EE_J^\top
\right)
\le
N_\delta.
\end{equation}

Since the net centers are true embedding vectors, for every $(i,j)$ there
exist indices $i'\in[N]$ and $j'\in[J]$ such that
$m_{ij}^{\delta}=m_{i'j'}^*$.
Because
$\EEE\widehat y_{i'j'}\in[a_{NJ},b_{NJ}]$
and
$\left| \EEE\widehat y_{i'j'} - f(m_{i'j'}^*) \right| \le r_{NJ},$
we obtain
$$
\left|
\widehat y_{ij}^{\delta}
-
a_{NJ}
\right|
\le
b_{NJ}-a_{NJ}+r_{NJ}.
$$
Consequently,
\begin{equation}\label{ineq:II}
\begin{aligned}
\text{(II)}
&\le
\sqrt{N_\delta}
\left\|
\widehat\YY_\delta
-
a_{NJ}\EE_N\EE_J^\top
\right\|_F\\
&\le
c\delta^{-K/2}
\sqrt{NJ}
\left(
b_{NJ}-a_{NJ}+r_{NJ}
\right).
\end{aligned}
\end{equation}

Substituting \eqref{ineq:I} and \eqref{ineq:II} into
\eqref{eq:decomp_nuclear}, and then into
\eqref{eq:variance_bound_1}, we obtain
\begin{align*}
\frac{1}{NJ}
\EEE
\left(
\|
\widehat\XX-\EEE(\widehat\YY)
\|_F^2
\right)
\le\;&
cR_{NJ}(r_{NJ}+\delta)
+
cR_{NJ}(R_{NJ}+r_{NJ})
\frac{\delta^{-K/2}}{\sqrt J}\\
&+
cR_{NJ}^2e^{-cN}
+
c\frac{R_{NJ}^2}{J}.
\end{align*}
Combining this with \eqref{eq:bias_pure}, we obtain
$$
\begin{aligned}
\frac{1}{NJ}
\EEE
\|
\widehat\XX-\YY^*
\|_F^2
\le\;&
cR_{NJ}(r_{NJ}+\delta)
+
cR_{NJ}(R_{NJ}+r_{NJ})
\frac{\delta^{-K/2}}{\sqrt J}\\
&+
cr_{NJ}^2
+
cR_{NJ}^2e^{-cN}
+
c\frac{R_{NJ}^2}{J}.
\end{aligned}
$$

Choose
$$
\delta
=
\left\{
\frac{(R_{NJ}+r_{NJ})^2}{J}
\right\}^{\frac{1}{K+2}}.
$$
Under the assumption
$R_{NJ}+r_{NJ}=o(\sqrt J)$,
we have $\delta=o(1)$. Substituting this choice and absorbing the
remainder terms gives
\begin{equation}\label{eq:X_rate}
\frac{1}{NJ}
\EEE
\|
\widehat\XX-\YY^*
\|_F^2
\le
cR_{NJ}
\left\{
\frac{(R_{NJ}+r_{NJ})^2}{J}
\right\}^{\frac{1}{K+2}}
+
cr_{NJ}(R_{NJ}+r_{NJ}).
\end{equation}

Since
$m_{ij}^*\in[-C_0,C_0]
\subseteq[-C_{\mathrm{proj}},C_{\mathrm{proj}}]$,
we have
$$
y_{ij}^*
=
f(m_{ij}^*)
\in
[f(-C_{\mathrm{proj}}),f(C_{\mathrm{proj}})].
$$
Hence,
\begin{align*}
|
\widehat m_{ij}^{\SVD}-m_{ij}^*
|
&\le
\left|
f^{-1}
\left(
\PPP_{[f(-C_{\mathrm{proj}}),f(C_{\mathrm{proj}})]}
(\widehat x_{ij})
\right)
-
f^{-1}(y_{ij}^*)
\right|\\
&\le
\ell^\dagger
|
\widehat x_{ij}-y_{ij}^*
|,
\end{align*}
where
$$
\ell^\dagger
:=
\left\{
\inf_{|m|\le C_{\mathrm{proj}}}
f'(m)
\right\}^{-1}
<
\infty.
$$
Therefore,
$$
\|
\widehat\MM^{\SVD}-\MM^*
\|_F^2
\le
(\ell^\dagger)^2
\|
\widehat\XX-\YY^*
\|_F^2.
$$
Combining this inequality with \eqref{eq:X_rate} and applying Markov's
inequality yields
$$
\frac{1}{NJ}
\|
\widehat\MM^{\SVD}-\MM^*
\|_F^2
=
O_p
\left(
R_{NJ}
\left\{
\frac{(R_{NJ}+r_{NJ})^2}{J}
\right\}^{\frac{1}{K+2}}
+
r_{NJ}(R_{NJ}+r_{NJ})
\right).
$$

\subsection{Unipartite Spectral Denoising}

For unipartite data, we set $J=N$. The main arguments from the
bipartite setting continue to apply, but the observation matrix is
symmetric, only the upper-triangular observations are mutually
independent, and the diagonal entries are not observed. We record the
required modifications separately.

\subsubsection{Identity Link}

For unipartite data, let
$\EEEE_{ij}=y_{ij}-m_{ij}^*$ for $i\neq j$, and define
$\WW=(w_{ij})_{N\times N}$ by
\[
w_{ij}
=
\begin{cases}
\EEEE_{ij}, & i<j,\\
\EEEE_{ji}, & i>j,\\
0, & i=j.
\end{cases}
\]
Recall that $y_{ii}=0$ is not treated as an observation, while the
algebraically completed signal satisfies $m_{ii}^*=\tau_0$. Therefore,
\[
\YY
=
\MM^*+\WW-\tau_0\II_N.
\]
By Assumption~\ref{assumption:algo1_bounded},
$|\tau_0|\le C_0$. Conditional on the true latent configuration, the
upper-triangular entries $\{w_{ij}:i<j\}$ are independent, mean zero,
and uniformly sub-exponential. Hence
$\|\WW\|_2=O_p(\sqrt N)$ and therefore
$\|\WW-\tau_0\II_N\|_2=O_p(\sqrt N)$.
Repeating the preceding Eckart--Young argument and using
$C_{\mathrm{proj}}\ge C_0$ gives
\[
\frac{1}{N^2}
\|\widehat\MM^{\SVD}-\MM^*\|_F^2
=
O_p(N^{-1}),
\]
which is $O_p(J^{-1})$ since $J=N$.

We next establish the max-norm bound. Put $r=K+2$ and write
\[
\MM^*=\mathbf U\mathbf D\mathbf V^\top,
\qquad
\widehat\XX=\SVD_r(\YY).
\]
The deterministic relations
\eqref{eq:idmax_offspace}--\eqref{eq:idmax_reconstruction_bound}
remain valid because they do not require independence of the entries
of the perturbation matrix. Set
\[
\mathbf E
=
\YY-\MM^*
=
\mathbf W-\tau_0\mathbf I_N,
\qquad
L=\log(N^2).
\]
Let $\mathbf W_+$ denote the strictly upper-triangular part of
$\mathbf W$. Its entries, including the deterministic zeros, are
independent and have uniformly bounded second and fourth moments.
By Theorem~2 of \citet{latala2005estimates},
\[
\EEE\|\mathbf W_+\|
\lesssim
\sqrt N.
\]
Since $\mathbf W=\mathbf W_++\mathbf W_+^\top$,
\[
\|\mathbf E\|
=
O_p(\sqrt N),
\qquad
\|\mathbf E\|_{2\to\infty}
=
O_p(\sqrt N),
\qquad
\|\YY\|_{2\to\infty}
=
O_p(\sqrt N).
\]

For each fixed $i\in[N]$ and $k\in[r]$,
\[
(\mathbf W\mathbf V)_{ik}
=
\sum_{j\ne i}W_{ij}V_{jk}.
\]
The variables in this sum correspond to distinct unordered pairs and
are therefore independent. Using the incoherence of $\mathbf V$ and
the weighted Bernstein inequality, and proceeding analogously for
$\mathbf W^\top\mathbf U$, gives
\[
\|\mathbf W\mathbf V\|_{2\to\infty}
+
\|\mathbf W^\top\mathbf U\|_{2\to\infty}
=
O_p(\sqrt L).
\]
The deterministic diagonal correction satisfies
\[
\|\tau_0\mathbf V\|_{2\to\infty}
+
\|\tau_0\mathbf U\|_{2\to\infty}
=
O(N^{-1/2}),
\]
so
\begin{equation}\label{eq:idmax_uni_projected_noise}
\|\mathbf E\mathbf V\|_{2\to\infty}
+
\|\mathbf E^\top\mathbf U\|_{2\to\infty}
=
O_p(\sqrt L).
\end{equation}

Since
$\sigma_r(\MM^*)\gtrsim N$
and
$\|\mathbf E\|=O_p(\sqrt N)$,
we have, with probability tending to one,
$\lambda_Y\gtrsim N^2$.
Moreover,
\[
\begin{aligned}
\|\mathbf H\|
&\le
2\|\MM^*\|\|\mathbf E\|+\|\mathbf E\|^2
=
O_p(N^{3/2}),\\
\|\mathbf H\|_{2\to\infty}
&\le
\|\MM^{*\top}\|_{2\to\infty}\|\mathbf E\|
+
\|\mathbf E^\top\mathbf U\|_{2\to\infty}\|\mathbf D\|
+
\|\mathbf E\|^2\\
&=
O_p(N+N\sqrt L).
\end{aligned}
\]
Thus, by \eqref{eq:idmax_projector_deterministic},
\[
\|\widehat{\mathbf P}-\mathbf P\|_{2\to\infty}
=
O_p\left(
\frac{1+\sqrt L}{N}
\right).
\]
Combining this with
\eqref{eq:idmax_reconstruction_bound} and
\eqref{eq:idmax_uni_projected_noise} yields
\[
\begin{aligned}
\|\widehat\XX-\MM^*\|_{\max}
&=
O_p\left\{
\sqrt{\frac LN}
+
\sqrt N\,
\frac{1+\sqrt L}{N}
\right\}\\
&=
O_p\left(
\sqrt{\frac LN}
\right).
\end{aligned}
\]
Since entrywise projection cannot increase the error relative to
$\MM^*$,
\[
\|\widehat\MM^{\mathrm{SVD}}-\MM^*\|_{\max}
=
O_p\left(
\sqrt{\frac{\log N}{N}}
\right)
=
O_p\left(
\sqrt{\frac{\log(NJ)}{J}}
\right),
\]
where $J=N$. This proves the identity-link bounds in the unipartite
setting.

\subsubsection{Nonlinear Link}

We now consider the unipartite case, where $J=N$,
$y_{ji}=y_{ij}$, and $y_{ii}=0$. Let
$$
\YY^*=f(\MM^*)=(y_{ij}^*)_{N\times N},
\qquad
y_{ij}^*=f(m_{ij}^*),
$$
and let $R_{NN}:=b_{NN}-a_{NN}$.
Recall that
$$
r_{NN}
=
\max_{i\neq j}
\left|
\EEE\widehat y_{ij}
-
y_{ij}^*
\right|.
$$
For the diagonal entries, $y_{ii}=0$, while the algebraically completed
signal satisfies $m_{ii}^*=\tau_0$. Hence, under the condition
$f(\tau_0)\in[a_{NN},b_{NN}]$,
$$
\left|
\EEE\widehat y_{ii}
-
y_{ii}^*
\right|
=
\left|
\PPP_{[a_{NN},b_{NN}]}(0)
-
f(\tau_0)
\right|
\le
R_{NN}.
$$
We then decompose
\begin{equation}\label{eq:uni_bias_variance_split}
\begin{aligned}
\frac{1}{N^2}
\EEE
\|
\widehat\XX-\YY^*
\|_F^2
\le\;&
\frac{2}{N^2}
\EEE
\left\|
\widehat\XX-\EEE(\widehat\YY)
\right\|_F^2\\
&+
\frac{2}{N^2}
\left\|
\EEE(\widehat\YY)-\YY^*
\right\|_F^2.
\end{aligned}
\end{equation}
By the definition of $r_{NN}$, the off-diagonal entries satisfy
$| \EEE\widehat y_{ij} - y_{ij}^* | \le r_{NN}, \qquad i\neq j.$
Combining this with the diagonal bound above yields
\begin{equation}\label{eq:uni_bias_pure}
\frac{2}{N^2}
\left\|
\EEE(\widehat\YY)-\YY^*
\right\|_F^2
\le
2r_{NN}^2
+
\frac{2R_{NN}^2}{N}.
\end{equation}

Applying the symmetric analogue of the modified USVT argument above gives
\begin{equation}\label{eq:uni_variance_bound}
\frac{1}{N^2}
\EEE
\left\|
\widehat\XX-\EEE(\widehat\YY)
\right\|_F^2
\le
c
\frac{R_{NN}}{N^{3/2}}
\left\|
\EEE(\widehat\YY)
-
a_{NN}\EE_N\EE_N^\top
\right\|_*
+
cR_{NN}^2e^{-cN}
+
c\frac{R_{NN}^2}{N}.
\end{equation}

Define the diagonal completion
$\overline\YY=(\overline y_{ij})_{N\times N}$ by
$$
\overline y_{ij}
=
\begin{cases}
\EEE\widehat y_{ij},
& i\neq j,\\
f(\tau_0),
& i=j.
\end{cases}
$$
Since
$\EEE\widehat y_{ii}
=
\PPP_{[a_{NN},b_{NN}]}(0)$,
we have
\begin{equation}\label{eq:uni_diagonal_completion}
\left\|
\EEE(\widehat\YY)
-
\overline\YY
\right\|_*
\le
NR_{NN}.
\end{equation}
Consequently,
\begin{equation}\label{eq:uni_nuclear_decomp}
\left\|
\EEE(\widehat\YY)
-
a_{NN}\EE_N\EE_N^\top
\right\|_*
\le
NR_{NN}
+
\left\|
\overline\YY
-
a_{NN}\EE_N\EE_N^\top
\right\|_*.
\end{equation}

Let $\mathcal G$ be a maximal $\delta$-separated subset of
$\{\ttt_i^*:i\in[N]\}$. Then
$|\mathcal G| \le N_\delta := c(C_0/\delta)^K.$
For each $i\in[N]$, let
$p(\ttt_i^*)\in\mathcal G$ satisfy
$\| \ttt_i^*-p(\ttt_i^*) \| \le \delta,$
and define
$$
m_{ij}^{\delta}
:=
\tau_0
-
\|
p(\ttt_i^*)
-
p(\ttt_j^*)
\|^2,
\qquad
\widehat y_{ij}^{\delta}
:=
f(m_{ij}^{\delta}).
$$
Let
$\widehat\YY_\delta = (\widehat y_{ij}^{\delta})_{N\times N}.$

Let
$L_f = \sup_{|m|\le C_0}|f'(m)| < \infty.$
As in the bipartite case,
$| m_{ij}^*-m_{ij}^{\delta} | \le 8C_0\delta.$

We decompose
\begin{equation}\label{eq:uni_completed_nuclear_decomposition}
\begin{aligned}
\left\|
\overline\YY
-
a_{NN}\EE_N\EE_N^\top
\right\|_*
\le\;&
\underbrace{
\left\|
\overline\YY-\widehat\YY_\delta
\right\|_*
}_{\mathrm{(I)}}
+
\underbrace{
\left\|
\widehat\YY_\delta
-
a_{NN}\EE_N\EE_N^\top
\right\|_*
}_{\mathrm{(II)}}.
\end{aligned}
\end{equation}

For Part~$\mathrm{(I)}$, if $i\neq j$,
$$
\left|
\overline y_{ij}
-
\widehat y_{ij}^{\delta}
\right|
\le
r_{NN}+8C_0L_f\delta.
$$
For $i=j$,
$\overline y_{ii} = \widehat y_{ii}^{\delta} = f(\tau_0).$
Therefore,
\begin{equation}\label{eq:uni_part_I}
\mathrm{(I)}
\le
N^{3/2}(r_{NN}+c\delta).
\end{equation}

For Part~$\mathrm{(II)}$,
$$
\rank
\left(
\widehat\YY_\delta
-
a_{NN}\EE_N\EE_N^\top
\right)
\le
N_\delta.
$$
Moreover,
$$
\left|
\widehat y_{ij}^{\delta}
-
a_{NN}
\right|
\le
R_{NN}+r_{NN},
$$
and hence
\begin{equation}\label{eq:uni_part_II}
\mathrm{(II)}
\le
c\delta^{-K/2}
N(R_{NN}+r_{NN}).
\end{equation}

Combining the preceding bounds gives
\begin{equation}\label{eq:uni_nuclear_bound}
\left\|
\EEE(\widehat\YY)
-
a_{NN}\EE_N\EE_N^\top
\right\|_*
\le
N^{3/2}(r_{NN}+c\delta)
+
cN(R_{NN}+r_{NN})\delta^{-K/2}
+
NR_{NN}.
\end{equation}
Substituting this into \eqref{eq:uni_variance_bound} and combining with
\eqref{eq:uni_bias_pure} gives
\begin{equation}\label{eq:uni_X_pre_rate}
\begin{aligned}
\frac{1}{N^2}
\EEE
\|
\widehat\XX-\YY^*
\|_F^2
\le\;&
cR_{NN}(r_{NN}+\delta)
+
cR_{NN}(R_{NN}+r_{NN})
\frac{\delta^{-K/2}}{\sqrt N}
+
cr_{NN}^2\\
&+
c\frac{R_{NN}^2}{\sqrt N}
+
cR_{NN}^2e^{-cN}.
\end{aligned}
\end{equation}

Choose
$$
\delta
=
\left\{
\frac{(R_{NN}+r_{NN})^2}{N}
\right\}^{\frac{1}{K+2}}.
$$
Under
$R_{NN}+r_{NN}=o(\sqrt N)$,
we obtain
\begin{equation}\label{eq:uni_X_rate}
\frac{1}{N^2}
\EEE
\|
\widehat\XX-\YY^*
\|_F^2
\le
cR_{NN}
\left\{
\frac{(R_{NN}+r_{NN})^2}{N}
\right\}^{\frac{1}{K+2}}
+
cr_{NN}(R_{NN}+r_{NN}).
\end{equation}

Finally, since
$m_{ij}^*\in[-C_0,C_0]
\subseteq[-C_{\mathrm{proj}},C_{\mathrm{proj}}]$,
the same inverse-link argument as in the bipartite case gives
$$
\frac{1}{N^2}
\left\|
\widehat\MM^{\SVD}-\MM^*
\right\|_F^2
=
O_p
\left(
R_{NN}
\left\{
\frac{(R_{NN}+r_{NN})^2}{N}
\right\}^{\frac{1}{K+2}}
+
r_{NN}(R_{NN}+r_{NN})
\right).
$$
Since $J=N$, this is the claimed bound. Together with the preceding bipartite arguments, this completes the
proof of Theorem~\ref{thm:converge in prob}.

\subsection{Proof of Corollary~\ref{cor:support_adaptive_truncation}}

\begin{proof}
By Assumption~\ref{assumption:algo1_bounded},
$|m_{ij}^*|\le C_0$ uniformly over the observed entries. Since $f$ is
continuous, there exists a fixed constant $C_\mu>0$ such that
$\max_{i,j} |f(m_{ij}^*)| \le C_\mu$
for bipartite data, with the maximum taken over $i\neq j$ in the
unipartite case.

By Assumption~\ref{assumption:algo1_subexponential}, there exist fixed
constants $c_0,C_{\rm tail}>0$ such that
\[
\Pr\left(
|y_{ij}-f(m_{ij}^*)|>t
\right)
\le
C_{\rm tail}e^{-c_0t},
\qquad t\ge0,
\]
uniformly over all observed entries for all sufficiently large
$N$ and $J$.
Choose a fixed constant
\[
\kappa>\frac{1}{c_0(K+2)}.
\]

Recall that
\[
a_{NJ}
=
\max\left\{
\inf\mathcal Y_{NJ},-\kappa\log J
\right\},
\qquad
b_{NJ}
=
\min\left\{
\sup\mathcal Y_{NJ},\kappa\log J
\right\}.
\]
Since
$\inf\mathcal Y_{NJ}\le \EEE(y_{ij})\le\sup\mathcal Y_{NJ}$
and
$|\EEE(y_{ij})|=|f(m_{ij}^*)|\le C_\mu$,
the condition
$\inf\mathcal Y_{NJ}<\sup\mathcal Y_{NJ}$
implies, for all sufficiently large $J$,
\[
-\kappa\log J
\le
a_{NJ}
<
b_{NJ}
\le
\kappa\log J.
\]
Hence,
\begin{equation}\label{eq:cor_R_rate}
R_{NJ}
=
b_{NJ}-a_{NJ}
=
O(\log J).
\end{equation}

We next bound the truncation bias. For every observed entry,
\[
\begin{aligned}
&
\left|
\EEE\left\{
\PPP_{[a_{NJ},b_{NJ}]}(y_{ij})
-
f(m_{ij}^*)
\right\}
\right|\\
&\qquad=
\left|
\EEE\left\{
\PPP_{[a_{NJ},b_{NJ}]}(y_{ij})
-
y_{ij}
\right\}
\right|\\
&\qquad\le
\EEE\{a_{NJ}-y_{ij}\}_+
+
\EEE\{y_{ij}-b_{NJ}\}_+.
\end{aligned}
\]

Consider first the lower-tail term. If
$a_{NJ}=\inf\mathcal Y_{NJ}$, then
$y_{ij}\ge a_{NJ}$ almost surely for every observed entry and therefore
$\EEE\{a_{NJ}-y_{ij}\}_+=0.$
Otherwise,
$a_{NJ}=-\kappa\log J$. For all sufficiently large $J$,
\[
f(m_{ij}^*)-a_{NJ}
=
f(m_{ij}^*)+\kappa\log J
\ge
\kappa\log J-C_\mu.
\]
By tail integration,
\[
\begin{aligned}
\EEE\{a_{NJ}-y_{ij}\}_+
&=
\int_0^\infty
\Pr(y_{ij}<a_{NJ}-s)\,ds\\
&\le
\int_0^\infty
C_{\rm tail}
\exp\left[
-c_0
\{f(m_{ij}^*)-a_{NJ}+s\}
\right]ds\\
&\lesssim
J^{-c_0\kappa}.
\end{aligned}
\]

Similarly, if
$b_{NJ}=\sup\mathcal Y_{NJ}$, then
$\EEE\{y_{ij}-b_{NJ}\}_+=0.$
Otherwise,
$b_{NJ}=\kappa\log J$, and for all sufficiently large $J$,
$b_{NJ}-f(m_{ij}^*) \ge \kappa\log J-C_\mu.$
Thus,
\[
\begin{aligned}
\EEE\{y_{ij}-b_{NJ}\}_+
&=
\int_0^\infty
\Pr(y_{ij}>b_{NJ}+s)\,ds\\
&\le
\int_0^\infty
C_{\rm tail}
\exp\left[
-c_0
\{b_{NJ}+s-f(m_{ij}^*)\}
\right]ds\\
&\lesssim
J^{-c_0\kappa}.
\end{aligned}
\]
Consequently,
\begin{equation}\label{eq:cor_r_rate}
r_{NJ}
=
O\left(
J^{-c_0\kappa}
\right).
\end{equation}

Since
$R_{NJ}+r_{NJ} = O(\log J) = o(\sqrt J),$
the nonlinear-link part of
Theorem~\ref{thm:converge in prob} applies. Moreover,
\[
\begin{aligned}
R_{NJ}
\left\{
\frac{(R_{NJ}+r_{NJ})^2}{J}
\right\}^{\frac{1}{K+2}}
&=
O\left[
(\log J)
\left\{
\frac{(\log J)^2}{J}
\right\}^{\frac{1}{K+2}}
\right]\\
&=
O\left(
J^{-\frac{1}{K+2}}
(\log J)^{\frac{K+4}{K+2}}
\right).
\end{aligned}
\]
Also, since
$c_0\kappa>\frac{1}{K+2},$
we have
\[
\begin{aligned}
r_{NJ}(R_{NJ}+r_{NJ})
&=
O\left(
J^{-c_0\kappa}\log J
\right)\\
&=
o\left(
J^{-\frac{1}{K+2}}
(\log J)^{\frac{K+4}{K+2}}
\right).
\end{aligned}
\]
Therefore,
\[
\frac{1}{NJ}
\left\|
\widehat\MM^{\mathrm{SVD}}-\MM^*
\right\|_F^2
=
O_p\left(
J^{-\frac{1}{K+2}}
(\log J)^{\frac{K+4}{K+2}}
\right).
\]

We next consider the uniformly bounded-support case. Suppose
$\sup_{N,J} \sup_{y\in\mathcal Y_{NJ}} |y| < \infty.$
Then there exists a fixed constant $B>0$ such that
$\mathcal Y_{NJ}\subseteq[-B,B]$
for all $N,J$. Since $\kappa\log J>B$ for all sufficiently large $J$,
\[
a_{NJ}
=
\inf\mathcal Y_{NJ},
\qquad
b_{NJ}
=
\sup\mathcal Y_{NJ}.
\]
Hence
$\PPP_{[a_{NJ},b_{NJ}]}(y_{ij}) = y_{ij}$
almost surely for every observed entry, and therefore
$r_{NJ}=0.$
Moreover,
$R_{NJ} = b_{NJ}-a_{NJ} \le 2B = O(1).$
Applying Theorem~\ref{thm:converge in prob} again gives
\[
\frac{1}{NJ}
\left\|
\widehat\MM^{\mathrm{SVD}}-\MM^*
\right\|_F^2
=
O_p\left(
J^{-\frac{1}{K+2}}
\right).
\]

For unipartite data, the same argument applies to the off-diagonal
observations with $J=N$. The additional condition
$f(\tau_0)\in[a_{NN},b_{NN}]$
for all sufficiently large $N$ is precisely the condition required in
the unipartite part of Theorem~\ref{thm:converge in prob}. This
completes the proof.
\end{proof}

\section{Proofs for One-Step Refinement}
\label{app:one-step-refinement-proofs}

Throughout Appendix~\ref{app:one-step-refinement-proofs}, $f$ is assumed to be nonlinear, as in Theorem~\ref{thm:refinement}. Let $C_0>0$ denote the fixed population bound in
Assumption~\ref{assumption:algo1_bounded}, so that
$\|\MM^*\|_{\max}\le C_0$, and let $C_{\mathrm{proj}}$ denote the
projection radius used in Algorithm~\ref{alg:rsd}, chosen so that
$C_{\mathrm{proj}}\ge C_0$.

\subsection{Preliminary Factor Perturbation}

 Throughout the proof of Theorem~\ref{thm:refinement}, we apply the refinement step of Algorithm~\ref{alg:rsd} with $\MM^{\mathrm{init}}$ in place of $\widehat\MM^{\mathrm{SVD}}$, while retaining the notation used therein. In particular, $\widehat\VVV$ denotes the matrix of the leading $K+2$ right singular vectors of $\MM^{\mathrm{init}}$, and $\widehat\AAA$, $\widetilde\BBB$, and $\widetilde\AAA$ are constructed accordingly as in Algorithm~\ref{alg:rsd}. 
 
 Recall that $\MM^*=\UUU^*\DDDD^*\VVV^{*\top}$, with the rescaled SVD factors $\sqrt J\,\VVV^*$ and $J^{-1/2}\UUU^*\DDDD^*$. To account for the rotational non-identifiability of the factorization, let 
 $$ 
 \widehat\PP \in \argmin_{\PP\in\OOO_{K+2}} \|\widehat\VVV-\VVV^*\PP\|_F. 
 $$
 For notational simplicity, throughout Appendix~\ref{app:one-step-refinement-proofs} we use $\AAA^*$ and $\BBB^*$ to denote the corresponding aligned true factors, 
 $$ \AAA^* := \sqrt J\,\VVV^*\widehat\PP, \qquad \BBB^* := J^{-1/2}\UUU^*\DDDD^*\widehat\PP. 
 $$ 
 Since $\widehat\PP$ is orthogonal, 
 $\MM^*=\BBB^*\AAA^{*\top}.$ 

Thus, all factor estimation errors below are measured relative to this
optimally aligned representation. By
Assumptions~\ref{assumption:algo1_bounded}
and~\ref{assumption:algo2_incoherence}, together with the orthogonality
of $\widehat\PP$, there exist fixed constants
$\bar C_1,\bar C_2>0$, independent of $(N,J)$, such that
$$ 
\|\BBB^*\|_{2\to\infty}\le \bar C_1,
\qquad
\|\AAA^*\|_{2\to\infty}\le \bar C_2.
$$
Fix an arbitrary constant $\eta_B>0$ and set $C_1:=\bar C_1+\eta_B$. Then
$\|\BBB^*\|_{2\to\infty} \le \bar C_1 = C_1-\eta_B.$
We take $C_{\mathrm{proj}}$ sufficiently large so that
$C_{\mathrm{proj}}\ge C_0\vee\bar C_2$, and set
$C_2:=C_{\mathrm{proj}}$. Then
$$
\|\AAA^*\|_{2\to\infty}\le C_2,
\qquad
\|\widehat\AAA\|_{2\to\infty}\le C_2.
$$

We first establish the following lemma.

\begin{lemma}\label{lem:error_of_a}
Suppose that $\|\MM^{\mathrm{init}}-\MM^*\|_F \le \frac12\sigma_{K+2}(\MM^*)$. Then
$$
\|\widehat\AAA-\AAA^*\|_F
\le
8\sqrt J\,
\sigma_{K+2}^{-1}(\MM^*)
\|\MM^{\mathrm{init}}-\MM^*\|_F.
$$
\end{lemma}

\begin{proof}
Let
$\Delta_{\MM}:=\MM^{\mathrm{init}}-\MM^*$.
By Weyl's inequality,
$$
\begin{aligned}
\sigma_{K+2}(\MM^{\mathrm{init}})
&\ge
\sigma_{K+2}(\MM^*)-\|\Delta_{\MM}\|_2\\
&\ge
\sigma_{K+2}(\MM^*)-\|\Delta_{\MM}\|_F\\
&\ge
\frac12\sigma_{K+2}(\MM^*).
\end{aligned}
$$
Since $\MM^*$ has rank $K+2$, Wedin's $\sin\Theta$ theorem gives
$$
\inf_{\PP\in\OOO_{K+2}}
\|\widehat\VVV-\VVV^*\PP\|_F
\le
\frac{2\|\MM^{\mathrm{init}}-\MM^*\|_F}
{\sigma_{K+2}(\MM^{\mathrm{init}})}.
$$
By the definition of $\widehat\PP$,
$$
\|\widehat\VVV-\VVV^*\widehat\PP\|_F
\le
\frac{4\|\MM^{\mathrm{init}}-\MM^*\|_F}
{\sigma_{K+2}(\MM^*)}.
$$
Since
$\AAA^*=\sqrt J\,\VVV^*\widehat\PP$, it follows that
$$
\|\sqrt J\,\widehat\VVV-\AAA^*\|_F
\le
4\sqrt J\,
\sigma_{K+2}^{-1}(\MM^*)
\|\MM^{\mathrm{init}}-\MM^*\|_F.
$$

Let
$$
\mathcal C_A
:=
\left\{
\AAA\in\mathbb R^{J\times(K+2)}:
\|\AAA\|_{2\to\infty}\le C_{\mathrm{proj}}
\right\}.
$$
By the choice of $C_{\mathrm{proj}}$ above,
$\AAA^*\in\mathcal C_A$.
Since
$\widehat\AAA
=
\PPP_{C_{\mathrm{proj}},2\to\infty}(\sqrt J\,\widehat\VVV)$,
the definition of the projection yields
$$
\|\widehat\AAA-\sqrt J\,\widehat\VVV\|_F
\le
\|\AAA^*-\sqrt J\,\widehat\VVV\|_F.
$$
Therefore,
$$
\begin{aligned}
\|\widehat\AAA-\AAA^*\|_F
&\le
\|\widehat\AAA-\sqrt J\,\widehat\VVV\|_F
+
\|\sqrt J\,\widehat\VVV-\AAA^*\|_F\\
&\le
2\|\sqrt J\,\widehat\VVV-\AAA^*\|_F\\
&\le
8\sqrt J\,
\sigma_{K+2}^{-1}(\MM^*)
\|\MM^{\mathrm{init}}-\MM^*\|_F.
\end{aligned}
$$
\end{proof}

The remainder of the proof is organized by data structure.
Section~\ref{app:bi-one-step-refinement} develops the bipartite refinement analysis, including the
deterministic estimating-equation bounds, the required stochastic
controls, and the resulting two-to-infinity factor bounds.
Section~\ref{app:uni-one-step-refinement} then treats the unipartite case by accounting for diagonal
exclusion, symmetry, and the final symmetrization and rank-$(K+2)$
truncation. Section~\ref{app:proof-refinement-corollary} proves Corollary~\ref{cor:refinement_algo1}.

\subsection{Bipartite One-Step Refinement}\label{app:bi-one-step-refinement}

We first establish deterministic properties of the row- and column-wise
estimating equations, then derive the required stochastic bounds and
two-to-infinity perturbation controls, and finally complete the proof of
Theorem~\ref{thm:refinement} for bipartite data.

\subsubsection{Deterministic Bounds for Solutions to the Estimating Equations}

For $\ell\in\{1,2\}$, define
$$
\kappa_\ell(x)
:=
\sup_{|m|\le x}|f^{(\ell)}(m)|,
\qquad
\delta_1(x)
:=
\inf_{|m|\le x}f'(m).
$$
Define
$$
\kappa_1^*
:=
\kappa_1(C_0),
\qquad
\kappa_2^*
:=
\kappa_2(2C_1C_2),
\qquad
\delta_1^*
:=
\delta_1(C_0).
$$
By Assumption~\ref{assumption:algo2_link}, $\kappa_1^*,\kappa_2^*<\infty$ and $\delta_1^*>0$. For each $i\in[N]$, define
$$
S_{1,i}(\bbb_i;\AAA)
:=
\sum_{j=1}^{J}
\{y_{ij}-f(\aaa_j^\top\bbb_i)\}\aaa_j.
$$
The next lemma gives a deterministic bound for a local solution to
$S_{1,i}(\bbb_i;\AAA)=\mathbf 0_{K+2}$.

\begin{lemma}\label{lem:local_bi_existence_1} Let $\BBB^* \in \R^{N \times (K+2)}$ and $\AAA^* \in \R^{J \times (K+2)}$ be such that $\MM^* = \BBB^* \AAA^{*\top}$, and let $\EEEE = (\EEEE_{ij})_{N \times J}$ with $\EEEE_{ij} = y_{ij} - f(m_{ij}^*)$. Suppose Assumption~\ref{assumption:algo2_link} holds, $\|\MM^*\|_{\max}\le C_0$, and $\rank(\AAA)=K+2$. If $\|\BBB^*\|_{2\to\infty}\le C_1$, $\|\AAA^*\|_{2\to\infty},\|\AAA\|_{2\to\infty}\le C_2$, and there exists $\xi>0$ such that 
\begin{equation}\label{eq:item_1_new} 
\begin{aligned} 
&2\sigma_{K+2}^{-1}(\III_{1,i}(\AAA))
\Bigl\{
\|\EEEE_{i\cdot}\AAA\|
+\|\BB_{1,i}(\AAA)\|
+\beta_{1,i}(\AAA)\kappa_2(C_2(C_1+\xi))
\Bigr\}
\le \xi,\\
&2\gamma_{1,i}(\AAA)
\kappa_2(C_2(C_1+\xi))\,\xi
\le
\sigma_{K+2}(\III_{1,i}(\AAA)).
\end{aligned}
\end{equation}
where we define $\EEEE_{i\cdot}=(\EEEE_{ij})_{j\in[J]} \in \R^{1\times J}$, 
\begin{equation}
    \BB_{1,i}(\AAA)
    :=
    \sum_{j=1}^{J}
    f'(m_{ij}^*)\,\aaa_j(\aaa_j-\aaa_j^*)^\top \bbb_i^*
    \in \R^{K+2},
\end{equation}
\begin{equation}
    \III_{1,i}(\AAA)
    :=
    \sum_{j=1}^{J}
    f'(m_{ij}^*)\,\aaa_j\aaa_j^\top
    \in \R^{(K+2)\times (K+2)},
\end{equation}
and
\begin{equation}
    \beta_{1,i}(\AAA)
    :=
    \sup_{\|\uu\|=1}
    \sum_{j=1}^{J}
    \bigl((\aaa_j-\aaa_j^*)^\top \bbb_i^*\bigr)^2
    |\aaa_j^\top \uu|,
    \qquad
    \gamma_{1,i}(\AAA)
    :=
    \sup_{\|\uu\|=1}
    \sum_{j=1}^{J}
    |\aaa_j^\top \uu|^3.
\end{equation}
Then there exists a unique $\tilde{\bbb}_i$ such that $\|\tilde{\bbb}_i-\bbb_i^*\|\le \xi$ and $S_{1,i}(\tilde{\bbb}_i;\AAA)=\mathbf{0}_{K+2}$.
\end{lemma}

\begin{proof}
The proof is divided into two parts: existence and uniqueness.

\textbf{Existence.}
For $\bbb_i\in\R^{K+2}$, let
$m_{ij}:=\aaa_j^\top\bbb_i$.
Consider any $\bbb_i$ satisfying
$\|\bbb_i-\bbb_i^*\|\le\xi$.
By
$\|\BBB^*\|_{2\to\infty}\le C_1$ and
$\|\AAA^*\|_{2\to\infty},\|\AAA\|_{2\to\infty}\le C_2$,
we have
$$
|m_{ij}^*|
=
|\aaa_j^{*\top}\bbb_i^*|
\le
C_1C_2
\le
C_2(C_1+\xi),
$$
and
$$
|m_{ij}|
=
|\aaa_j^\top\bbb_i|
\le
\|\aaa_j\|\|\bbb_i\|
\le
C_2(C_1+\xi).
$$
Hence, every point between $m_{ij}^*$ and $m_{ij}$ lies in $[-C_2(C_1+\xi),\,C_2(C_1+\xi)]$. By Assumption~\ref{assumption:algo2_link}, $f$ is twice continuously differentiable on this interval.

We write
$$
S_{1,i}(\bbb_i;\AAA)
=
\sum_{j=1}^{J}\{y_{ij}-f(m_{ij}^*)\}\aaa_j
-
\sum_{j=1}^{J}\{f(m_{ij})-f(m_{ij}^*)\}\aaa_j.
$$
Applying a second-order Taylor expansion to $f(m_{ij})$ around
$m_{ij}^*$ gives
$$
f(m_{ij})-f(m_{ij}^*)
=
f'(m_{ij}^*)(m_{ij}-m_{ij}^*)
+
2^{-1}f''(\widetilde m_{ij})(m_{ij}-m_{ij}^*)^2,
$$
for some $\widetilde m_{ij}$ between $m_{ij}^*$ and $m_{ij}$.
Therefore,
\begin{equation}\label{eq:taylor_S}
\begin{aligned}
S_{1,i}(\bbb_i;\AAA)
&=
\AAA^\top\EEEE_{i\cdot}^\top
-
\sum_{j=1}^{J}
f'(m_{ij}^*)(m_{ij}-m_{ij}^*)\aaa_j\\
&\quad
-
2^{-1}\sum_{j=1}^{J}
f''(\widetilde m_{ij})(m_{ij}-m_{ij}^*)^2\aaa_j.
\end{aligned}
\end{equation}
Since
$$
m_{ij}-m_{ij}^*
=
\aaa_j^\top(\bbb_i-\bbb_i^*)
+
(\aaa_j-\aaa_j^*)^\top\bbb_i^*,
$$
we obtain
$$
\begin{aligned}
S_{1,i}(\bbb_i;\AAA)
&=
\AAA^\top\EEEE_{i\cdot}^\top
-
\sum_{j=1}^{J}
f'(m_{ij}^*)\aaa_j\aaa_j^\top
(\bbb_i-\bbb_i^*)\\
&\quad
-
\sum_{j=1}^{J}
f'(m_{ij}^*)\aaa_j
(\aaa_j-\aaa_j^*)^\top\bbb_i^*\\
&\quad
-
2^{-1}\sum_{j=1}^{J}
f''(\widetilde m_{ij})(m_{ij}-m_{ij}^*)^2\aaa_j.
\end{aligned}
$$
Multiplying both sides from the left by
$(\bbb_i-\bbb_i^*)^\top$ gives
$$
\begin{aligned}
(\bbb_i-\bbb_i^*)^\top
S_{1,i}(\bbb_i;\AAA)
&=
(\bbb_i-\bbb_i^*)^\top
\AAA^\top\EEEE_{i\cdot}^\top\\
&\quad
-
(\bbb_i-\bbb_i^*)^\top
\III_{1,i}(\AAA)
(\bbb_i-\bbb_i^*)\\
&\quad
-
(\bbb_i-\bbb_i^*)^\top
\BB_{1,i}(\AAA)\\
&\quad
-
2^{-1}(\bbb_i-\bbb_i^*)^\top
\sum_{j=1}^{J}
f''(\widetilde m_{ij})
(m_{ij}-m_{ij}^*)^2\aaa_j.
\end{aligned}
$$

Now consider $\bbb_i$ satisfying
$\|\bbb_i-\bbb_i^*\|=\xi$.
The first term satisfies
\begin{equation}\label{eq:thm_1_ineq_1}
\left|
(\bbb_i-\bbb_i^*)^\top
\AAA^\top\EEEE_{i\cdot}^\top
\right|
\le
\xi\|\EEEE_{i\cdot}\AAA\|.
\end{equation}

Since $|m_{ij}^*|\le C_0$, Assumption~\ref{assumption:algo2_link}
implies
$f'(m_{ij}^*)\ge\delta_1^*>0.$
Hence,
$$
\III_{1,i}(\AAA)
=
\sum_{j=1}^{J}
f'(m_{ij}^*)\aaa_j\aaa_j^\top
\succeq
\delta_1^*\AAA^\top\AAA.
$$
Since $\rank(\AAA)=K+2$,
$$
\sigma_{K+2}(\III_{1,i}(\AAA))
\ge
\delta_1^*\sigma_{K+2}^2(\AAA)
>0.
$$
Therefore,
\begin{equation}\label{eq:thm_1_ineq_2}
-(\bbb_i-\bbb_i^*)^\top
\III_{1,i}(\AAA)
(\bbb_i-\bbb_i^*)
\le
-\xi^2\sigma_{K+2}(\III_{1,i}(\AAA)).
\end{equation}
Moreover,
\begin{equation}\label{eq:thm_1_ineq_3}
\left|
(\bbb_i-\bbb_i^*)^\top
\BB_{1,i}(\AAA)
\right|
\le
\xi\|\BB_{1,i}(\AAA)\|.
\end{equation}

For the last term, by the preceding bounds,
$|\widetilde m_{ij}| \le C_2(C_1+\xi).$
Writing
$\bbb_i-\bbb_i^*=\xi\uu$ with $\|\uu\|=1$, we obtain
\begin{equation}\label{eq:thm_1_ineq_4}
\begin{aligned}
&\left|
2^{-1}(\bbb_i-\bbb_i^*)^\top
\sum_{j=1}^{J}
f''(\widetilde m_{ij})
(m_{ij}-m_{ij}^*)^2\aaa_j
\right|\\
&\le
2^{-1}
\kappa_2(C_2(C_1+\xi))\,\xi
\sup_{\|\uu\|=1}
\sum_{j=1}^{J}
\left\{
(\aaa_j-\aaa_j^*)^\top\bbb_i^*
+
\xi\aaa_j^\top\uu
\right\}^2
|\aaa_j^\top\uu|\\
&\le
\kappa_2(C_2(C_1+\xi))
\left\{
\xi\beta_{1,i}(\AAA)
+
\xi^3\gamma_{1,i}(\AAA)
\right\},
\end{aligned}
\end{equation}
where the second inequality follows from
$(x+y)^2\le2x^2+2y^2$.

Combining
\eqref{eq:thm_1_ineq_1}--\eqref{eq:thm_1_ineq_4} gives
\begin{equation}\label{eq:cubic_bound_new}
\begin{aligned}
(\bbb_i-\bbb_i^*)^\top
S_{1,i}(\bbb_i;\AAA)
\le{}&
-\sigma_{K+2}(\III_{1,i}(\AAA))\xi^2\\
&+
\gamma_{1,i}(\AAA)
\kappa_2(C_2(C_1+\xi))\xi^3\\
&+
\Bigl\{
\|\EEEE_{i\cdot}\AAA\|
+
\|\BB_{1,i}(\AAA)\|
+
\beta_{1,i}(\AAA)
\kappa_2(C_2(C_1+\xi))
\Bigr\}\xi.
\end{aligned}
\end{equation}

By the first inequality in \eqref{eq:item_1_new},
$$
\Bigl\{
\|\EEEE_{i\cdot}\AAA\|
+
\|\BB_{1,i}(\AAA)\|
+
\beta_{1,i}(\AAA)
\kappa_2(C_2(C_1+\xi))
\Bigr\}\xi
\le
\frac12
\sigma_{K+2}(\III_{1,i}(\AAA))\xi^2.
$$
By the second inequality in \eqref{eq:item_1_new},
$$
\gamma_{1,i}(\AAA)
\kappa_2(C_2(C_1+\xi))\xi^3
\le
\frac12
\sigma_{K+2}(\III_{1,i}(\AAA))\xi^2.
$$
Substituting these two bounds into
\eqref{eq:cubic_bound_new} yields
$$
\sup_{\|\bbb_i-\bbb_i^*\|=\xi}
(\bbb_i-\bbb_i^*)^\top
S_{1,i}(\bbb_i;\AAA)
\le0.
$$
By Result 6.3.4 of Ortega and Rheinboldt (2000), there exists at least one
$\widetilde\bbb_i$ satisfying
$$
\|\widetilde\bbb_i-\bbb_i^*\|\le\xi,
\qquad
S_{1,i}(\widetilde\bbb_i;\AAA)
=
\mathbf 0_{K+2}.
$$

\textbf{Uniqueness.}
Take any
$\bbb^{(1)},\bbb^{(2)}\in\R^{K+2}$ and let
$\Delta:=\bbb^{(1)}-\bbb^{(2)}$.
Then
$$
\begin{aligned}
&\Delta^\top
\left\{
S_{1,i}(\bbb^{(1)};\AAA)
-
S_{1,i}(\bbb^{(2)};\AAA)
\right\}\\
&=
-\sum_{j=1}^{J}
\left\{
f(\aaa_j^\top\bbb^{(1)})
-
f(\aaa_j^\top\bbb^{(2)})
\right\}
\aaa_j^\top\Delta.
\end{aligned}
$$
Since Assumption~\ref{assumption:algo2_link} implies that $f$ is strictly
increasing, for every $j$,
$$
\left\{
f(\aaa_j^\top\bbb^{(1)})
-
f(\aaa_j^\top\bbb^{(2)})
\right\}
\aaa_j^\top\Delta
\ge0,
$$
with strict inequality whenever
$\aaa_j^\top\Delta\neq0$.
Because $\rank(\AAA)=K+2$, for every $\Delta\neq0$ there exists at least
one $j$ such that $\aaa_j^\top\Delta\neq0$.
Hence,
$$
\Delta^\top
\left\{
S_{1,i}(\bbb^{(1)};\AAA)
-
S_{1,i}(\bbb^{(2)};\AAA)
\right\}
<0
\qquad
\text{for every }\Delta\neq0.
$$
Therefore, two distinct zeros of
$S_{1,i}(\cdot;\AAA)$ cannot exist, and the solution obtained above is
unique.
\end{proof}

\begin{lemma}\label{lem:local_bi_existence_2}
Let $\BBB^* \in \R^{N \times (K+2)}$ and
$\AAA^* \in \R^{J \times (K+2)}$ be such that
$\MM^*=\BBB^*\AAA^{*\top}$, and let
$\EEEE=(\EEEE_{ij})_{N\times J}$ with
$\EEEE_{ij}=y_{ij}-f(m_{ij}^*)$.
Suppose Assumption~\ref{assumption:algo2_link} holds,
$\|\MM^*\|_{\max}\le C_0$,
$\rank(\AAA)=K+2$,
$\|\BBB^*\|_{2\to\infty}\le C_1$,
$\|\AAA^*\|_{2\to\infty},\|\AAA\|_{2\to\infty}\le C_2$.
If
\begin{equation}\label{eq:key_equation_new}
\begin{aligned}
&2\Bigl\{
\|\EEEE_{i\cdot}\AAA\|
+\|\BB_{1,i}(\AAA)\|
+\beta_{1,i}(\AAA)\kappa_2^*
\Bigr\}
\le
\sigma_{K+2}(\III_{1,i}(\AAA))C_1,\\
&4\gamma_{1,i}(\AAA)\kappa_2^*
\Bigl\{
\|\EEEE_{i\cdot}\AAA\|
+\|\BB_{1,i}(\AAA)\|
+\beta_{1,i}(\AAA)\kappa_2^*
\Bigr\}
\le
\sigma_{K+2}^2(\III_{1,i}(\AAA)),
\end{aligned}
\end{equation}
then there exists a unique $\widetilde{\bbb}_i$ such that
$S_{1,i}(\widetilde{\bbb}_i;\AAA)=\mathbf{0}_{K+2}$, and
\begin{equation}
\label{eq:bi_error_bound_new}
\|\widetilde{\bbb}_i-\bbb_i^*\|
\le
2\sigma_{K+2}^{-1}(\III_{1,i}(\AAA))
\Bigl\{
\|\EEEE_{i\cdot}\AAA\|
+
\|\BB_{1,i}(\AAA)\|
+
\beta_{1,i}(\AAA)\kappa_2^*
\Bigr\}.
\end{equation}
Moreover, the solution $\widetilde{\bbb}_i$ also satisfies
$\|\widetilde{\bbb}_i-\bbb_i^*\|\le C_1$.
\end{lemma}

\begin{proof}
Let
$$
\xi
=
2\sigma_{K+2}^{-1}(\III_{1,i}(\AAA))
\Bigl\{
\|\EEEE_{i\cdot}\AAA\|
+
\|\BB_{1,i}(\AAA)\|
+
\beta_{1,i}(\AAA)\kappa_2^*
\Bigr\}.
$$
By the first inequality in \eqref{eq:key_equation_new},
$\xi \le C_1.$
Therefore,
$C_2(C_1+\xi) \le 2C_1C_2.$
By the monotonicity of $\kappa_2(\cdot)$,
\begin{equation}
\label{eq:kappa_compare_new}
\kappa_2(C_2(C_1+\xi))
\le
\kappa_2(2C_1C_2)
=
\kappa_2^*.
\end{equation}
Hence,
\begin{equation}\label{eq:item_6_new}
\begin{aligned}
&2\sigma_{K+2}^{-1}(\III_{1,i}(\AAA))
\Bigl\{
\|\EEEE_{i\cdot}\AAA\|
+
\|\BB_{1,i}(\AAA)\|\\
&\hspace{35mm}
+
\beta_{1,i}(\AAA)
\kappa_2(C_2(C_1+\xi))
\Bigr\}
\le
\xi.
\end{aligned}
\end{equation}

Moreover, by \eqref{eq:kappa_compare_new}, the definition of $\xi$, and the
second inequality in \eqref{eq:key_equation_new},
\begin{equation}\label{eq:item_7_new}
\begin{aligned}
&2\gamma_{1,i}(\AAA)
\kappa_2(C_2(C_1+\xi))\,\xi\\
&\le
2\gamma_{1,i}(\AAA)\kappa_2^*\,\xi\\
&=
4\gamma_{1,i}(\AAA)\kappa_2^*
\sigma_{K+2}^{-1}(\III_{1,i}(\AAA))
\Bigl\{
\|\EEEE_{i\cdot}\AAA\|
+
\|\BB_{1,i}(\AAA)\|
+
\beta_{1,i}(\AAA)\kappa_2^*
\Bigr\}\\
&\le
\sigma_{K+2}(\III_{1,i}(\AAA)).
\end{aligned}
\end{equation}

Therefore, \eqref{eq:item_6_new} and \eqref{eq:item_7_new} imply that
all the conditions in \eqref{eq:item_1_new} of
Lemma~\ref{lem:local_bi_existence_1} hold with this choice of $\xi$.
Applying Lemma~\ref{lem:local_bi_existence_1}, there exists a unique
$\widetilde{\bbb}_i$ such that
$$
\|\widetilde{\bbb}_i-\bbb_i^*\|
\le
\xi
\qquad\text{and}\qquad
S_{1,i}(\widetilde{\bbb}_i;\AAA)
=
\mathbf{0}_{K+2}.
$$
Substituting the definition of $\xi$ gives
\eqref{eq:bi_error_bound_new}. Finally, since $\xi\le C_1$,
$\|\widetilde{\bbb}_i-\bbb_i^*\| \le C_1.$
\end{proof}

\subsubsection{Non-Asymptotic Stochastic Bounds}

\begin{lemma}\label{lem:max_tail_prob_2}
Let $\EEEE = (\EEEE_{ij})_{N \times J}$ and let $\EEEE_{ij}$ be independent
mean-zero sub-exponential random variables with parameters
$(\nu_{ij}, b_{ij})$. Define
$$
\nu = \max_{i\in[N],\,j\in[J]} \nu_{ij},
\qquad
b = \max_{i\in[N],\,j\in[J]} b_{ij}.
$$
Then, for any $\delta\in(0,1)$, with probability at least $1-\delta$,
$$
\|\EEEE\|_{\max}
\le
\max\left\{
\nu\sqrt{2\log\frac{2NJ}{\delta}},
\;
2b\log\frac{2NJ}{\delta}
\right\}.
$$
\end{lemma}

\begin{proof}
For each $(i,j)$ and any $t>0$, the sub-exponential tail bound gives
$$
\Pr(|\EEEE_{ij}|\ge t)
\le
2\left(
e^{-t^2/(2\nu_{ij}^2)}
\vee
e^{-t/(2b_{ij})}
\right)
\le
2\left(
e^{-t^2/(2\nu^2)}
\vee
e^{-t/(2b)}
\right).
$$
Therefore, by the union bound,
$$
\begin{aligned}
\Pr(\|\EEEE\|_{\max}\ge t)
&=
\Pr\left(
\max_{i\in[N],\,j\in[J]}|\EEEE_{ij}|\ge t
\right)\\
&\le
\sum_{i=1}^N\sum_{j=1}^J
\Pr(|\EEEE_{ij}|\ge t)\\
&\le
2NJ
\left(
e^{-t^2/(2\nu^2)}
\vee
e^{-t/(2b)}
\right).
\end{aligned}
$$
Now choose
$$
t
=
\max\left\{
\nu\sqrt{2\log\frac{2NJ}{\delta}},
\;
2b\log\frac{2NJ}{\delta}
\right\}.
$$
Then
$$
e^{-t^2/(2\nu^2)}
\le
\frac{\delta}{2NJ},
\qquad
e^{-t/(2b)}
\le
\frac{\delta}{2NJ},
$$
and hence
$\Pr(\|\EEEE\|_{\max}\ge t)\le\delta.$
\end{proof}

\begin{lemma}[Upper bound for $\max_{i\in[N]}\|\EEEE_{i\cdot}\widehat\AAA\|$]
\label{lem:item_1}
Let $\EEEE=(\EEEE_{ij})_{N\times J}$ and let $\EEEE_{ij}$ be independent
mean-zero sub-exponential random variables with parameters
$(\nu_{ij},b_{ij})$. Define
$$
\nu=\max_{i\in[N],\,j\in[J]}\nu_{ij},
\qquad
b=\max_{i\in[N],\,j\in[J]}b_{ij},
$$
and
$$
L_{N,J}
=
\log\bigl(2N^2J(K+2)^2\bigr).
$$
Assume that $\|\AAA^*\|_{2\to\infty}\le C_2$. Then, with probability
at least $1-2/(N(K+2))$,
\begin{equation}
\label{eq:item_1_final}
\max_{i\in[N]}
\|\EEEE_{i\cdot}\widehat\AAA\|
\le
2C_2\sqrt{K+2}\,L_{N,J}\,
(\sqrt J\,\nu\vee b)
+
2L_{N,J}(\nu\vee b)\sqrt J\,
\|\widehat\AAA-\AAA^*\|_F.
\end{equation}
\end{lemma}

\begin{proof}
For each $i\in[N]$,
$$
\|\EEEE_{i\cdot}\widehat\AAA\|
\le
\|\EEEE_{i\cdot}\AAA^*\|
+
\|\EEEE_{i\cdot}(\widehat\AAA-\AAA^*)\|.
$$
Moreover,
$$
\begin{aligned}
\|\EEEE_{i\cdot}(\widehat\AAA-\AAA^*)\|
&=
\left\|
\sum_{j=1}^J
\EEEE_{ij}(\widehat\aaa_j-\aaa_j^*)
\right\|\\
&\le
\|\EEEE\|_{\max}
\sum_{j=1}^J
\|\widehat\aaa_j-\aaa_j^*\|\\
&\le
\|\EEEE\|_{\max}\sqrt J\,
\|\widehat\AAA-\AAA^*\|_F,
\end{aligned}
$$
where the last inequality follows from the Cauchy--Schwarz inequality.
Therefore,
\begin{equation}
\label{eq:decomp_item_1}
\max_{i\in[N]}
\|\EEEE_{i\cdot}\widehat\AAA\|
\le
\underbrace{
\max_{i\in[N]}
\|\EEEE_{i\cdot}\AAA^*\|
}_{(I)}
+
\underbrace{
\|\EEEE\|_{\max}\sqrt J\,
\|\widehat\AAA-\AAA^*\|_F
}_{(II)}.
\end{equation}

\noindent\textbf{Bound item (I).}
Recall that
$\AAA^* = \sqrt J\,\VVV^*\widehat\PP.$
Although $\widehat\PP$ is random, its randomness does not affect the norm in
item $(I)$ because $\widehat\PP$ is orthogonal. Indeed,
$$
\|\EEEE_{i\cdot}\AAA^*\|
=
\|\EEEE_{i\cdot}\sqrt J\,\VVV^*\widehat\PP\|
=
\|\EEEE_{i\cdot}\sqrt J\,\VVV^*\|.
$$
Define
$\AAA^0:=\sqrt J\,\VVV^*.$
Conditional on the true latent configuration, $\AAA^0$ is deterministic.
Moreover, by the orthogonality of $\widehat\PP$,
$$
\|\AAA^0\|_{2\to\infty}
=
\|\AAA^0\widehat\PP\|_{2\to\infty}
=
\|\AAA^*\|_{2\to\infty}
\le C_2.
$$
Consequently,
$(I) = \max_{i\in[N]} \|\EEEE_{i\cdot}\AAA^0\|.$

For each $i\in[N]$ and $k\in[K+2]$,
$$
\EEEE_{i\cdot}\AAA^0_{\cdot k}
=
\sum_{j=1}^J
\EEEE_{ij}\AAA^0_{jk}.
$$
Since $\|\AAA^0\|_{2\to\infty}\le C_2$,
$$
\max_{j\in[J]}|\AAA^0_{jk}|
\le C_2,
\qquad
\|\AAA^0_{\cdot k}\|^2
\le
JC_2^2.
$$
By the standard tail bound for weighted sums of independent centered
sub-exponential random variables,
$$
\Pr\left(
|\EEEE_{i\cdot}\AAA^0_{\cdot k}|
\ge t
\right)
\le
2\left(
\exp\left\{
-\frac{t^2}{2JC_2^2\nu^2}
\right\}
\vee
\exp\left\{
-\frac{t}{2C_2b}
\right\}
\right).
$$
Hence, for any $t>0$,
$$
\begin{aligned}
\Pr\left(
\|\EEEE_{i\cdot}\AAA^0\|
\ge t
\right)
&\le
\sum_{k=1}^{K+2}
\Pr\left(
|\EEEE_{i\cdot}\AAA^0_{\cdot k}|
\ge
\frac{t}{\sqrt{K+2}}
\right)\\
&\le
2(K+2)
\left(
\exp\left\{
-\frac{t^2}
{2JC_2^2(K+2)\nu^2}
\right\}
\vee
\exp\left\{
-\frac{t}
{2C_2\sqrt{K+2}\,b}
\right\}
\right).
\end{aligned}
$$
Applying the union bound over $i\in[N]$ gives
\begin{equation}
\label{eq:bound_I_prob}
\begin{aligned}
&\Pr\left(
\max_{i\in[N]}
\|\EEEE_{i\cdot}\AAA^0\|
\ge t
\right)\\
&\le
2N(K+2)
\left(
\exp\left\{
-\frac{t^2}
{2JC_2^2(K+2)\nu^2}
\right\}
\vee
\exp\left\{
-\frac{t}
{2C_2\sqrt{K+2}\,b}
\right\}
\right).
\end{aligned}
\end{equation}

Set
$t_1 = 2C_2\sqrt{K+2}\, L_{N,J}\, (\sqrt J\,\nu\vee b).$
Since $L_{N,J}\ge\log 2$, we have
$2L_{N,J}\ge\sqrt{2L_{N,J}}$, and therefore
$$
t_1
\ge
C_2\sqrt{2J(K+2)L_{N,J}}\,\nu,
\qquad
t_1
\ge
2C_2\sqrt{K+2}\,L_{N,J}\,b.
$$
Substituting $t=t_1$ into \eqref{eq:bound_I_prob} yields
$$
\begin{aligned}
\Pr\left((I)\ge t_1\right)
&\le
2N(K+2)e^{-L_{N,J}}\\
&=
\frac{1}{NJ(K+2)}
\le
\frac{1}{N(K+2)}.
\end{aligned}
$$
Hence, with probability at least
$1-\{N(K+2)\}^{-1}$,
\begin{equation}
\label{eq:bound_I_final}
(I)
\le
2C_2\sqrt{K+2}\,
L_{N,J}\,
(\sqrt J\,\nu\vee b).
\end{equation}

\noindent\textbf{Bound item (II).}
By Lemma~\ref{lem:max_tail_prob_2}, with probability at least
$1-\delta$,
$$
\|\EEEE\|_{\max}
\le
\nu\sqrt{2\log(2NJ/\delta)}
\vee
2b\log(2NJ/\delta).
$$
Choose
$\delta=\{N(K+2)\}^{-1}.$
Then
$\log(2NJ/\delta) = \log\bigl(2N^2J(K+2)\bigr) \le L_{N,J}.$
Since
$2L_{N,J}\ge\sqrt{2L_{N,J}}$,
we obtain
$\|\EEEE\|_{\max} \le 2L_{N,J}(\nu\vee b)$
with probability at least
$1-\{N(K+2)\}^{-1}$.
Therefore, with the same probability,
\begin{equation}
\label{eq:bound_II_final}
(II)
\le
2L_{N,J}(\nu\vee b)\sqrt J\,
\|\widehat\AAA-\AAA^*\|_F.
\end{equation}

Finally, combining
\eqref{eq:decomp_item_1},
\eqref{eq:bound_I_final}, and
\eqref{eq:bound_II_final}, and applying the union bound once more,
we conclude that \eqref{eq:item_1_final} holds with probability at least
$1-2/(N(K+2))$.
\end{proof}

\begin{lemma}[Upper bound for $\max_{i\in[N]}\|\BB_{1,i}(\widehat\AAA)\|$]
\label{lem:item_2}
Let
$\AAA^*=\sqrt J\,\VVV^*\widehat\PP$ and
$\BBB^*=J^{-1/2}\UUU^*\DDDD^*\widehat\PP$.
Assume
$\|\widehat\AAA\|_{2\to\infty}\le C_2$
and
$\|\BBB^*\|_{2\to\infty}\le C_1$.
Then,
\begin{equation}
\max_{i\in[N]}
\|\BB_{1,i}(\widehat\AAA)\|
\le
C_1C_2\kappa_1^*
\sqrt J\,
\|\widehat\AAA-\AAA^*\|_F.
\end{equation}
\end{lemma}

\begin{proof}
Recall that
$$
\BB_{1,i}(\widehat\AAA)
=
\sum_{j=1}^J
f'(m_{ij}^*)\,
\widehat\aaa_j
(\widehat\aaa_j-\aaa_j^*)^\top
\bbb_i^*.
$$
Since
$|m_{ij}^*|\le C_0$,
we have
$|f'(m_{ij}^*)| \le \kappa_1(C_0) = \kappa_1^*.$
Therefore, for each $i\in[N]$,
$$
\begin{aligned}
\|\BB_{1,i}(\widehat\AAA)\|
&\le
\sum_{j=1}^J
|f'(m_{ij}^*)|\,
\big\|
\widehat\aaa_j
(\widehat\aaa_j-\aaa_j^*)^\top
\bbb_i^*
\big\|\\
&=
\sum_{j=1}^J
|f'(m_{ij}^*)|\,
\|\widehat\aaa_j\|\,
\big|
(\widehat\aaa_j-\aaa_j^*)^\top
\bbb_i^*
\big|\\
&\le
\sum_{j=1}^J
|f'(m_{ij}^*)|\,
\|\widehat\aaa_j\|\,
\|\widehat\aaa_j-\aaa_j^*\|\,
\|\bbb_i^*\|\\
&\le
C_1C_2\kappa_1^*
\sum_{j=1}^J
\|\widehat\aaa_j-\aaa_j^*\|.
\end{aligned}
$$
Since the right-hand side does not depend on $i$, taking the maximum over
$i\in[N]$ gives
$$
\max_{i\in[N]}
\|\BB_{1,i}(\widehat\AAA)\|
\le
C_1C_2\kappa_1^*
\sum_{j=1}^J
\|\widehat\aaa_j-\aaa_j^*\|.
$$
By the Cauchy--Schwarz inequality,
$$
\sum_{j=1}^J
\|\widehat\aaa_j-\aaa_j^*\|
\le
\sqrt J
\left(
\sum_{j=1}^J
\|\widehat\aaa_j-\aaa_j^*\|^2
\right)^{1/2}
=
\sqrt J\,
\|\widehat\AAA-\AAA^*\|_F.
$$
Hence,
$$
\max_{i\in[N]}
\|\BB_{1,i}(\widehat\AAA)\|
\le
C_1C_2\kappa_1^*
\sqrt J\,
\|\widehat\AAA-\AAA^*\|_F.
$$
\end{proof}

\begin{lemma}[Bound for $\max_{i\in[N]}\beta_{1,i}(\widehat\AAA)$]
\label{lem:item_3}
Let
$\AAA^*=\sqrt J\,\VVV^*\widehat\PP$ and
$\BBB^*=J^{-1/2}\UUU^*\DDDD^*\widehat\PP$.
Assume
$\|\widehat\AAA\|_{2\to\infty}\le C_2$
and
$\|\BBB^*\|_{2\to\infty}\le C_1$.
Then
\begin{equation}
\max_{i\in[N]}
\beta_{1,i}(\widehat\AAA)
\le
C_1^2C_2
\|\widehat\AAA-\AAA^*\|_F^2.
\end{equation}
\end{lemma}

\begin{proof}
For each $i\in[N]$,
$$
\begin{aligned}
\beta_{1,i}(\widehat\AAA)
&=
\sup_{\|\uu\|=1}
\sum_{j=1}^J
\bigl(
(\widehat\aaa_j-\aaa_j^*)^\top
\bbb_i^*
\bigr)^2
|\widehat\aaa_j^\top\uu|\\
&\le
C_1^2C_2
\sum_{j=1}^J
\|\widehat\aaa_j-\aaa_j^*\|^2\\
&=
C_1^2C_2
\|\widehat\AAA-\AAA^*\|_F^2.
\end{aligned}
$$
Since the right-hand side does not depend on $i$, taking the maximum over
$i\in[N]$ yields
$$
\max_{i\in[N]}
\beta_{1,i}(\widehat\AAA)
\le
C_1^2C_2
\|\widehat\AAA-\AAA^*\|_F^2.
$$
\end{proof}

\begin{lemma}[Bound for $\gamma_{1,i}(\widehat\AAA)$]
\label{lem:item_4}
If $\|\widehat\AAA\|_{2\to\infty}\le C_2$, then
\begin{equation}
\gamma_{1,i}(\widehat\AAA)
\le
JC_2^3.
\end{equation}
\end{lemma}

\begin{proof}
By the definition of $\gamma_{1,i}$,
$$
\begin{aligned}
\gamma_{1,i}(\widehat\AAA)
&=
\sup_{\|\uu\|=1}
\sum_{j=1}^J
|\widehat\aaa_j^\top\uu|^3\\
&\le
\sum_{j=1}^J
\|\widehat\aaa_j\|^3\\
&\le
JC_2^3.
\end{aligned}
$$
\end{proof}

\subsubsection{Two-to-Infinity Bounds for the Refined Factors}

\begin{lemma}[Asymptotic analysis of $\widetilde{\BBB}$]
\label{lem:2 to infty beta}
Let
$\AAA^*=\sqrt J\,\VVV^*\widehat\PP$ and
$\BBB^*=J^{-1/2}\UUU^*\DDDD^*\widehat\PP$.
Suppose Assumption~\ref{assumption:algo1} holds.
Assume that
$\|\widehat\AAA-\AAA^*\|_F=O_p(e_{\AAA,F})$,
where $e_{\AAA,F}$ is deterministic and satisfies
$$
e_{\AAA,F}=\Omega(1),
\qquad
e_{\AAA,F}
=
o\bigl(\sqrt J(\log(NJ))^{-1}\bigr).
$$
Then, with probability tending to one, for every $i\in[N]$, the equation
$$
S_{1,i}(\bbb_i;\widehat\AAA)=\mathbf 0_{K+2}
$$
admits a unique solution $\widetilde{\bbb}_i$ satisfying
$\|\widetilde{\bbb}_i-\bbb_i^*\|\le C_1$.
Moreover,
$$
\|\widetilde{\BBB}-\BBB^*\|_{2\to\infty}
=
O_p\bigl(
J^{-1/2}\log(NJ)e_{\AAA,F}
\bigr).
$$
\end{lemma}

\begin{proof}
Let
$L_{N,J}:=\log(NJ).$
Under Assumption~\ref{assumption:algo1_subexponential}, the centered noises
$\EEEE_{ij}=y_{ij}-f(m_{ij}^*)$ are uniformly sub-exponential.
Hence, the parameters $\nu$ and $b$ in
Lemmas~\ref{lem:max_tail_prob_2} and~\ref{lem:item_1}
are uniformly bounded.
Since $K$ is fixed,
$\log\bigl(2N^2J(K+2)^2\bigr) \asymp L_{N,J}.$
Therefore, Lemma~\ref{lem:item_1} and
$\|\widehat\AAA-\AAA^*\|_F=O_p(e_{\AAA,F})$ imply
$$
\begin{aligned}
\max_{i\in[N]}
\|\EEEE_{i\cdot}\widehat\AAA\|
&=
O_p\left(
\sqrt J L_{N,J}
+
\sqrt J L_{N,J}e_{\AAA,F}
\right)\\
&=
O_p\left(
\sqrt J L_{N,J}e_{\AAA,F}
\right),
\end{aligned}
$$
where the second equality uses
$e_{\AAA,F}=\Omega(1)$.

Similarly, Lemmas~\ref{lem:item_2} and~\ref{lem:item_3} give
$$
\max_{i\in[N]}
\|\BB_{1,i}(\widehat\AAA)\|
=
O_p\left(
\sqrt J\,e_{\AAA,F}
\right),
$$
and
$$
\max_{i\in[N]}
\beta_{1,i}(\widehat\AAA)
=
O_p\left(
e_{\AAA,F}^2
\right).
$$

For each $i\in[N]$, define
$$
T_{1,i}
:=
\|\EEEE_{i\cdot}\widehat\AAA\|
+
\|\BB_{1,i}(\widehat\AAA)\|
+
\beta_{1,i}(\widehat\AAA)\kappa_2^*.
$$
Since $\kappa_2^*$ is a fixed constant,
$$
\max_{i\in[N]}T_{1,i}
=
O_p\left(
\sqrt J L_{N,J}e_{\AAA,F}
+
\sqrt J\,e_{\AAA,F}
+
e_{\AAA,F}^2
\right).
$$
By
$e_{\AAA,F} = o\bigl(\sqrt J L_{N,J}^{-1}\bigr),$
all three terms on the right-hand side are $o(J)$.
Hence,
\begin{equation}\label{eq:B3_T_row_small}
\max_{i\in[N]}T_{1,i}
=
o_p(J).
\end{equation}

By Lemma~\ref{lem:item_4},
\begin{equation}\label{eq:B3_gamma_row}
\max_{i\in[N]}
\gamma_{1,i}(\widehat\AAA)
\lesssim
J.
\end{equation}

Next, since
$\AAA^* = \sqrt J\,\VVV^*\widehat\PP,$
we have
$\sigma_{K+2}(\AAA^*)=\sqrt J.$
Moreover,
$$
\|\widehat\AAA-\AAA^*\|_2
\le
\|\widehat\AAA-\AAA^*\|_F
=
O_p(e_{\AAA,F})
=
o_p(\sqrt J).
$$
Therefore, Weyl's inequality yields
\begin{equation}\label{eq:B3_A_singular}
\sigma_{K+2}(\widehat\AAA)
\asymp
\sqrt J
\end{equation}
with probability tending to one.
In particular,
$\rank(\widehat\AAA)=K+2$ with probability tending to one.

Since
$|m_{ij}^*|\le C_0$,
Assumption~\ref{assumption:algo2_link} gives
$f'(m_{ij}^*) \ge \delta_1^* > 0.$
Thus,
$$
\III_{1,i}(\widehat\AAA)
=
\sum_{j=1}^J
f'(m_{ij}^*)
\widehat\aaa_j\widehat\aaa_j^\top
\succeq
\delta_1^*
\widehat\AAA^\top\widehat\AAA.
$$
Together with \eqref{eq:B3_A_singular}, uniformly over $i\in[N]$,
\begin{equation}\label{eq:B3_I_row}
\sigma_{K+2}
\bigl(\III_{1,i}(\widehat\AAA)\bigr)
\gtrsim
J
\end{equation}
with probability tending to one.

We now verify the two conditions in
\eqref{eq:key_equation_new}.
By \eqref{eq:B3_T_row_small} and \eqref{eq:B3_I_row},
$$
\max_{i\in[N]}
\frac{
2T_{1,i}
}{
\sigma_{K+2}
(\III_{1,i}(\widehat\AAA))C_1
}
=
o_p(1).
$$
Hence, with probability tending to one,
$2T_{1,i} \le \sigma_{K+2} (\III_{1,i}(\widehat\AAA))C_1$
simultaneously for all $i\in[N]$.

Moreover, by
\eqref{eq:B3_gamma_row},
\eqref{eq:B3_T_row_small}, and
\eqref{eq:B3_I_row},
$$
\max_{i\in[N]}
\frac{
4\gamma_{1,i}(\widehat\AAA)\kappa_2^*T_{1,i}
}{
\sigma_{K+2}^2
(\III_{1,i}(\widehat\AAA))
}
=
o_p(1).
$$
Thus, with probability tending to one,
$$
4\gamma_{1,i}(\widehat\AAA)\kappa_2^*T_{1,i}
\le
\sigma_{K+2}^2
(\III_{1,i}(\widehat\AAA))
$$
simultaneously for all $i\in[N]$. 

Therefore, Lemma~\ref{lem:local_bi_existence_2} applies simultaneously to
all $i\in[N]$ with probability tending to one.
Hence, for every $i\in[N]$, there exists a unique
$\widetilde\bbb_i$ satisfying
$$
S_{1,i}(\widetilde\bbb_i;\widehat\AAA)
=
\mathbf 0_{K+2},
\qquad
\|\widetilde\bbb_i-\bbb_i^*\|
\le
C_1.
$$

Furthermore, by \eqref{eq:bi_error_bound_new} and
\eqref{eq:B3_I_row},
$$
\begin{aligned}
\|\widetilde\BBB-\BBB^*\|_{2\to\infty}
&\lesssim
J^{-1}
\max_{i\in[N]}T_{1,i}\\
&=
O_p\left(
J^{-1/2}L_{N,J}e_{\AAA,F}
+
J^{-1/2}e_{\AAA,F}
+
J^{-1}e_{\AAA,F}^2
\right).
\end{aligned}
$$
Since
$L_{N,J}\ge1$ and
$e_{\AAA,F}
=o(\sqrt J L_{N,J}^{-1})$,
the last two terms are dominated by the first one.
Therefore,
$$
\|\widetilde\BBB-\BBB^*\|_{2\to\infty}
=
O_p\left(
J^{-1/2}L_{N,J}e_{\AAA,F}
\right).
$$
Since $L_{N,J}=\log(NJ)$,
$$
\|\widetilde\BBB-\BBB^*\|_{2\to\infty}
=
O_p\bigl(
J^{-1/2}\log(NJ)e_{\AAA,F}
\bigr).
$$
\end{proof}

\begin{lemma}[Asymptotic analysis of $\widetilde{\AAA}$]
\label{lem:2 to infty alpha}
Suppose the conditions of Lemma~\ref{lem:2 to infty beta} hold, with the
rate condition strengthened to
$$
e_{\AAA,F}
=
o\bigl(
\sqrt J(\log(NJ))^{-2}
\bigr).
$$
Then, with probability tending to one, for every $j\in[J]$, the equation
$$
S_{2,j}(\aaa_j;\widetilde\BBB)
=
\mathbf 0_{K+2}
$$
admits a unique solution $\widetilde{\aaa}_j$ satisfying
$\|\widetilde{\aaa}_j-\aaa_j^*\|\le C_2$.
Moreover,
$$
\|\widetilde{\AAA}-\AAA^*\|_{2\to\infty}
=
O_p\left(
N^{-1/2}\log(NJ)
+
J^{-1/2}(\log(NJ))^2e_{\AAA,F}
\right).
$$
\end{lemma}

\begin{proof}
Let
$L_{N,J}:=\log(NJ).$
By Lemma~\ref{lem:2 to infty beta},
$$
\|\widetilde\BBB-\BBB^*\|_{2\to\infty}
=
O_p\left(
J^{-1/2}L_{N,J}e_{\AAA,F}
\right).
$$
Since
$e_{\AAA,F} = o\bigl( \sqrt J L_{N,J}^{-2} \bigr),$
we have
$J^{-1/2}L_{N,J}e_{\AAA,F} = o(L_{N,J}^{-1}) = o(1).$
Hence,
\begin{equation}\label{eq:B3_B_row_consistency}
\|\widetilde\BBB-\BBB^*\|_{2\to\infty}
=
o_p(1).
\end{equation}

By the fixed-slack construction above,
$\|\BBB^*\|_{2\to\infty} \le C_1-\eta_B.$
Together with
\eqref{eq:B3_B_row_consistency}, this implies
$$
\Pr\left(
\|\widetilde\BBB-\BBB^*\|_{2\to\infty}
\le \eta_B
\right)
\to 1.
$$
Hence, with probability tending to one,
\begin{equation}\label{eq:B3_B_row_bound}
\begin{aligned}
\|\widetilde\BBB\|_{2\to\infty}
&\le
\|\BBB^*\|_{2\to\infty}
+
\|\widetilde\BBB-\BBB^*\|_{2\to\infty}\\
&\le
(C_1-\eta_B)+\eta_B\\
&=
C_1.
\end{aligned}
\end{equation}
Furthermore,
\begin{equation}\label{eq:B3_B_F}
\|\widetilde\BBB-\BBB^*\|_F
\le
\sqrt N\,
\|\widetilde\BBB-\BBB^*\|_{2\to\infty}
=
O_p\left(
\sqrt N J^{-1/2}
L_{N,J}e_{\AAA,F}
\right).
\end{equation}

Applying Lemma~\ref{lem:item_1} to $\EEEE^\top$ and interchanging the
roles of $(\AAA,J)$ and $(\BBB,N)$ gives
\begin{align}
\max_{j\in[J]}
\|\EEEE_{\cdot j}^\top\widetilde\BBB\|
&=
O_p\left(
\sqrt N L_{N,J}
+
\sqrt N L_{N,J}
\|\widetilde\BBB-\BBB^*\|_F
\right)\notag\\
&=
O_p\left(
\sqrt N L_{N,J}
+
NJ^{-1/2}L_{N,J}^2e_{\AAA,F}
\right).
\label{eq:B3_noise_col}
\end{align}

By the column-wise analogue of Lemma~\ref{lem:item_2},
\begin{equation}\label{eq:B3_B_col}
\max_{j\in[J]}
\|\BB_{2,j}(\widetilde\BBB)\|
=
O_p\left(
NJ^{-1/2}L_{N,J}e_{\AAA,F}
\right).
\end{equation}
Similarly, the column-wise analogue of Lemma~\ref{lem:item_3} gives
\begin{equation}\label{eq:B3_beta_col}
\max_{j\in[J]}
\beta_{2,j}(\widetilde\BBB)
=
O_p\left(
NJ^{-1}L_{N,J}^2e_{\AAA,F}^2
\right).
\end{equation}
Also, by the column-wise analogue of Lemma~\ref{lem:item_4},
\begin{equation}\label{eq:B3_gamma_col}
\max_{j\in[J]}
\gamma_{2,j}(\widetilde\BBB)
\lesssim
N
\end{equation}
with probability tending to one.

For each $j\in[J]$, define
$$
T_{2,j}
:=
\|\EEEE_{\cdot j}^\top\widetilde\BBB\|
+
\|\BB_{2,j}(\widetilde\BBB)\|
+
\beta_{2,j}(\widetilde\BBB)\kappa_2^*.
$$
Combining
\eqref{eq:B3_noise_col}--\eqref{eq:B3_beta_col}, we obtain
$$
\begin{aligned}
\max_{j\in[J]}T_{2,j}
=
O_p\Bigl(&
\sqrt N L_{N,J}
+
NJ^{-1/2}L_{N,J}^2e_{\AAA,F}\\
&+
NJ^{-1/2}L_{N,J}e_{\AAA,F}
+
NJ^{-1}L_{N,J}^2e_{\AAA,F}^2
\Bigr).
\end{aligned}
$$

Since $N\ge J$,
$L_{N,J} = \log(NJ) \le 2\log N = o(\sqrt N),$
and therefore
$\sqrt N L_{N,J} = o(N).$
Moreover, by
$e_{\AAA,F}
=o(\sqrt J L_{N,J}^{-2})$,
$NJ^{-1/2}L_{N,J}^2e_{\AAA,F} = o(N).$
The remaining two terms are also $o(N)$.
Consequently,
\begin{equation}\label{eq:B3_T_col_small}
\max_{j\in[J]}T_{2,j}
=
o_p(N).
\end{equation}

We next control the smallest singular value of
$\widetilde\BBB$.
Since
$\BBB^* = J^{-1/2}\UUU^*\DDDD^*\widehat\PP,$
we have
$\sigma_{K+2}(\BBB^*) = J^{-1/2} \sigma_{K+2}(\MM^*).$
By Assumptions~\ref{assumption:algo1_bounded} and
\ref{assumption:algo2_spikes},
$\sigma_{K+2}(\MM^*) \asymp \sqrt{NJ},$
and hence
$\sigma_{K+2}(\BBB^*) \asymp \sqrt N.$
Furthermore, by \eqref{eq:B3_B_F},
$$
\begin{aligned}
\|\widetilde\BBB-\BBB^*\|_2
&\le
\|\widetilde\BBB-\BBB^*\|_F\\
&=
O_p\left(
\sqrt N J^{-1/2}
L_{N,J}e_{\AAA,F}
\right)\\
&=
o_p(\sqrt N).
\end{aligned}
$$
Therefore, Weyl's inequality yields
\begin{equation}\label{eq:B3_B_singular}
\sigma_{K+2}(\widetilde\BBB)
\asymp
\sqrt N
\end{equation}
with probability tending to one.
In particular,
$\rank(\widetilde\BBB)=K+2$ with probability tending to one.

Since
$|m_{ij}^*|\le C_0$,
Assumption~\ref{assumption:algo2_link} implies
$f'(m_{ij}^*) \ge \delta_1^* > 0.$
Thus,
$$
\III_{2,j}(\widetilde\BBB)
=
\sum_{i=1}^N
f'(m_{ij}^*)
\widetilde\bbb_i\widetilde\bbb_i^\top
\succeq
\delta_1^*
\widetilde\BBB^\top\widetilde\BBB.
$$
Together with \eqref{eq:B3_B_singular}, uniformly over $j\in[J]$,
\begin{equation}\label{eq:B3_I_col}
\sigma_{K+2}
\bigl(\III_{2,j}(\widetilde\BBB)\bigr)
\gtrsim
N
\end{equation}
with probability tending to one.

We now verify the two product-form conditions in the column-wise analogue of
\eqref{eq:key_equation_new}.
By \eqref{eq:B3_T_col_small} and \eqref{eq:B3_I_col},
$$
\max_{j\in[J]}
\frac{
2T_{2,j}
}{
\sigma_{K+2}
(\III_{2,j}(\widetilde\BBB))C_2
}
=
o_p(1).
$$
Hence, with probability tending to one,
$2T_{2,j} \le \sigma_{K+2} (\III_{2,j}(\widetilde\BBB))C_2$
simultaneously for all $j\in[J]$.

Furthermore, by
\eqref{eq:B3_gamma_col},
\eqref{eq:B3_T_col_small}, and
\eqref{eq:B3_I_col},
$$
\max_{j\in[J]}
\frac{
4\gamma_{2,j}(\widetilde\BBB)\kappa_2^*T_{2,j}
}{
\sigma_{K+2}^2
(\III_{2,j}(\widetilde\BBB))
}
=
o_p(1).
$$
Thus, with probability tending to one,
$$
4\gamma_{2,j}(\widetilde\BBB)\kappa_2^*T_{2,j}
\le
\sigma_{K+2}^2
(\III_{2,j}(\widetilde\BBB))
$$
simultaneously for all $j\in[J]$.

Therefore, the column-wise analogue of
Lemma~\ref{lem:local_bi_existence_2} applies simultaneously to all
$j\in[J]$ with probability tending to one.
Hence, for every $j\in[J]$, there exists a unique
$\widetilde\aaa_j$ satisfying
$$
S_{2,j}(\widetilde\aaa_j;\widetilde\BBB)
=
\mathbf 0_{K+2},
\qquad
\|\widetilde\aaa_j-\aaa_j^*\|
\le
C_2.
$$

Moreover,
$\|\widetilde\aaa_j-\aaa_j^*\| \lesssim N^{-1}T_{2,j}.$
Taking the maximum over $j\in[J]$ yields
$$
\begin{aligned}
\|\widetilde\AAA-\AAA^*\|_{2\to\infty}
=
O_p\Bigl(&
N^{-1/2}L_{N,J}
+
J^{-1/2}L_{N,J}^2e_{\AAA,F}\\
&+
J^{-1/2}L_{N,J}e_{\AAA,F}
+
J^{-1}L_{N,J}^2e_{\AAA,F}^2
\Bigr).
\end{aligned}
$$
Since $L_{N,J}\ge1$ and
$e_{\AAA,F}
=o(\sqrt J L_{N,J}^{-2})$,
the last two terms are dominated by the first two.
Therefore,
$$
\|\widetilde\AAA-\AAA^*\|_{2\to\infty}
=
O_p\left(
N^{-1/2}L_{N,J}
+
J^{-1/2}L_{N,J}^2e_{\AAA,F}
\right).
$$
Since $L_{N,J}=\log(NJ)$,
$$
\|\widetilde\AAA-\AAA^*\|_{2\to\infty}
=
O_p\left(
N^{-1/2}\log(NJ)
+
J^{-1/2}(\log(NJ))^2e_{\AAA,F}
\right).
$$
\end{proof}

\subsubsection{Completion of the Bipartite Proof of Theorem~\ref{thm:refinement}}

\begin{proof}
Let
$e_{\AAA,F} := \sqrt J\, \sigma_{K+2}^{-1}(\MM^*)e_{\MM,F}.$

By Assumptions~\ref{assumption:algo1_bounded} and
\ref{assumption:algo2_spikes},
$\sigma_{K+2}(\MM^*) \asymp \sqrt{NJ}.$
Hence,
\begin{equation}\label{eq:B4_eA_rate}
e_{\AAA,F}
\asymp
N^{-1/2}e_{\MM,F}.
\end{equation}

We first verify that Lemma~\ref{lem:error_of_a} is applicable.
By the assumption
$\|\MM^{\mathrm{init}}-\MM^*\|_F = O_p(e_{\MM,F})$
and
$e_{\MM,F} \ll \sqrt{NJ}(\log(NJ))^{-2},$
we have
$$
\frac{
\|\MM^{\mathrm{init}}-\MM^*\|_F
}{
\sigma_{K+2}(\MM^*)
}
=
O_p\left(
\frac{e_{\MM,F}}{\sqrt{NJ}}
\right)
=
o_p(1).
$$
Therefore,
$$
\Pr\left(
\|\MM^{\mathrm{init}}-\MM^*\|_F
\le
\frac12
\sigma_{K+2}(\MM^*)
\right)
\to1.
$$
Lemma~\ref{lem:error_of_a} thus gives
\begin{equation}\label{eq:B4_A_init}
\|\widehat\AAA-\AAA^*\|_F
=
O_p(e_{\AAA,F}).
\end{equation}

We next verify the rate conditions required by
Lemmas~\ref{lem:2 to infty beta} and~\ref{lem:2 to infty alpha}.
Since
$\sqrt N\lesssim e_{\MM,F},$
\eqref{eq:B4_eA_rate} implies
$e_{\AAA,F} = \Omega(1).$
Moreover,
$e_{\MM,F} \ll \sqrt{NJ}(\log(NJ))^{-2}$
implies
\begin{equation}\label{eq:B4_eA_upper}
e_{\AAA,F}
=
o\bigl(
\sqrt J(\log(NJ))^{-2}
\bigr).
\end{equation}
In particular,
$e_{\AAA,F} = o\bigl( \sqrt J(\log(NJ))^{-1} \bigr).$
Therefore, all the conditions of
Lemmas~\ref{lem:2 to infty beta} and~\ref{lem:2 to infty alpha}
are satisfied.

It follows that
\begin{equation}\label{eq:B4_B_rate}
\|\widetilde\BBB-\BBB^*\|_{2\to\infty}
=
O_p\bigl(
J^{-1/2}\log(NJ)e_{\AAA,F}
\bigr),
\end{equation}
and
\begin{equation}\label{eq:B4_A_rate}
\|\widetilde\AAA-\AAA^*\|_{2\to\infty}
=
O_p\left(
N^{-1/2}\log(NJ)
+
J^{-1/2}(\log(NJ))^2e_{\AAA,F}
\right).
\end{equation}

By \eqref{eq:B4_eA_upper},
$J^{-1/2}\log(NJ)e_{\AAA,F} = o(1).$
Together with
$\|\BBB^*\|_{2\to\infty}\lesssim1$,
\eqref{eq:B4_B_rate} yields
\begin{equation}\label{eq:B4_B_bounded}
\|\widetilde\BBB\|_{2\to\infty}
=
O_p(1).
\end{equation}

For bipartite data, by construction,
$$
\widehat\MM^{\mathrm{ref}}
=
\widetilde\BBB\widetilde\AAA^\top,
\qquad
\MM^*
=
\BBB^*\AAA^{*\top}.
$$
Therefore,
$$
\begin{aligned}
\|\widehat\MM^{\mathrm{ref}}-\MM^*\|_{\max}
&=
\|
\widetilde\BBB\widetilde\AAA^\top
-
\BBB^*\AAA^{*\top}
\|_{\max}\\
&\le
\|
(\widetilde\BBB-\BBB^*)\AAA^{*\top}
\|_{\max}
+
\|
\widetilde\BBB
(\widetilde\AAA-\AAA^*)^\top
\|_{\max}\\
&\le
\|\widetilde\BBB-\BBB^*\|_{2\to\infty}
\|\AAA^*\|_{2\to\infty}\\
&\quad+
\|\widetilde\BBB\|_{2\to\infty}
\|\widetilde\AAA-\AAA^*\|_{2\to\infty}.
\end{aligned}
$$
Since
$\|\AAA^*\|_{2\to\infty}\lesssim1$, combining
\eqref{eq:B4_B_rate}--\eqref{eq:B4_B_bounded} gives
$$
\begin{aligned}
\|\widehat\MM^{\mathrm{ref}}-\MM^*\|_{\max}
=
O_p\Bigl(
&
J^{-1/2}\log(NJ)e_{\AAA,F}\\
&+
N^{-1/2}\log(NJ)\\
&+
J^{-1/2}(\log(NJ))^2e_{\AAA,F}
\Bigr).
\end{aligned}
$$

The first term is dominated by the third term. Moreover, by
\eqref{eq:B4_eA_rate},
$$
J^{-1/2}(\log(NJ))^2e_{\AAA,F}
\asymp
(NJ)^{-1/2}(\log(NJ))^2e_{\MM,F}.
$$
For the remaining term, since
$e_{\MM,F}\gtrsim\sqrt N$ and $N\ge J$,
$$
\begin{aligned}
N^{-1/2}\log(NJ)
&\lesssim
(NJ)^{-1/2}
\log(NJ)e_{\MM,F}\\
&\le
(NJ)^{-1/2}
(\log(NJ))^2e_{\MM,F}.
\end{aligned}
$$
Consequently,
$$
\|\widehat\MM^{\mathrm{ref}}-\MM^*\|_{\max}
=
O_p\left(
(NJ)^{-1/2}
(\log(NJ))^2
e_{\MM,F}
\right).
$$
This proves Theorem~\ref{thm:refinement} for the bipartite case.
\end{proof}

\subsection{Unipartite One-Step Refinement}\label{app:uni-one-step-refinement}

We now prove Theorem~\ref{thm:refinement} for unipartite data.
The argument follows the same factor-refinement logic as in Section~B.2,
with three structural modifications: diagonal observations are excluded
from the estimating equations, the observation matrix is symmetric, and
the final estimator is symmetrized and truncated to rank $K+2$.

Let
$r:=K+2, \qquad L_N:=\log(N^2),$
and define, as in the bipartite proof,
$$
\widehat\PP
\in
\argmin_{\PP\in\OOO_r}
\|\widehat\VVV-\VVV^*\PP\|_F,
$$
together with the aligned factors
$$
\AAA^*
:=
\sqrt N\,\VVV^*\widehat\PP,
\qquad
\BBB^*
:=
N^{-1/2}\UUU^*\DDDD^*\widehat\PP.
$$
Then
$\MM^* = \BBB^*\AAA^{*\top}.$
By Assumptions~\ref{assumption:algo1_bounded} and
\ref{assumption:algo2_incoherence},
$$
\|\AAA^*\|_{2\to\infty}\lesssim1,
\qquad
\|\BBB^*\|_{2\to\infty}\lesssim1.
$$

Define
$e_{\AAA,F} := \sqrt N\, \sigma_r^{-1}(\MM^*)e_{\MM,F}.$
Since $J=N$, Assumptions~\ref{assumption:algo1_bounded} and
\ref{assumption:algo2_spikes} imply
$\sigma_r(\MM^*) \asymp N,$
and hence
\begin{equation}\label{eq:B_uni_eA_rate}
e_{\AAA,F}
\asymp
N^{-1/2}e_{\MM,F}.
\end{equation}
The assumptions on $e_{\MM,F}$ therefore give
\begin{equation}\label{eq:B_uni_eA_conditions}
e_{\AAA,F}
=
\Omega(1),
\qquad
e_{\AAA,F}
=
o\left(
\sqrt N\,L_N^{-2}
\right).
\end{equation}

Moreover,
$$
\frac{
\|\MM^{\mathrm{init}}-\MM^*\|_F
}{
\sigma_r(\MM^*)
}
=
O_p\left(
\frac{e_{\MM,F}}{N}
\right)
=
o_p(1).
$$
Thus, with probability tending to one,
$\|\MM^{\mathrm{init}}-\MM^*\|_F \le \frac12\sigma_r(\MM^*).$
Applying Lemma~\ref{lem:error_of_a} with $J=N$ gives
\begin{equation}\label{eq:B_uni_A_init}
\|\widehat\AAA-\AAA^*\|_F
=
O_p(e_{\AAA,F}).
\end{equation}

We next consider the two refinement steps. The deterministic arguments in
Lemmas~\ref{lem:local_bi_existence_1} and
\ref{lem:local_bi_existence_2} remain valid after deleting the single
diagonal summand from each estimating equation. The probabilistic arguments
in Lemmas~\ref{lem:item_1}--\ref{lem:item_4} and in the proofs of
Lemmas~\ref{lem:2 to infty beta} and
\ref{lem:2 to infty alpha} also remain valid with $J=N$. Indeed, for each
fixed row $i$, the collection
$\{y_{ij}-f(m_{ij}^*):j\ne i\}$ consists of independent centered
sub-exponential random variables, and the same is true for each fixed
column $j$. Dependence across different rows or columns caused by symmetry
does not affect these arguments, since concentration is applied to each
fixed row or column and the resulting bounds are combined by a union bound.
We give below the modifications needed to account for the omitted diagonal
terms.

For the row refinement, define
$$
S_{1,i}^{\mathrm{uni}}(\bbb_i;\AAA)
:=
\sum_{j\ne i}
\{y_{ij}-f(\aaa_j^\top\bbb_i)\}\aaa_j.
$$
The analogues of Lemmas~\ref{lem:item_1}--\ref{lem:item_4}, together with
\eqref{eq:B_uni_A_init}, give
\begin{equation}\label{eq:B_uni_row_bounds}
\begin{aligned}
\max_{i\in[N]}
\left\|
\sum_{j\ne i}
\{y_{ij}-f(m_{ij}^*)\}\widehat\aaa_j
\right\|
&=
O_p\left(
\sqrt N\,L_N e_{\AAA,F}
\right),\\
\max_{i\in[N]}
\|\BB_{1,i}^{\mathrm{uni}}(\widehat\AAA)\|
&=
O_p\left(
\sqrt N\,e_{\AAA,F}
\right),\\
\max_{i\in[N]}
\beta_{1,i}^{\mathrm{uni}}(\widehat\AAA)
&=
O_p(e_{\AAA,F}^2),\\
\max_{i\in[N]}
\gamma_{1,i}^{\mathrm{uni}}(\widehat\AAA)
&\lesssim N,
\end{aligned}
\end{equation}
where the quantities with superscript $\mathrm{uni}$ are defined as their
bipartite counterparts with the summation restricted to $j\ne i$.

It remains to check that deleting the $i$th summand does not destroy the
information lower bound. By \eqref{eq:B_uni_A_init},
\eqref{eq:B_uni_eA_conditions}, and Weyl's inequality,
$\sigma_r(\widehat\AAA) \asymp \sqrt N$
with probability tending to one. Since
$\|\widehat\AAA\|_{2\to\infty}\le C_2$ by construction,
uniformly over $i\in[N]$,
$$
\begin{aligned}
\lambda_{\min}
\left(
\sum_{j\ne i}
\widehat\aaa_j\widehat\aaa_j^\top
\right)
&=
\lambda_{\min}
\left(
\widehat\AAA^\top\widehat\AAA
-
\widehat\aaa_i\widehat\aaa_i^\top
\right)\\
&\ge
\sigma_r^2(\widehat\AAA)
-
\|\widehat\aaa_i\|^2
\gtrsim N.
\end{aligned}
$$
Since $f'(m_{ij}^*)\ge\delta_1^*>0$,
\begin{equation}\label{eq:B_uni_I_row}
\min_{i\in[N]}
\sigma_r
\left(
\III_{1,i}^{\mathrm{uni}}(\widehat\AAA)
\right)
\gtrsim N
\end{equation}
with probability tending to one.

Combining \eqref{eq:B_uni_row_bounds},
\eqref{eq:B_uni_I_row}, and
\eqref{eq:B_uni_eA_conditions} verifies the two conditions in the
unipartite analogue of \eqref{eq:key_equation_new}, uniformly over
$i\in[N]$. Hence, with probability tending to one, every row-wise estimating
equation admits a unique solution, and the same argument as in
Lemma~\ref{lem:2 to infty beta} yields
\begin{equation}\label{eq:B_uni_B_rate}
\|\widetilde\BBB-\BBB^*\|_{2\to\infty}
=
O_p\left(
N^{-1/2}L_Ne_{\AAA,F}
\right).
\end{equation}
In view of \eqref{eq:B_uni_eA_conditions},
$\|\widetilde\BBB-\BBB^*\|_{2\to\infty} = o_p(1).$
Using the fixed slack in the choice of $C_1$ from the bipartite proof, we
may therefore work on an event whose probability tends to one on which
\begin{equation}\label{eq:B_uni_B_bounded}
\|\widetilde\BBB\|_{2\to\infty}
\le C_1.
\end{equation}
Furthermore,
\begin{equation}\label{eq:B_uni_B_F}
\|\widetilde\BBB-\BBB^*\|_F
\le
\sqrt N
\|\widetilde\BBB-\BBB^*\|_{2\to\infty}
=
O_p(L_Ne_{\AAA,F}).
\end{equation}

For the column refinement, define
$$
S_{2,j}^{\mathrm{uni}}(\aaa_j;\BBB)
:=
\sum_{i\ne j}
\{y_{ij}-f(\aaa_j^\top\bbb_i)\}\bbb_i.
$$
Repeating the proof of Lemma~\ref{lem:2 to infty alpha}, with the diagonal
summand omitted and using \eqref{eq:B_uni_B_F}, gives
\begin{equation}\label{eq:B_uni_col_bounds}
\begin{aligned}
\max_{j\in[N]}
\left\|
\sum_{i\ne j}
\{y_{ij}-f(m_{ij}^*)\}\widetilde\bbb_i
\right\|
&=
O_p\left(
\sqrt N\,L_N
+
\sqrt N\,L_N^2e_{\AAA,F}
\right),\\
\max_{j\in[N]}
\|\BB_{2,j}^{\mathrm{uni}}(\widetilde\BBB)\|
&=
O_p\left(
\sqrt N\,L_Ne_{\AAA,F}
\right),\\
\max_{j\in[N]}
\beta_{2,j}^{\mathrm{uni}}(\widetilde\BBB)
&=
O_p\left(
L_N^2e_{\AAA,F}^2
\right).
\end{aligned}
\end{equation}
On the event \eqref{eq:B_uni_B_bounded},
the column-wise analogue of Lemma~\ref{lem:item_4} further gives
$$
\max_{j\in[N]}
\gamma_{2,j}^{\mathrm{uni}}(\widetilde\BBB)
\lesssim N.
$$

Also,
$\sigma_r(\BBB^*) = N^{-1/2}\sigma_r(\MM^*) \asymp \sqrt N.$
By \eqref{eq:B_uni_B_F} and
\eqref{eq:B_uni_eA_conditions},
$$
\|\widetilde\BBB-\BBB^*\|_2
\le
\|\widetilde\BBB-\BBB^*\|_F
=
o_p(\sqrt N),
$$
and hence
$\sigma_r(\widetilde\BBB) \asymp \sqrt N$
with probability tending to one. Together with
\eqref{eq:B_uni_B_bounded}, this yields, uniformly over $j\in[N]$,
$$
\begin{aligned}
\lambda_{\min}
\left(
\sum_{i\ne j}
\widetilde\bbb_i\widetilde\bbb_i^\top
\right)
&=
\lambda_{\min}
\left(
\widetilde\BBB^\top\widetilde\BBB
-
\widetilde\bbb_j\widetilde\bbb_j^\top
\right)\\
&\ge
\sigma_r^2(\widetilde\BBB)
-
\|\widetilde\bbb_j\|^2
\gtrsim N.
\end{aligned}
$$
Therefore,
\begin{equation}\label{eq:B_uni_I_col}
\min_{j\in[N]}
\sigma_r
\left(
\III_{2,j}^{\mathrm{uni}}(\widetilde\BBB)
\right)
\gtrsim N
\end{equation}
with probability tending to one.

Equations \eqref{eq:B_uni_col_bounds},
\eqref{eq:B_uni_I_col}, and
\eqref{eq:B_uni_eA_conditions} verify the column-wise analogues of the
conditions in \eqref{eq:key_equation_new}. Thus, with probability tending
to one, every column-wise estimating equation admits a unique solution, and
the same argument as in Lemma~\ref{lem:2 to infty alpha} gives
\begin{equation}\label{eq:B_uni_A_rate}
\|\widetilde\AAA-\AAA^*\|_{2\to\infty}
=
O_p\left(
N^{-1/2}L_N
+
N^{-1/2}L_N^2e_{\AAA,F}
\right).
\end{equation}

We first consider the estimator before the final rank truncation. Let
$\widetilde\MM_0 := \widetilde\BBB\widetilde\AAA^\top.$
Since $\MM^*=\BBB^*\AAA^{*\top}$,
$$
\begin{aligned}
\|\widetilde\MM_0-\MM^*\|_{\max}
&\le
\|\widetilde\BBB-\BBB^*\|_{2\to\infty}
\|\AAA^*\|_{2\to\infty}\\
&\quad+
\|\widetilde\BBB\|_{2\to\infty}
\|\widetilde\AAA-\AAA^*\|_{2\to\infty}.
\end{aligned}
$$
Combining \eqref{eq:B_uni_B_rate},
\eqref{eq:B_uni_B_bounded}, and
\eqref{eq:B_uni_A_rate} yields
$$
\|\widetilde\MM_0-\MM^*\|_{\max}
=
O_p\left(
N^{-1/2}L_N
+
N^{-1/2}L_N^2e_{\AAA,F}
\right).
$$
By \eqref{eq:B_uni_eA_rate},
$N^{-1/2}L_N^2e_{\AAA,F} \asymp N^{-1}L_N^2e_{\MM,F},$
whereas $e_{\MM,F}\gtrsim\sqrt N$ implies
$N^{-1/2}L_N \lesssim N^{-1}L_N^2e_{\MM,F}.$
Consequently,
\begin{equation}\label{eq:B_uni_raw_max}
\|\widetilde\MM_0-\MM^*\|_{\max}
=
O_p(\varepsilon_N),
\qquad
\varepsilon_N
:=
N^{-1}L_N^2e_{\MM,F}
=
o(1).
\end{equation}

Define the symmetrized estimator
$$
\overline\MM
:=
\frac{
\widetilde\BBB\widetilde\AAA^\top
+
\widetilde\AAA\widetilde\BBB^\top
}{2}.
$$
Since $\MM^*$ is symmetric,
\begin{equation}\label{eq:B_uni_sym_max}
\|\overline\MM-\MM^*\|_{\max}
\le
\|\widetilde\MM_0-\MM^*\|_{\max}
=
O_p(\varepsilon_N).
\end{equation}

It remains to verify that the final rank-$r$ truncation preserves this
entrywise rate. We use the following deterministic lemma.

\begin{lemma}\label{lem:uni_svd_max_stability}
Let $\MM,\widetilde\MM\in\R^{N\times N}$ be symmetric and suppose that
$\rank(\MM)=r$, where $r$ is fixed. Assume
$$
\sigma_r(\MM)\gtrsim N,
\qquad
\|\MM\|_2\lesssim N.
$$
Let $\UUU\in\R^{N\times r}$ contain orthonormal eigenvectors spanning the
nonzero eigenspace of $\MM$, and suppose
$$
\|\UUU\|_{2\to\infty}
\lesssim
N^{-1/2}.
$$
If, for some deterministic sequence $\varepsilon_N=o(1)$,
$$
\|\widetilde\MM-\MM\|_{\max}
\lesssim
\varepsilon_N,
$$
then
$$
\|\SVD_r(\widetilde\MM)-\MM\|_{\max}
\lesssim
\varepsilon_N.
$$
\end{lemma}

\begin{proof}
Let
$\Delta_{\MM}:=\widetilde\MM-\MM.$
Then
$$
\|\Delta_{\MM}\|_2
\le
\|\Delta_{\MM}\|_F
\le
N\|\Delta_{\MM}\|_{\max}
\lesssim
N\varepsilon_N
=
o(N).
$$
By Weyl's inequality, the $r$ singular values of $\widetilde\MM$
corresponding to the signal are of order $N$, whereas
$\sigma_{r+1}(\widetilde\MM) \le \|\Delta_{\MM}\|_2 = o(N).$

Let $\widehat\UUU\in\R^{N\times r}$ contain the eigenvectors corresponding
to the $r$ eigenvalues of $\widetilde\MM$ with largest absolute values, and
let $\widehat\Lambda$ be the corresponding signed eigenvalue matrix. Define
$$
\Pi:=\UUU\UUU^\top,
\qquad
\widehat\Pi:=\widehat\UUU\widehat\UUU^\top.
$$
Then
$$
\SVD_r(\widetilde\MM)
=
\widehat\Pi\widetilde\MM\widehat\Pi,
\qquad
\|\widehat\Lambda^{-1}\|_2
\lesssim
N^{-1}.
$$

We first control $\widehat\UUU$ row-wisely. Since
$\widetilde\MM\widehat\UUU=\widehat\UUU\widehat\Lambda$, for every
$i\in[N]$,
$$
\begin{aligned}
\|\ee_i^\top\widehat\UUU\|
&\le
\left(
\|\ee_i^\top\MM\widehat\UUU\|
+
\|\ee_i^\top\Delta_{\MM}\widehat\UUU\|
\right)
\|\widehat\Lambda^{-1}\|_2\\
&\le
\left(
\|\ee_i^\top\UUU\|\|\MM\|_2
+
\|\ee_i^\top\Delta_{\MM}\|
\right)
\|\widehat\Lambda^{-1}\|_2\\
&\lesssim
\left(
\sqrt N+\sqrt N\,\varepsilon_N
\right)N^{-1}
\lesssim
N^{-1/2}.
\end{aligned}
$$
Hence
\begin{equation}\label{eq:B_uni_Uhat_incoherent}
\|\widehat\UUU\|_{2\to\infty}
\lesssim
N^{-1/2}.
\end{equation}

Next, because $(\II_N-\Pi)\MM=0$,
$$
(\II_N-\Pi)\widehat\UUU
=
(\II_N-\Pi)\Delta_{\MM}\widehat\UUU
\widehat\Lambda^{-1}.
$$
Since
$$
\|\ee_i^\top\Delta_{\MM}\|
\le
\sqrt N\,\|\Delta_{\MM}\|_{\max}
\lesssim
\sqrt N\,\varepsilon_N
$$
and
$\|\ee_i^\top\Pi\| = \|\ee_i^\top\UUU\| \lesssim N^{-1/2},$
we obtain
$$
\begin{aligned}
\|\ee_i^\top(\II_N-\Pi)\widehat\UUU\|
&\le
\left\{
\|\ee_i^\top\Delta_{\MM}\|
+
\|\ee_i^\top\Pi\|\|\Delta_{\MM}\|_2
\right\}
\|\widehat\Lambda^{-1}\|_2\\
&\lesssim
N^{-1/2}\varepsilon_N.
\end{aligned}
$$
Therefore,
\begin{equation}\label{eq:B_uni_proj_left}
\|(\II_N-\Pi)\widehat\Pi\|_{2\to\infty}
\lesssim
N^{-1/2}\varepsilon_N.
\end{equation}

On the other hand, Wedin's perturbation bound gives
$$
\|(\II_N-\widehat\Pi)\Pi\|_2
\lesssim
\frac{\|\Delta_{\MM}\|_2}{\sigma_r(\MM)}
\lesssim
\varepsilon_N.
$$
Consequently,
$$
\begin{aligned}
\|\Pi(\II_N-\widehat\Pi)\|_{2\to\infty}
&=
\max_{i\in[N]}
\|
\ee_i^\top\UUU
\UUU^\top(\II_N-\widehat\Pi)
\|\\
&\le
\|\UUU\|_{2\to\infty}
\|
\UUU^\top(\II_N-\widehat\Pi)
\|_2\\
&\lesssim
N^{-1/2}\varepsilon_N.
\end{aligned}
$$
Since
$$
\widehat\Pi-\Pi
=
(\II_N-\Pi)\widehat\Pi
-
\Pi(\II_N-\widehat\Pi),
$$
we conclude that
\begin{equation}\label{eq:B_uni_projector_2inf}
\|\widehat\Pi-\Pi\|_{2\to\infty}
\lesssim
N^{-1/2}\varepsilon_N.
\end{equation}
Together with \eqref{eq:B_uni_Uhat_incoherent}, this also gives
$\|\widehat\Pi\|_{2\to\infty} \lesssim N^{-1/2}.$

Let
$\Delta_{\Pi}:=\widehat\Pi-\Pi.$
Since $\Pi\MM=\MM\Pi=\MM$,
$$
\SVD_r(\widetilde\MM)-\MM
=
\widehat\Pi\Delta_{\MM}\widehat\Pi
+
\Delta_{\Pi}\MM
+
\MM\Delta_{\Pi}
+
\Delta_{\Pi}\MM\Delta_{\Pi}.
$$
For the first term,
$$
\|\widehat\Pi\Delta_{\MM}\widehat\Pi\|_{\max}
\le
\|\widehat\Pi\|_{2\to\infty}^2
\|\Delta_{\MM}\|_2
\lesssim
\varepsilon_N.
$$
Moreover,
$$
\max_{j\in[N]}\|\MM\ee_j\|
\le
\|\MM\|_2\|\UUU\|_{2\to\infty}
\lesssim
\sqrt N.
$$
Thus, by \eqref{eq:B_uni_projector_2inf},
$$
\|\Delta_{\Pi}\MM\|_{\max}
+
\|\MM\Delta_{\Pi}\|_{\max}
\lesssim
\varepsilon_N.
$$
Finally,
$$
\|\Delta_{\Pi}\MM\Delta_{\Pi}\|_{\max}
\le
\|\Delta_{\Pi}\|_{2\to\infty}^2
\|\MM\|_2
\lesssim
\varepsilon_N^2
\lesssim
\varepsilon_N.
$$
Combining the preceding bounds gives
$\|\SVD_r(\widetilde\MM)-\MM\|_{\max} \lesssim \varepsilon_N,$
which completes the proof.
\end{proof}

We apply Lemma~\ref{lem:uni_svd_max_stability} with
$\widetilde\MM=\overline\MM, \qquad \MM=\MM^*.$
Assumption~\ref{assumption:algo2_incoherence} gives
$\|\UUU^*\|_{2\to\infty} \lesssim N^{-1/2},$
whereas Assumptions~\ref{assumption:algo1_bounded} and
\ref{assumption:algo2_spikes} imply
$\sigma_r(\MM^*) \asymp N, \qquad \|\MM^*\|_2 \lesssim N.$
Together with \eqref{eq:B_uni_sym_max},
Lemma~\ref{lem:uni_svd_max_stability} yields
$$
\left\|
\SVD_r(\overline\MM)-\MM^*
\right\|_{\max}
=
O_p(\varepsilon_N).
$$
By Algorithm~\ref{alg:rsd},
$\widehat\MM^{\mathrm{ref}} = \SVD_r(\overline\MM),$
and hence
$$
\|\widehat\MM^{\mathrm{ref}}-\MM^*\|_{\max}
=
O_p\left(
N^{-1}(\log(N^2))^2e_{\MM,F}
\right).
$$
Since $J=N$,
$N^{-1} = (NJ)^{-1/2}, \qquad \log(N^2)=\log(NJ),$
and therefore
$$
\|\widehat\MM^{\mathrm{ref}}-\MM^*\|_{\max}
=
O_p\left(
(NJ)^{-1/2}
(\log(NJ))^2e_{\MM,F}
\right).
$$
This proves Theorem~\ref{thm:refinement} for the unipartite case.

\subsection{Proof of Corollary~\ref{cor:refinement_algo1}}\label{app:proof-refinement-corollary}

\begin{proof}
We apply Theorem~\ref{thm:refinement} with
$\MM^{\mathrm{init}}=\widehat\MM^{\mathrm{SVD}}$.

By Corollary~\ref{cor:support_adaptive_truncation},
\[
\|\widehat\MM^{\mathrm{SVD}}-\MM^*\|_F
=
O_p(e_{\MM,F}),
\]
where
\[
e_{\MM,F}
=
\sqrt{NJ}\,
J^{-\frac{1}{2(K+2)}}
(\log J)^{\frac{K+4}{2(K+2)}}.
\]
Since $N=J^{O(1)}$, we have $\log(NJ)=O(\log J)$, and therefore
\[
\sqrt N
\lesssim
e_{\MM,F}
\ll
\sqrt{NJ}\{\log(NJ)\}^{-2}.
\]
Hence, Theorem~\ref{thm:refinement} gives
\[
\|\widehat\MM^{\mathrm{RSD}}-\MM^*\|_{\max}
=
O_p\left(
(\log(NJ))^2
J^{-\frac{1}{2(K+2)}}
(\log J)^{\frac{K+4}{2(K+2)}}
\right).
\]

If the response supports are uniformly bounded,
Corollary~\ref{cor:support_adaptive_truncation} gives
\[
\|\widehat\MM^{\mathrm{SVD}}-\MM^*\|_F
=
O_p\left(
\sqrt{NJ}\,
J^{-\frac{1}{2(K+2)}}
\right).
\]
Thus, we may take
\[
e_{\MM,F}
=
\sqrt{NJ}\,
J^{-\frac{1}{2(K+2)}}.
\]
The conditions of Theorem~\ref{thm:refinement} remain satisfied, and
therefore
\[
\|\widehat\MM^{\mathrm{RSD}}-\MM^*\|_{\max}
=
O_p\left(
(\log(NJ))^2
J^{-\frac{1}{2(K+2)}}
\right).
\]

For unipartite data, the same arguments apply with $J=N$, under the
additional condition
$f(\tau_0)\in[a_{NN},b_{NN}]$.
\end{proof}

\section{Proofs for Configuration Recovery}
\label{app:configuration-recovery-proofs}

The configuration recovery arguments differ substantially between the two
observation structures. We therefore present the unipartite proof first,
followed by the auxiliary results and perturbation analysis required for
the bipartite case.

Throughout Appendix~\ref{app:configuration-recovery-proofs}, the configuration recovery results are treated as
deterministic perturbation statements. Let $\xi=\xi_{NJ}>0$ denote a
deterministic sequence such that
$\|\widehat\MM-\MM^*\|_{\max}\le\xi, \qquad \xi=o(1).$
For the bipartite parameter space $\GGamma$, we write $C_\Gamma$ for the
fixed radius appearing in its definition, so that
$\|\rrr\|\le C_\Gamma$ for every $\rrr\in\GGamma$.

\subsection{Unipartite Configuration Recovery}

\begin{proof}
Let
\[
\BB^*
=
\frac{1}{2}\JJ_N\MM^*\JJ_N
=
\TTT^*\TTT^{*\top},
\qquad
\widehat\BB
=
\frac{1}{2}\JJ_N\widehat\MM\JJ_N,
\]
where the first equality follows from
$\EE_N^\top\TTT^*=\mathbf 0$.
Set
$\Delta_{\BB} = \widehat\BB-\BB^*.$
Since
$\|\widehat\MM-\MM^*\|_{\max}\le\xi$,
double centering gives
\begin{equation}
\label{eq:uni_gram_perturbation}
\|\Delta_{\BB}\|_{\max}\lesssim\xi,
\qquad
\|\Delta_{\BB}\|\lesssim N\xi,
\qquad
\|\Delta_{\BB}\|_{2\to\infty}
\lesssim
\sqrt N\,\xi.
\end{equation}

Let
$\GG = \TTT^{*\top}\TTT^*.$
By assumption,
$\lambda_{\min}(\GG)\gtrsim N$,
while Assumption~\ref{assumption:algo1_bounded} gives
$\lambda_{\max}(\GG)\lesssim N$.
Thus the $K$ nonzero eigenvalues of $\BB^*$ are of order $N$.

Write
\[
\Eigen_K(\widehat\BB)
=
\widehat\UU\widehat\Lambda\widehat\UU^\top,
\qquad
\widehat\Lambda
=
\diag(\widehat\lambda_1,\ldots,\widehat\lambda_K).
\]
By Weyl's inequality and
\eqref{eq:uni_gram_perturbation}, since $\xi=o(1)$,
$\widehat\lambda_K \gtrsim N$
for all sufficiently large $N$.
Hence
$(\widehat\lambda_k)_+=\widehat\lambda_k$
for $k\in[K]$, and the configuration estimator in
Section~\ref{subsec:method_conf_recover} can be written as
$\widehat\TTT = \widehat\UU\widehat\Lambda^{1/2}.$

For $k>K$, the corresponding population eigenvalues of $\BB^*$ are zero.
Hence Weyl's inequality gives
$|\widehat\lambda_k| \le \|\Delta_{\BB}\|, \qquad k>K.$
Therefore,
\[
\begin{aligned}
\left\|
\widehat\TTT\widehat\TTT^\top-\BB^*
\right\|
&=
\left\|
\Eigen_K(\widehat\BB)-\BB^*
\right\|\\
&\le
\left\|
\Eigen_K(\widehat\BB)-\widehat\BB
\right\|
+
\|\widehat\BB-\BB^*\|\\
&\lesssim
\|\Delta_{\BB}\|.
\end{aligned}
\]
Together with \eqref{eq:uni_gram_perturbation}, this yields
\begin{equation}
\label{eq:uni_rankK_perturbation}
\left\|
\widehat\TTT\widehat\TTT^\top-\BB^*
\right\|
\lesssim
N\xi.
\end{equation}

Let
$\PP^* = \TTT^*\GG^{-1}\TTT^{*\top}$
be the orthogonal projector onto the column space of $\TTT^*$.
We first control the component of $\widehat\TTT$ orthogonal to this
space. Since
$(\II_N-\PP^*)\BB^*=0$
and
$\widehat\BB\widehat\TTT = \widehat\TTT\widehat\Lambda,$
we have
\[
(\II_N-\PP^*)\widehat\TTT
=
(\II_N-\PP^*)\Delta_{\BB}\,
\widehat\TTT\widehat\Lambda^{-1}.
\]
Furthermore,
\[
\|\widehat\TTT\widehat\Lambda^{-1}\|
=
\|\widehat\UU\widehat\Lambda^{-1/2}\|
\lesssim
N^{-1/2}.
\]
Since
\[
\|\ee_i^\top\PP^*\|
\le
\|\ttt_i^*\|
\,
\|\GG^{-1}\TTT^{*\top}\|
\lesssim
N^{-1/2},
\]
we have
\[
\begin{aligned}
\|\ee_i^\top(\II_N-\PP^*)\Delta_{\BB}\|
&\le
\|\ee_i^\top\Delta_{\BB}\|
+
\|\ee_i^\top\PP^*\|
\|\Delta_{\BB}\|\\
&\lesssim
\sqrt N\,\xi.
\end{aligned}
\]
Therefore,
$\|(\II_N-\PP^*)\Delta_{\BB}\|_{2\to\infty} \lesssim \sqrt N\,\xi,$
and consequently
\begin{equation}
\label{eq:uni_orthogonal_component}
\|(\II_N-\PP^*)\widehat\TTT\|_{2\to\infty}
\lesssim
\xi.
\end{equation}

It remains to control the component within the column space of
$\TTT^*$. Define
$\HHH = \GG^{-1}\TTT^{*\top}\widehat\TTT.$
Then
$\PP^*\widehat\TTT = \TTT^*\HHH.$
Moreover,
\[
\HHH\HHH^\top-\II_K
=
\GG^{-1}\TTT^{*\top}
\left(
\widehat\TTT\widehat\TTT^\top
-
\TTT^*\TTT^{*\top}
\right)
\TTT^*\GG^{-1}.
\]
Since
\[
\|\GG^{-1}\TTT^{*\top}\|
=
\lambda_{\min}(\GG)^{-1/2}
\lesssim
N^{-1/2},
\]
\eqref{eq:uni_rankK_perturbation} yields
\begin{equation}
\label{eq:uni_inner_transform}
\|\HHH\HHH^\top-\II_K\|
\lesssim
\xi.
\end{equation}

Thus, for all sufficiently large $N$, $\HHH$ is nonsingular. Let
$\QQ = (\HHH\HHH^\top)^{-1/2}\HHH \in \OOO_K$
be its polar factor. By
\eqref{eq:uni_inner_transform},
\begin{equation}
\label{eq:uni_polar_transform}
\|\HHH-\QQ\|
\lesssim
\xi.
\end{equation}

Finally,
\[
\widehat\TTT-\TTT^*\QQ
=
(\II_N-\PP^*)\widehat\TTT
+
\TTT^*(\HHH-\QQ).
\]
Therefore, by \eqref{eq:uni_orthogonal_component}, \eqref{eq:uni_polar_transform}, and Assumption~\ref{assumption:algo1_bounded},
\[
\|\widehat\TTT-\TTT^*\QQ\|_{2\to\infty}
\lesssim
\xi.
\]
Since $\QQ\in\OOO_K$, taking this $\QQ$ and zero translation in the
definition of $d_c$ gives
\[
d_c(\widehat\TTT,\TTT^*)
\lesssim
\xi.
\]
This completes the proof.
\end{proof}

\subsection{Bipartite Configuration Recovery}

We first collect the auxiliary results needed for the perturbation analysis
and then prove Theorem~\ref{thm:bi_conf}.

\subsubsection{Auxiliary Lemmas and Proofs}

\begin{lemma}\label{lem:bi_alignment}
Suppose Assumption~\ref{assumption:algo1_bounded} and
Assumption~\ref{assumption:algo3_diag} hold. Then, for all sufficiently
large $N$ and $J$, there exists $\QQ_Z\in\mathcal S_K$ such that
\[
\|\widehat\Z\QQ_Z-\Z^*\|_{2\to\infty}
\lesssim
\xi.
\]
Moreover, the optimization problems based on $\widehat\Z$ and
$\widehat\Z\QQ_Z$ have the same minimum value, and their global
minimizers are in one-to-one correspondence under
\[
(\tau,\ve(\FF)^\top,\bb^\top)^\top
\mapsto
(\tau,\ve(\QQ_Z\FF)^\top,\bb^\top)^\top.
\]
In particular, if
\[
\widehat\rrr
=
(\widehat\tau,\ve(\widehat\FF)^\top,\widehat\bb^\top)^\top
\]
is a global minimizer of \eqref{eq:opti}, then, for $s=1,2$,
\[
\TTT_{\mathcal P_s}
\left(
\left(
\widehat\tau,\,
\ve(\QQ_Z\widehat\FF)^\top,\,
\widehat\bb^\top
\right)^\top;
\widehat\Z\QQ_Z
\right)
=
\TTT_{\mathcal P_s}(\widehat\rrr;\widehat\Z).
\]
\end{lemma}

\begin{proof}
Let
$\MM_c^* = \frac12\JJ_N\MM^*\JJ_J = \UU^*\SG^*\VV^{*\top}$
and
$\widehat\MM_c = \frac12\JJ_N\widehat\MM\JJ_J.$
Set
$\Delta_c = \widehat\MM_c-\MM_c^*.$
Since
$\|\widehat\MM-\MM^*\|_{\max}\le\xi$,
\[
\|\Delta_c\|_{\max}
\lesssim
\xi,
\qquad
\|\Delta_c\|
\lesssim
\sqrt{NJ}\,\xi.
\]

By Assumption~\ref{assumption:algo3_diag}, the relevant population
singular values are separated by gaps of order $\sqrt{NJ}$. Therefore,
the left and right singular vectors can be matched individually. The
only remaining ambiguity for each matched singular-vector pair is its
sign. Hence, by Weyl's inequality and Wedin's theorem, for all
sufficiently large $N$ and $J$, there exists
$\QQ_Z\in\mathcal S_K$ such that
\[
\|\widehat\UU\QQ_Z-\UU^*\|
+
\|\widehat\VV\QQ_Z-\VV^*\|
\lesssim
\xi,
\qquad
\|\widehat\SG-\SG^*\|
\lesssim
\sqrt{NJ}\,\xi.
\]
In particular,
$\|\widehat\SG^{-1}\| \lesssim (NJ)^{-1/2}$
and
$\|\widehat\SG^{-1}-\SG^{*-1}\| \lesssim \frac{\xi}{\sqrt{NJ}}.$

Since $\QQ_Z$ commutes with the diagonal matrix $\widehat\SG$,
\[
\begin{aligned}
\widehat\UU\QQ_Z-\UU^*
={}&
\Delta_c\widehat\VV\QQ_Z\widehat\SG^{-1}
+
\MM_c^*
(\widehat\VV\QQ_Z-\VV^*)
\widehat\SG^{-1}\\
&+
\UU^*\SG^*
(\widehat\SG^{-1}-\SG^{*-1}).
\end{aligned}
\]
Under Assumption~\ref{assumption:algo1_bounded},
$\|\MM_c^*\|_{2\to\infty} \lesssim \sqrt J$
and
$\|\UU^*\|_{2\to\infty} \lesssim N^{-1/2}.$
Moreover,
$\|\ee_i^\top\Delta_c\| \lesssim \sqrt J\,\xi.$
Hence
$\sqrt N\, \|\widehat\UU\QQ_Z-\UU^*\|_{2\to\infty} \lesssim \xi.$

Applying the same argument to $\widehat\MM_c^\top$ gives
$\sqrt J\, \|\widehat\VV\QQ_Z-\VV^*\|_{2\to\infty} \lesssim \xi.$
Since
\[
\widehat\WW
=
\frac{1}{\sqrt{NJ}}
\widehat\VV\widehat\SG,
\qquad
\WW^*
=
\frac{1}{\sqrt{NJ}}
\VV^*\SG^*,
\]
we further obtain
$\sqrt J\, \|\widehat\WW\QQ_Z-\WW^*\|_{2\to\infty} \lesssim \xi.$
Therefore,
$\|\widehat\Z\QQ_Z-\Z^*\|_{2\to\infty} \lesssim \xi.$

Since
$\QQ_Z^\top=\QQ_Z^{-1}=\QQ_Z$,
direct substitution into \eqref{eq:unified_TA} gives
\[
\sqrt N\,
\widehat\UU\QQ_Z(\QQ_Z\widehat\FF)
=
\sqrt N\,
\widehat\UU\widehat\FF
\]
and
\[
\sqrt J\,
\widehat\WW\QQ_Z
(\QQ_Z\widehat\FF)^{-\top}
+
\mathbf 1_J\widehat\bb^\top
=
\sqrt J\,
\widehat\WW\widehat\FF^{-\top}
+
\mathbf 1_J\widehat\bb^\top.
\]
Thus, for $s=1,2$,
\[
\TTT_{\mathcal P_s}
\left(
\left(
\widehat\tau,\,
\ve(\QQ_Z\widehat\FF)^\top,\,
\widehat\bb^\top
\right)^\top;
\widehat\Z\QQ_Z
\right)
=
\TTT_{\mathcal P_s}(\widehat\rrr;\widehat\Z),
\]
and consequently
\[
\MM
\left(
\left(
\widehat\tau,\,
\ve(\QQ_Z\widehat\FF)^\top,\,
\widehat\bb^\top
\right)^\top;
\widehat\Z\QQ_Z
\right)
=
\MM(\widehat\rrr;\widehat\Z).
\]

Moreover,
$(\QQ_Z\FF)^\top(\QQ_Z\FF) = \FF^\top\FF,$
so the transformation preserves $\GGamma$. Since the transformation is
its own inverse, it defines a bijection between the feasible sets of
the two problems while preserving the objective value. The claimed
one-to-one correspondence of global minimizers follows.
\end{proof}

\begin{lemma}\label{lem:bi_parameter_recovery}
Suppose Assumption~\ref{assumption:algo1_bounded} and
Assumption~\ref{assumption:algo3} hold, and suppose
\[
\xi^2=o(c_{NJ}),
\qquad
\rho_{NJ}
=
o\left(
\frac{q_{NJ}}{\sqrt{NJ}}
\right).
\]
Then, for all sufficiently large $N$ and $J$, the minimum in
\eqref{eq:opti} is attained.

Moreover, for any global minimizer
\[
\widehat\rrr
=
(\widehat\tau,\ve(\widehat\FF)^\top,\widehat\bb^\top)^\top
\]
and the sign matrix $\QQ_Z$ in Lemma~\ref{lem:bi_alignment},
\begin{equation}
\label{eq:bi_parameter_localization_statement}
\inf_{\widetilde\rrr\in[\rrr_0]}
\left\|
\left(
\widehat\tau,\,
\ve(\QQ_Z\widehat\FF)^\top,\,
\widehat\bb^\top
\right)^\top
-
\widetilde\rrr
\right\|
<
\rho_{NJ}.
\end{equation}
Furthermore, there exists $\QQ\in\mathcal S_K$ attaining the preceding
infimum such that
\[
|\widehat\tau-\tau_0|
+
\|\QQ_Z\widehat\FF-\FF_0\QQ\|_F
+
\|\widehat\bb-\QQ^\top\bb_0\|
\lesssim
\frac{\sqrt{NJ}}{q_{NJ}}\xi.
\]
\end{lemma}

\begin{proof}
Let
$\DD_0 = \partial\MM(\rrr_0;\Z^*).$
Since the parameter dimension is fixed and the rows of $\DD_0$ are
uniformly bounded,
$\|\DD_0\| \lesssim \sqrt{NJ}.$
Assumption~\ref{assumption:algo3_identifiability} therefore implies
$q_{NJ} \lesssim \sqrt{NJ}.$
Hence
$\rho_{NJ} = o\left( \frac{q_{NJ}}{\sqrt{NJ}} \right) = o(1).$

We first show that minimizing sequences stay uniformly away from singular
$\FF$. By Lemma~\ref{lem:bi_alignment} and
Assumption~\ref{assumption:algo3_diag},
\[
\lambda_{\min}
\left\{
(\widehat\WW\QQ_Z)^\top
(\widehat\WW\QQ_Z)
\right\}
=
\frac{\widehat\sigma_K^2}{NJ}
\gtrsim
1.
\]
Since the columns of $\widehat\UU\QQ_Z$ and
$\widehat\WW\QQ_Z$ are centered, for any $\rrr\in\GGamma$,
\[
\begin{aligned}
\frac{1}{NJ}
\sum_{i,j}
&
\left\|
\ttt_i(\rrr;\widehat\Z\QQ_Z)
-
\ttt_{N+j}(\rrr;\widehat\Z\QQ_Z)
\right\|^2\\
&\ge
\tr\left\{
\FF^{-1}
(\widehat\WW\QQ_Z)^\top
(\widehat\WW\QQ_Z)
\FF^{-\top}
\right\}\\
&\gtrsim
\sigma_{\min}^{-2}(\FF).
\end{aligned}
\]
Since
$|\tau| \le \|\rrr\| \le C_\Gamma,$
Jensen's inequality gives
\[
\frac{1}{NJ}
\|\MM(\rrr;\widehat\Z\QQ_Z)\|_F^2
\longrightarrow
\infty
\]
uniformly as
$\sigma_{\min}(\FF)\to0$.

Since $\FF_0$ and $\FF_0^{-1}$ are uniformly bounded,
Lemma~\ref{lem:bi_alignment} and the smoothness of
\eqref{eq:unified_TA}--\eqref{eq:unified_M} give
$\|\MM(\rrr_0;\widehat\Z\QQ_Z)-\MM^*\|_F \lesssim \sqrt{NJ}\,\xi.$
Hence
\[
\frac{1}{NJ}
l(\rrr_0;\widehat\Z\QQ_Z,\widehat\MM)
\lesssim
\xi^2,
\]
while Assumption~\ref{assumption:algo1_bounded} and
$\xi=o(1)$ imply
$\frac{1}{NJ} \|\widehat\MM\|_F^2 \lesssim 1.$
Therefore, there exists a fixed $c_F>0$ such that every minimizing
sequence eventually satisfies
$\sigma_{\min}(\FF) \ge c_F.$
The corresponding subset of $\GGamma$ is compact. Hence the aligned
problem attains its minimum. By the bijection in
Lemma~\ref{lem:bi_alignment}, the original problem
\eqref{eq:opti} also attains its minimum.

On the set
$\sigma_{\min}(\FF)\ge c_F$,
$\|\FF\|+\|\FF^{-1}\| \lesssim 1.$
Thus, Lemma~\ref{lem:bi_alignment} and smoothness imply
\begin{equation}
\label{eq:bi_plugin_bound}
\sup_{\substack{
\rrr\in\GGamma:\\
\sigma_{\min}(\FF)\ge c_F
}}
\|\MM(\rrr;\widehat\Z\QQ_Z)
-\MM(\rrr;\Z^*)\|_F
\lesssim
\sqrt{NJ}\,\xi.
\end{equation}

By Lemma~\ref{lem:bi_alignment},
\[
\overline\rrr
:=
\left(
\widehat\tau,\,
\ve(\QQ_Z\widehat\FF)^\top,\,
\widehat\bb^\top
\right)^\top
\]
is a global minimizer of the aligned problem. Hence, by optimality,
\begin{equation}
\label{eq:bi_optimal_residual}
\left\|
\MM(\overline\rrr;\widehat\Z\QQ_Z)
-
\widehat\MM
\right\|_F
\lesssim
\sqrt{NJ}\,\xi.
\end{equation}

We next localize this minimizer. Suppose, to the contrary, that
\[
\inf_{\widetilde\rrr\in[\rrr_0]}
\|\overline\rrr-\widetilde\rrr\|
\ge
\rho_{NJ}.
\]
By Assumption~\ref{assumption:algo3_global}, there exists a fixed
constant $c_1>0$ such that, for all sufficiently large $N$ and $J$,
$\|\MM(\overline\rrr;\Z^*)-\MM^*\|_F \ge c_1 \sqrt{NJc_{NJ}}.$
By \eqref{eq:bi_plugin_bound} and
$\|\widehat\MM-\MM^*\|_F \le \sqrt{NJ}\,\xi,$
there exists a fixed constant $c_2>0$ such that
\[
\begin{aligned}
\left\|
\MM(\overline\rrr;\widehat\Z\QQ_Z)
-
\widehat\MM
\right\|_F
&\ge
\|\MM(\overline\rrr;\Z^*)-\MM^*\|_F\\
&\quad
-
\|\MM(\overline\rrr;\widehat\Z\QQ_Z)
-\MM(\overline\rrr;\Z^*)\|_F\\
&\quad
-
\|\widehat\MM-\MM^*\|_F\\
&\ge
c_1\sqrt{NJc_{NJ}}
-
c_2\sqrt{NJ}\,\xi.
\end{aligned}
\]
Since
$\xi^2=o(c_{NJ}),$
we have
$\xi=o(\sqrt{c_{NJ}}).$
Thus, the preceding lower bound is of order
$\sqrt{NJc_{NJ}}$, which is asymptotically larger than the upper bound
$O(\sqrt{NJ}\xi)$ in \eqref{eq:bi_optimal_residual}. This is a
contradiction. Therefore,
\begin{equation}
\label{eq:bi_rho_localization}
\inf_{\widetilde\rrr\in[\rrr_0]}
\|\overline\rrr-\widetilde\rrr\|
<
\rho_{NJ}.
\end{equation}

It remains to establish the local growth bound. We first prove that,
uniformly for $\rrr\in\GGamma$ satisfying
$\|\rrr-\rrr_0\| \le \rho_{NJ},$
\begin{equation}
\label{eq:bi_local_growth}
\|\MM(\rrr;\Z^*)-\MM^*\|_F
\gtrsim
q_{NJ}\|\rrr-\rrr_0\|.
\end{equation}
Let
$\hh = \rrr-\rrr_0$
and write its $\FF$-component as $\HHH$.
Rotational invariance gives
\[
\mathcal N_0
=
\left\{
\left(
0,\ve(\FF_0\RR)^\top,
(-\RR\bb_0)^\top
\right)^\top:
\RR^\top=-\RR
\right\}
\subseteq
\operatorname{Null}(\DD_0).
\]
The parameter dimension is
$1+K^2+K$, whereas
$\dim(\mathcal N_0) = \frac{K(K-1)}{2}.$
Assumption~\ref{assumption:algo3_identifiability} gives
$\rank(\DD_0) \ge \frac{(K+1)(K+2)}{2}.$
Since
$\mathcal N_0\subseteq\operatorname{Null}(\DD_0)$,
the reverse dimension bound also holds, and hence
\[
\rank(\DD_0)
=
\frac{(K+1)(K+2)}{2},
\qquad
\operatorname{Null}(\DD_0)
=
\mathcal N_0.
\]

Let
\[
\mathcal T_0
=
\left\{
(\delta\tau,\ve(\HHH)^\top,\delta\bb^\top)^\top:
\Delta(\FF_0^\top\HHH+\HHH^\top\FF_0)=0
\right\}
\]
be the tangent space of the constraint
$\Delta(\FF^\top\FF)=0$
at $\rrr_0$.
Since $\rrr_0\in\GGamma$, write
$\FF_0^\top\FF_0 = \diag(d_{1,NJ},\ldots,d_{K,NJ}).$
Assumption~\ref{assumption:algo3_diag} gives
$d_{k,NJ} \longrightarrow d_k, \qquad k\in[K],$
where
$d_1>\cdots>d_K>0.$
Hence, for all sufficiently large $N,J$,
\[
\min_k d_{k,NJ}
\gtrsim
1,
\qquad
\min_{p\ne q}
|d_{p,NJ}-d_{q,NJ}|
\gtrsim
1.
\]

For
\[
\left(
0,\ve(\FF_0\RR)^\top,
(-\RR\bb_0)^\top
\right)^\top
\in
\mathcal N_0,
\]
we have, for $p\ne q$,
\[
\left[
\Delta\{
\FF_0^\top\FF_0\RR
-
\RR\FF_0^\top\FF_0
\}
\right]_{pq}
=
(d_{p,NJ}-d_{q,NJ})R_{pq}.
\]
Consequently,
\[
\left\|
\Delta\{
\FF_0^\top\FF_0\RR
-
\RR\FF_0^\top\FF_0
\}
\right\|_F
\gtrsim
\|\RR\|_F.
\]
Since $\FF_0$ is uniformly nonsingular and $\bb_0$ is uniformly
bounded, the norm of the corresponding element of $\mathcal N_0$ is
equivalent to $\|\RR\|_F$. Thus
$\mathcal T_0\cap\mathcal N_0 = \{0\},$
and the smallest principal angle between $\mathcal T_0$ and
$\mathcal N_0$ is uniformly bounded away from zero.

If
$\FF = \FF_0+\HHH$
is feasible, then
$\Delta(\FF_0^\top\HHH+\HHH^\top\FF_0) = -\Delta(\HHH^\top\HHH).$
Let
$\SSS = \Delta(\FF_0^\top\HHH+\HHH^\top\FF_0)$
and define a symmetric matrix $\AAA$ with zero diagonal by
$A_{pq} = \frac{S_{pq}}{d_{p,NJ}+d_{q,NJ}}, \qquad p\ne q.$
Since
$\min_kd_{k,NJ}\gtrsim1$,
$\|\AAA\|_F \lesssim \|\HHH\|_F^2,$
and
$\Delta\{ \FF_0^\top(\FF_0\AAA) + (\FF_0\AAA)^\top\FF_0 \} = \SSS.$
Therefore, replacing $\HHH$ by
$\HHH-\FF_0\AAA$
produces an element
$\hh_0\in\mathcal T_0$ satisfying
$\|\hh-\hh_0\| \lesssim \|\hh\|^2.$
By the uniform transversality of
$\mathcal T_0$ and $\mathcal N_0$,
$\operatorname{dist}(\hh_0,\mathcal N_0) \gtrsim \|\hh_0\|.$
Hence
\[
\begin{aligned}
\operatorname{dist}(\hh,\mathcal N_0)
&\ge
\operatorname{dist}(\hh_0,\mathcal N_0)
-
\|\hh-\hh_0\|\\
&\gtrsim
\|\hh\|
-
O(\|\hh\|^2)\\
&\gtrsim
\|\hh\|
\end{aligned}
\]
for all sufficiently small $\|\hh\|$. Equivalently,
$\|\Pi_{\mathcal N_0^\perp}\hh\| \gtrsim \|\hh\|.$

Since
$\operatorname{Null}(\DD_0) = \mathcal N_0,$
Assumption~\ref{assumption:algo3_identifiability} gives
\[
\|\DD_0\hh\|
=
\|\DD_0\Pi_{\mathcal N_0^\perp}\hh\|
\gtrsim
q_{NJ}\|\hh\|.
\]
The second derivatives of
$m_{ij}(\rrr;\Z^*)$
are uniformly bounded in a fixed neighborhood of $\rrr_0$, so
Taylor's theorem yields $\left\|\ve\bigl(\MM(\rrr;\Z^*)-\MM^*\bigr)-\DD_0\hh\right\| \lesssim\sqrt{NJ}\|\hh\|^2$.
Since
$\|\hh\| \le \rho_{NJ}$
and
$\rho_{NJ} = o\left( \frac{q_{NJ}}{\sqrt{NJ}} \right),$
the remainder is
$o(q_{NJ}\|\hh\|)$
uniformly. This proves \eqref{eq:bi_local_growth}.

Since the sign transformation defining $[\rrr_0]$ is an isometry,
preserves $\GGamma$, and leaves $\MM(\rrr;\Z^*)$ invariant, the same
bound holds in a $\rho_{NJ}$-neighborhood of every element of
$[\rrr_0]$.

Finally, \eqref{eq:bi_plugin_bound},
\eqref{eq:bi_optimal_residual}, and
$\|\widehat\MM-\MM^*\|_F \le \sqrt{NJ}\,\xi$
imply
\[
\|\MM(\overline\rrr;\Z^*)-\MM^*\|_F
\lesssim
\sqrt{NJ}\,\xi.
\]
Combining this with
\eqref{eq:bi_rho_localization} and
\eqref{eq:bi_local_growth} gives
\[
q_{NJ}
\inf_{\widetilde\rrr\in[\rrr_0]}
\|\overline\rrr-\widetilde\rrr\|
\lesssim
\sqrt{NJ}\,\xi.
\]
Since $[\rrr_0]$ is finite, the infimum is attained by some
$\widetilde\rrr_0\in[\rrr_0]$. By the definition of the
sign-equivalence class, there exists $\QQ\in\mathcal S_K$ such that
\[
\widetilde\rrr_0
=
\left(
\tau_0,\,
\ve(\FF_0\QQ)^\top,\,
(\QQ^\top\bb_0)^\top
\right)^\top.
\]
Hence,
\[
\left\|
\overline\rrr
-
\widetilde\rrr_0
\right\|
\lesssim
\frac{\sqrt{NJ}}{q_{NJ}}\xi,
\qquad
\left\|
\overline\rrr
-
\widetilde\rrr_0
\right\|
<
\rho_{NJ}.
\]
Since $K$ is fixed, equivalence of norms yields
\[
|\widehat\tau-\tau_0|
+
\|\QQ_Z\widehat\FF-\FF_0\QQ\|_F
+
\|\widehat\bb-\QQ^\top\bb_0\|
\lesssim
\frac{\sqrt{NJ}}{q_{NJ}}\xi.
\]
This proves the lemma.
\end{proof}

\begin{lemma}\label{lem:bi_configuration_stability}
Suppose Assumption~\ref{assumption:algo1_bounded} and
Assumption~\ref{assumption:algo3_diag} hold. Then there exist fixed
constants $\eta_0>0$ and $C_T>0$ such that, for any
\[
\rrr
=
(\tau,\ve(\FF)^\top,\bb^\top)^\top,
\qquad
\Z,
\qquad
\QQ\in\mathcal S_K
\]
satisfying
\[
|\tau-\tau_0|
+
\|\FF-\FF_0\QQ\|_F
+
\|\bb-\QQ^\top\bb_0\|
+
\|\Z-\Z^*\|_{2\to\infty}
\le
\eta_0,
\]
we have
\[
\max_{s=1,2}
\|\TTT_{\mathcal P_s}(\rrr;\Z)
-\TTT_{\mathcal P_s}^*\QQ\|_{2\to\infty}
\le
C_T
\left\{
\|\FF-\FF_0\QQ\|_F
+
\|\bb-\QQ^\top\bb_0\|
+
\|\Z-\Z^*\|_{2\to\infty}
\right\}.
\]
\end{lemma}

\begin{proof}
Write
$\Z = \bigl( \sqrt N\,\UU^\top, \sqrt J\,\WW^\top \bigr)^\top.$
For $\eta_0$ sufficiently small, $\FF$ and $\FF^{-1}$ are uniformly
bounded. By \eqref{eq:unified_TA},
\[
\begin{aligned}
\|\TTT_{\mathcal P_1}(\rrr;\Z)
-\TTT_{\mathcal P_1}^*\QQ\|_{2\to\infty}
&=
\|\sqrt N\,\UU\FF
-\sqrt N\,\UU^*\FF_0\QQ\|_{2\to\infty}\\
&\lesssim
\|\Z-\Z^*\|_{2\to\infty}
+
\|\FF-\FF_0\QQ\|_F.
\end{aligned}
\]

For the second configuration, since
$(\FF_0\QQ)^{-\top} = \FF_0^{-\top}\QQ$
and
$(\QQ^\top\bb_0)^\top = \bb_0^\top\QQ,$
we have
\[
\begin{aligned}
&
\|\TTT_{\mathcal P_2}(\rrr;\Z)
-\TTT_{\mathcal P_2}^*\QQ\|_{2\to\infty}\\
&\le
\sqrt J\,
\|\WW-\WW^*\|_{2\to\infty}
\|\FF^{-\top}\|\\
&\quad+
\sqrt J\,
\|\WW^*\|_{2\to\infty}
\|\FF^{-\top}-(\FF_0\QQ)^{-\top}\|\\
&\quad+
\|\bb-\QQ^\top\bb_0\|.
\end{aligned}
\]
Since $\FF$ and $\FF_0\QQ$ are uniformly nonsingular,
$\|\FF^{-\top}-(\FF_0\QQ)^{-\top}\| \lesssim \|\FF-\FF_0\QQ\|_F.$
The result follows.
\end{proof}

\subsubsection{Proof of Theorem~\ref{thm:bi_conf}}

\begin{proof}
Let
$\QQ_Z\in\mathcal S_K$
be given by Lemma~\ref{lem:bi_alignment}.
By Lemma~\ref{lem:bi_parameter_recovery}, the minimum in
\eqref{eq:opti} is attained. Let
\[
\widehat\rrr
=
(\widehat\tau,\ve(\widehat\FF)^\top,\widehat\bb^\top)^\top
\]
be any global minimizer, and define the aligned parameter
\[
\overline\rrr
=
\left(
\widehat\tau,\,
\ve(\QQ_Z\widehat\FF)^\top,\,
\widehat\bb^\top
\right)^\top.
\]
Lemma~\ref{lem:bi_parameter_recovery} gives a
$\QQ\in\mathcal S_K$, attaining the infimum in
\eqref{eq:bi_parameter_localization_statement}, such that
\[
\left\|
\overline\rrr
-
\left(
\tau_0,\,
\ve(\FF_0\QQ)^\top,\,
(\QQ^\top\bb_0)^\top
\right)^\top
\right\|
<
\rho_{NJ}
=
o(1),
\]
and
\[
|\widehat\tau-\tau_0|
+
\|\QQ_Z\widehat\FF-\FF_0\QQ\|_F
+
\|\widehat\bb-\QQ^\top\bb_0\|
\lesssim
\frac{\sqrt{NJ}}{q_{NJ}}\xi.
\]
Moreover, Lemma~\ref{lem:bi_alignment} gives
$\|\widehat\Z\QQ_Z-\Z^*\|_{2\to\infty} \lesssim \xi = o(1).$
Therefore, for all sufficiently large $N$ and $J$, the localization
bound and the alignment bound place
$\overline\rrr$ and $\widehat\Z\QQ_Z$ inside the fixed neighborhood
required by Lemma~\ref{lem:bi_configuration_stability}.

Applying Lemma~\ref{lem:bi_configuration_stability} with
$\Z = \widehat\Z\QQ_Z, \qquad \FF = \QQ_Z\widehat\FF,$
we obtain
\[
\begin{aligned}
\max_{s=1,2}
&
\left\|
\TTT_{\mathcal P_s}
\left(
\overline\rrr;
\widehat\Z\QQ_Z
\right)
-
\TTT_{\mathcal P_s}^*\QQ
\right\|_{2\to\infty}\\
&\lesssim
\xi
+
\frac{\sqrt{NJ}}{q_{NJ}}\xi.
\end{aligned}
\]
Since the rows of
$\partial\MM(\rrr_0;\Z^*)$
are uniformly bounded and the parameter dimension is fixed,
Assumption~\ref{assumption:algo3_identifiability} implies
$q_{NJ} \lesssim \sqrt{NJ}.$
Therefore,
$\xi \lesssim \frac{\sqrt{NJ}}{q_{NJ}}\xi,$
and hence
\[
\max_{s=1,2}
\left\|
\TTT_{\mathcal P_s}
\left(
\overline\rrr;
\widehat\Z\QQ_Z
\right)
-
\TTT_{\mathcal P_s}^*\QQ
\right\|_{2\to\infty}
\lesssim
\frac{\sqrt{NJ}}{q_{NJ}}\xi.
\]

Finally, Lemma~\ref{lem:bi_alignment} gives, for $s=1,2$,
\[
\TTT_{\mathcal P_s}
\left(
\overline\rrr;
\widehat\Z\QQ_Z
\right)
=
\TTT_{\mathcal P_s}(\widehat\rrr;\widehat\Z).
\]
Thus,
\[
\left\|
\TTT(\widehat\rrr;\widehat\Z)
-
\TTT^*\QQ
\right\|_{2\to\infty}
=
\max_{s=1,2}
\left\|
\TTT_{\mathcal P_s}(\widehat\rrr;\widehat\Z)
-
\TTT_{\mathcal P_s}^*\QQ
\right\|_{2\to\infty}
\lesssim
\frac{\sqrt{NJ}}{q_{NJ}}\xi.
\]
Since $\QQ\in\mathcal S_K\subset\OOO_K$, taking this $\QQ$ and zero
translation in the definition of $d_c$ gives
\[
d_c\!\left(
\TTT(\widehat\rrr;\widehat\Z),
\TTT^*
\right)
\lesssim
\frac{\sqrt{NJ}}{q_{NJ}}\xi.
\]
This completes the proof.
\end{proof}

\subsection{Proof of Corollary~\ref{cor:config_rsd}}

\begin{proof}
Let $\widehat\MM$ denote the output of Algorithm~\ref{alg:rsd}.
By Theorem~\ref{thm:converge in prob} when $f(x)=x$ and
Corollary~\ref{cor:refinement_algo1} when $f$ is nonlinear,
\[
\|\widehat\MM-\MM^*\|_{\max}
=
O_p(\delta_{NJ}).
\]
Under the standing condition $N=J^{O(1)}$,
$\delta_{NJ}=o(1)$ in both cases.

Fix any $\varepsilon>0$. By the definition of
$O_p(\delta_{NJ})$, there exists a constant
$M_\varepsilon>0$ such that, for all sufficiently large $N$ and $J$,
\[
\Pr\left(
\|\widehat\MM-\MM^*\|_{\max}
\le
M_\varepsilon\delta_{NJ}
\right)
\ge
1-\varepsilon.
\]
Denote this event by $\mathcal E_{NJ,\varepsilon}$.

For unipartite data, on $\mathcal E_{NJ,\varepsilon}$,
Theorem~\ref{thm:uni_conf} gives
\[
d_c(\widehat\TTT,\TTT^*)
\lesssim
M_\varepsilon\delta_{NJ}.
\]
Since $\varepsilon>0$ is arbitrary,
\[
d_c(\widehat\TTT,\TTT^*)
=
O_p(\delta_{NJ}).
\]

For bipartite data,
$\delta_{NJ}^2=o(c_{NJ})$ implies
$(M_\varepsilon\delta_{NJ})^2=o(c_{NJ})$.
Together with
$\rho_{NJ}=o(q_{NJ}/\sqrt{NJ})$,
Theorem~\ref{thm:bi_conf} gives, on
$\mathcal E_{NJ,\varepsilon}$,
\[
d_c(\widehat\TTT,\TTT^*)
\lesssim
M_\varepsilon
\frac{\sqrt{NJ}}{q_{NJ}}
\delta_{NJ}.
\]
Therefore,
\[
d_c(\widehat\TTT,\TTT^*)
=
O_p\left(
\frac{\sqrt{NJ}}{q_{NJ}}
\delta_{NJ}
\right).
\]

If the response supports are uniformly bounded, the sharper
nonlinear-link rate follows directly from
Corollary~\ref{cor:refinement_algo1}.
\end{proof}

\bibliographystyle{apalike}
\bibliography{ref}

@article{latala2005estimates,
  author  = {Lata{\l}a, Rafa{\l}},
  title   = {Some Estimates of Norms of Random Matrices},
  journal = {Proceedings of the American Mathematical Society},
  volume  = {133},
  number  = {5},
  pages   = {1273--1282},
  year    = {2005},
  doi     = {10.1090/S0002-9939-04-07800-1}
}

@book{vershynin2018high,
  author    = {Vershynin, Roman},
  title     = {High-Dimensional Probability: An Introduction with Applications in Data Science},
  publisher = {Cambridge University Press},
  series    = {Cambridge Series in Statistical and Probabilistic Mathematics},
  number    = {47},
  year      = {2018},
  doi       = {10.1017/9781108231596}
}

@article{xia2021tensor,
  title={Statistically Optimal and Computationally Efficient Low Rank Tensor
         Completion from Noisy Entries},
  author={Xia, Dong and Yuan, Ming and Zhang, Cun-Hui},
  journal={The Annals of Statistics},
  volume={49},
  number={1},
  pages={76--99},
  year={2021},
  doi={10.1214/20-AOS1942}
}

@article{lee2026deep,
  title={Deep Discrete Encoders: Identifiable Deep Generative Models for Rich Data with Discrete Latent Layers},
  author={Lee, Seunghyun and Gu, Yuqi},
  journal={Journal of the American Statistical Association},
  volume={121},
  number={553},
  pages={194--208},
  year={2026},
  doi={10.1080/01621459.2025.2587922}
}

@article{lyu2026spectral,
  title={Spectral Clustering with Likelihood Refinement for High-Dimensional Latent Class Recovery},
  author={Lyu, Zhongyuan and Gu, Yuqi},
  journal={Psychometrika},
  volume={91},
  number={2},
  pages={695--723},
  year={2026},
  doi={10.1017/psy.2026.10095}
}

@article{fan2016projected,
  title={Projected Principal Component Analysis in Factor Models},
  author={Fan, Jianqing and Liao, Yuan and Wang, Weichen},
  journal={The Annals of Statistics},
  volume={44},
  number={1},
  pages={219--254},
  year={2016},
  doi={10.1214/15-AOS1364}
}

@article{torgerson1952multidimensional,
  title={Multidimensional Scaling: I. Theory and Method},
  author={Torgerson, Warren S.},
  journal={Psychometrika},
  volume={17},
  number={4},
  pages={401--419},
  year={1952},
  doi={10.1007/BF02288916}
}

@article{schonemann1970metric,
  title={On Metric Multidimensional Unfolding},
  author={Sch{\"o}nemann, Peter H.},
  journal={Psychometrika},
  volume={35},
  number={3},
  pages={349--366},
  year={1970},
  doi={10.1007/BF02310794}
}

@book{borg2005modern,
  title={Modern Multidimensional Scaling: Theory and Applications},
  author={Borg, Ingwer and Groenen, Patrick J. F.},
  edition={2},
  publisher={Springer},
  address={New York},
  year={2005},
  doi={10.1007/0-387-28981-X}
}

@article{deleeuw2009smacof,
  title={Multidimensional Scaling Using Majorization: {SMACOF} in {R}},
  author={de Leeuw, Jan and Mair, Patrick},
  journal={Journal of Statistical Software},
  volume={31},
  number={3},
  pages={1--30},
  year={2009},
  doi={10.18637/jss.v031.i03}
}

@article{chen2021unfolding,
  title={Unfolding-Model-Based Visualization: Theory, Method and Applications},
  author={Chen, Yunxiao and Ying, Zhiliang and Zhang, Haoran},
  journal={Journal of Machine Learning Research},
  volume={22},
  number={11},
  pages={1--51},
  year={2021}
}

@article{hoff2002latent,
  title={Latent Space Approaches to Social Network Analysis},
  author={Hoff, Peter D. and Raftery, Adrian E. and Handcock, Mark S.},
  journal={Journal of the American Statistical Association},
  volume={97},
  number={460},
  pages={1090--1098},
  year={2002},
  doi={10.1198/016214502388618906}
}

@article{handcock2007model,
  title={Model-Based Clustering for Social Networks},
  author={Handcock, Mark S. and Raftery, Adrian E. and Tantrum, Jeremy M.},
  journal={Journal of the Royal Statistical Society: Series A},
  volume={170},
  number={2},
  pages={301--354},
  year={2007},
  doi={10.1111/j.1467-985X.2007.00471.x}
}

@article{ma2020universal,
  title={Universal Latent Space Model Fitting for Large Networks with Edge Covariates},
  author={Ma, Zhuang and Ma, Zongming and Yuan, Hongsong},
  journal={Journal of Machine Learning Research},
  volume={21},
  number={4},
  pages={1--67},
  year={2020}
}

@misc{li2025statistical,
  title={Statistical Inference on Latent Space Models for Network Data},
  author={Li, Jinming and Wu, Shihao and Cui, Chengyu and Xu, Gongjun and Zhu, Ji},
  year={2025},
  eprint={2312.06605},
  archivePrefix={arXiv},
  primaryClass={math.ST}
}

@article{bai2003inferential,
  title={Inferential Theory for Factor Models of Large Dimensions},
  author={Bai, Jushan},
  journal={Econometrica},
  volume={71},
  number={1},
  pages={135--171},
  year={2003},
  doi={10.1111/1468-0262.00392}
}

@article{chen2020structured,
  title={Structured Latent Factor Analysis for Large-Scale Data:
         Identifiability, Estimability, and Their Implications},
  author={Chen, Yunxiao and Li, Xiaoou and Zhang, Siliang},
  journal={Journal of the American Statistical Association},
  volume={115},
  number={532},
  pages={1756--1770},
  year={2020},
  doi={10.1080/01621459.2019.1635485}
}

@article{wang2022maximum,
  title={Maximum Likelihood Estimation and Inference for High Dimensional
         Generalized Factor Models with Application to Factor-Augmented Regressions},
  author={Wang, Fa},
  journal={Journal of Econometrics},
  volume={229},
  number={1},
  pages={180--200},
  year={2022},
  doi={10.1016/j.jeconom.2020.11.002}
}

@article{kidzinski2022generalized,
  title={Generalized Matrix Factorization: Efficient Algorithms for Fitting
         Generalized Linear Latent Variable Models to Large Data Arrays},
  author={Kidzi{\'n}ski, {\L}ukasz and Hui, Francis K. C. and
          Warton, David I. and Hastie, Trevor J.},
  journal={Journal of Machine Learning Research},
  volume={23},
  number={291},
  pages={1--29},
  year={2022}
}

@article{zhang2020note,
  title={A Note on Exploratory Item Factor Analysis by Singular Value Decomposition},
  author={Zhang, Haoran and Chen, Yunxiao and Li, Xiaoou},
  journal={Psychometrika},
  volume={85},
  number={2},
  pages={358--372},
  year={2020},
  doi={10.1007/s11336-020-09704-7}
}

@article{chen2024entrywise,
  title={A Note on Entrywise Consistency for Mixed-Data Matrix Completion},
  author={Chen, Yunxiao and Li, Xiaoou},
  journal={Journal of Machine Learning Research},
  volume={25},
  number={343},
  pages={1--66},
  year={2024}
}

@article{chen2024spectral,
  title={A Spectral Method for Identifiable Grade of Membership Analysis
         with Binary Responses},
  author={Chen, Ling and Gu, Yuqi},
  journal={Psychometrika},
  volume={89},
  number={2},
  pages={626--657},
  year={2024},
  doi={10.1007/s11336-024-09951-y}
}

@article{abbe2020entrywise,
  title={Entrywise Eigenvector Analysis of Random Matrices with Low Expected Rank},
  author={Abbe, Emmanuel and Fan, Jianqing and Wang, Kaizheng and Zhong, Yiqiao},
  journal={The Annals of Statistics},
  volume={48},
  number={3},
  pages={1452--1474},
  year={2020},
  doi={10.1214/19-AOS1854}
}

@article{agarwal2012noisy,
  author  = {Agarwal, Alekh and Negahban, Sahand and Wainwright, Martin J.},
  title   = {Noisy Matrix Decomposition via Convex Relaxation: Optimal Rates in High Dimensions},
  journal = {The Annals of Statistics},
  year    = {2012},
  volume  = {40},
  number  = {2},
  pages   = {1171--1197},
  doi     = {10.1214/12-AOS1000}
}

@article{greco2025twodimensional,
  author = {Greco, Stefano},
  title = {Two-Dimensional Political Competition and Europarties as Cleavage Mediators in the 2024 European Parliament Elections},
  journal = {JCMS: Journal of Common Market Studies},
  year = {2025},
  volume = {63},
  number = {6},
  pages = {1507-1525},
  doi = {10.1111/jcms.70069}
}

@article{hix2006dimensions,
  author = {Simon Hix and Abdul Noury and Gérard Roland},
  title = {Dimensions of Politics in the European Parliament},
  journal = {American Journal of Political Science},
  volume = {50},
  number = {2},
  pages = {494-520},
  year = {2006}
}

@article{li2020central,
  author  = {Li, Gongkai and Tang, Minh and Charon, Nicolas
             and Priebe, Carey E.},
  title   = {Central Limit Theorems for Classical
             Multidimensional Scaling},
  journal = {Electronic Journal of Statistics},
  year    = {2020},
  volume  = {14},
  number  = {1},
  pages   = {2362--2394},
  doi     = {10.1214/20-EJS1720}
}

@article{little2023analysis,
  author  = {Little, Anna and Xie, Yuying and Sun, Qiang},
  title   = {An Analysis of Classical Multidimensional Scaling
             with Applications to Clustering},
  journal = {Information and Inference:
             A Journal of the IMA},
  year    = {2023},
  volume  = {12},
  number  = {1},
  pages   = {72--112},
  doi     = {10.1093/imaiai/iaac004}
}

@article{zhang2016distance,
  author  = {Zhang, Luwan and Wahba, Grace and Yuan, Ming},
  title   = {Distance Shrinkage and Euclidean Embedding via
             Regularized Kernel Estimation},
  journal = {Journal of the Royal Statistical Society:
             Series B (Statistical Methodology)},
  year    = {2016},
  volume  = {78},
  number  = {4},
  pages   = {849--867},
  doi     = {10.1111/rssb.12138}
}

@article{hinton2006reducing,
  author  = {Hinton, Geoffrey E. and Salakhutdinov, Ruslan R.},
  title   = {Reducing the Dimensionality of Data with Neural Networks},
  journal = {Science},
  year    = {2006},
  volume  = {313},
  number  = {5786},
  pages   = {504--507},
  doi     = {10.1126/science.1127647}
}

@misc{mcinnes2018umap,
  author        = {McInnes, Leland and Healy, John and Melville, James},
  title         = {{UMAP}: Uniform Manifold Approximation and Projection
                   for Dimension Reduction},
  year          = {2018},
  eprint        = {1802.03426},
  archivePrefix = {arXiv},
  primaryClass  = {stat.ML}
}

@article{vandermaaten2008visualizing,
  author  = {van der Maaten, Laurens and Hinton, Geoffrey},
  title   = {Visualizing Data Using {t-SNE}},
  journal = {Journal of Machine Learning Research},
  year    = {2008},
  volume  = {9},
  pages   = {2579--2605}
}

@article{belkin2003laplacian,
  author  = {Belkin, Mikhail and Niyogi, Partha},
  title   = {Laplacian Eigenmaps for Dimensionality Reduction
             and Data Representation},
  journal = {Neural Computation},
  year    = {2003},
  volume  = {15},
  number  = {6},
  pages   = {1373--1396},
  doi     = {10.1162/089976603321780317}
}

@article{tenenbaum2000global,
  author  = {Tenenbaum, Joshua B. and de Silva, Vin and
             Langford, John C.},
  title   = {A Global Geometric Framework for Nonlinear
             Dimensionality Reduction},
  journal = {Science},
  year    = {2000},
  volume  = {290},
  number  = {5500},
  pages   = {2319--2323},
  doi     = {10.1126/science.290.5500.2319}
}

@misc{debodt2025low,
  author        = {de Bodt, Cyril and Diaz-Papkovich, Alex and
                   Bleher, Michael and Bunte, Kerstin and
                   Coupette, Corinna and Damrich, Sebastian and
                   Fita Sanmartin, Enrique and Hamprecht, Fred A. and
                   Horv{\'a}t, Em{\H{o}}ke-Agnes and Kohli, Dhruv and
                   Krishnaswamy, Smita and Lee, John A. and
                   Lelieveldt, Boudewijn P. F. and McInnes, Leland and
                   Nabney, Ian T. and Noichl, Maximilian and
                   Poli{\v{c}}ar, Pavlin G. and Rieck, Bastian and
                   Wolf, Guy and Mishne, Gal and Kobak, Dmitry},
  title         = {Low-Dimensional Embeddings of High-Dimensional Data},
  year          = {2025},
  eprint        = {2508.15929},
  archivePrefix = {arXiv},
  primaryClass  = {cs.LG}
}

@article{gabriel2002goodness,
  title={Goodness of fit of biplots and correspondence analysis},
  author={Gabriel, K Ruben},
  journal={Biometrika},
  volume={89},
  number={2},
  pages={423--436},
  year={2002},
  publisher={Oxford University Press}
}

@article{gabriel1971biplot,
  title={The biplot graphic display of matrices with application to principal component analysis},
  author={Gabriel, Karl Ruben},
  journal={Biometrika},
  volume={58},
  number={3},
  pages={453--467},
  year={1971},
  publisher={Oxford University Press}
}

@article{gower1966some,
  title={Some distance properties of latent root and vector methods used in multivariate analysis},
  author={Gower, John C.},
  journal={Biometrika},
  volume={53},
  number={3/4},
  pages={325--338},
  year={1966},
  publisher={Oxford University Press}
}

@article{kruskal1964multidimensional,
  title={Multidimensional scaling by optimizing goodness of fit to a nonmetric hypothesis},
  author={Kruskal, Joseph B},
  journal={Psychometrika},
  volume={29},
  pages={1--27},
  year={1964},
  publisher={Springer}
}

@article{busing2005avoiding,
  title={Avoiding degeneracy in multidimensional unfolding by penalizing on the coefficient of variation},
  author={Busing, Frank MTA and Groenen, Patrick JK and Heiser, Willem J},
  journal={Psychometrika},
  volume={70},
  pages={71--98},
  year={2005},
  publisher={Springer}
}

@article{bandeira2016sharp,
 author = {Afonso S. Bandeira and Ramon van Handel},
 journal = {The Annals of Probability},
 number = {4},
 pages = {2479--2506},
 title = {Sharp nonasymptotic bounds on the norm of random matrices with independent entries},
 volume = {44},
 year = {2016}
}

@book{skrondal2004generalized,
  title={Generalized latent variable modeling: Multilevel, longitudinal, and structural equation models},
  author={Skrondal, Anders and Rabe-Hesketh, Sophia},
  year={2004},
  publisher={Chapman and Hall/CRC}
}

@book{bartholomew2008analysis,
  title={Analysis of multivariate social science data},
  author={Bartholomew, David J and Steele, Fiona and Galbraith, Jane and Moustaki, Irini},
  year={2008},
  publisher={CRC press}
}

@article{chatterjee2015matrix,
author = {Sourav Chatterjee},
title = {Matrix estimation by Universal Singular Value Thresholding},
volume = {43},
journal = {The Annals of Statistics},
number = {1},
publisher = {Institute of Mathematical Statistics},
pages = {177 -- 214},
year = {2015},
doi = {10.1214/14-AOS1272},
URL = {https://doi.org/10.1214/14-AOS1272}
}

\end{document}